\documentclass{article} 
\usepackage{iclr2026_conference,times}

\usepackage{amsmath,amsfonts,bm}

\def\eqref#1{equation~\ref{#1}}

\def\1{\bm{1}}

\DeclareMathAlphabet{\mathsfit}{\encodingdefault}{\sfdefault}{m}{sl}
\SetMathAlphabet{\mathsfit}{bold}{\encodingdefault}{\sfdefault}{bx}{n}

\usepackage{hyperref}
\usepackage{url}

\usepackage[utf8]{inputenc}
\usepackage[T1]{fontenc}
\usepackage{algorithm}
\usepackage{algorithmic}
\usepackage{graphicx}
\usepackage{booktabs}
\usepackage{multirow}
\usepackage{amsmath}
\usepackage{amsfonts}
\usepackage{enumitem}
\usepackage{url} 
\usepackage{xspace}
\usepackage{xcolor}
\usepackage{hyperref}
\usepackage{epigraph}
\usepackage{amssymb}
\usepackage{wrapfig}
\usepackage{subcaption}

\definecolor{langblue}{rgb}{0, 0.4, 0.8}
\definecolor{langred}{rgb}{0.81, 0.09, 0.13}
\definecolor{langgreen}{rgb}{0.0, 0.6, 0.3}
\hypersetup{
    colorlinks=true,
    linkcolor=langblue,
    citecolor=langblue,
}

\makeatletter
\AtBeginDocument{%
  \let\oldref\ref
  \renewcommand{\ref}[1]{%
    \hyperref[{#1}]{\underline{\oldref{#1}}}%
  }%
}
\makeatother

\usepackage{titletoc}
\newcommand\DoToC{%
  \startcontents
  \printcontents{}{1}{\textbf{\large Contents of Appendix}\small\vskip3pt\hrule\vskip5pt}
  \vskip3pt\hrule\vskip5pt
}

\newcommand{\methodname}{TableMaster}
\newcommand{\method}{\textit{\methodname}\xspace}

\title{\textsc{\methodname}: A Recipe to Advance Table \\Understanding with Language Models}

\newcommand{\authicon}[2][2.5ex]{%
  \textsuperscript{%
    \raisebox{-0.15ex}{\includegraphics[height=#1]{#2}}%
  }%
}
\author{%
  Lang Cao\authicon{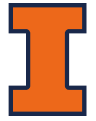} \hspace{0.3em} Hanbing Liu\authicon[2.8ex]{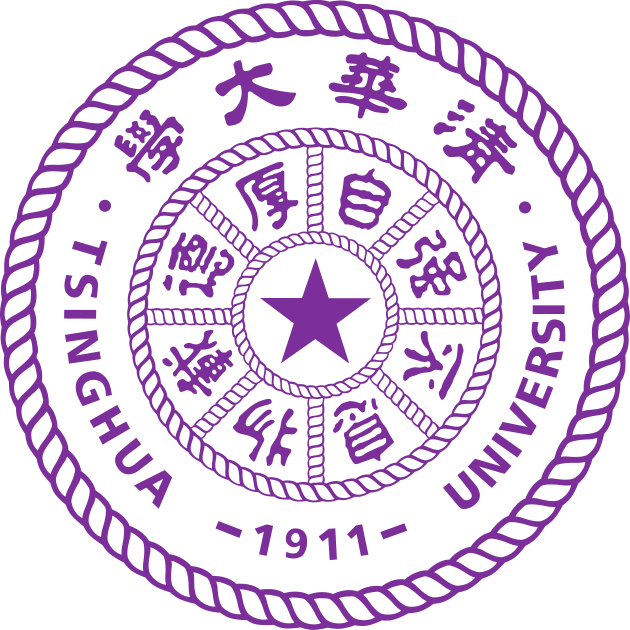} \\
  \authicon{icons/uiuc.png}University of Illinois Urbana-Champaign \quad\authicon[2.8ex]{icons/thu.png}Tsinghua University \\
  \texttt{langcao2@illinois.edu} \quad\texttt{liuhb24@mails.tsinghua.edu.cn} \\
}

\iclrfinalcopy 
\begin{document}

\maketitle



\begin{abstract}
Tables serve as a fundamental format for representing structured relational data. While current language models (LMs) excel at many text-based tasks, they still face challenges in table understanding due to the complex characteristics of tabular data, such as their structured nature. In this paper, we aim to enhance LMs for improved table understanding. We identify four key challenges: 1) difficulty in locating target data, 2) deficiency in table semantics, 3) numerical inaccuracies in textual reasoning, and 4) semantic inflexibility in symbolic reasoning. To address these issues, we propose \method, a recipe and comprehensive framework that integrates multiple solutions to overcome these obstacles. \method first extracts relevant table content and verbalizes it with enriched semantic context. Additionally, we introduce adaptive reasoning, a flexible approach that dynamically adjusts between textual and symbolic reasoning, tailoring the reasoning process to each query. Extensive analyses and experiments demonstrate our findings and the effectiveness of \method. On the WikiTQ dataset, \method achieves an accuracy of 78.13\% using GPT-4o-mini, surpassing existing baselines. We hope this work will serve as a practical step toward more robust and reliable table understanding.
\end{abstract}


\section{Introduction}
\setlength{\epigraphwidth}{0.6\textwidth}
\epigraph{\textit{“Data gains extraordinary power as it transcends the simplicity of one dimension to embrace the richness of higher dimensions.”}}{}

Tables are widely used in daily life and across various fields, such as healthcare \citep{ghasemi2016process} and finance \citep{li2020gftegraphbasedfinancialtable, yi2025tablepilot}, due to their unique ability to efficiently represent two-dimensional relational data. It is crucial to process tabular data with both efficiency and accuracy. Recently, large language models (LLMs) \citep{gunasekar2023textbooksneed, openai2024gpt4technicalreport, touvron2023llamaopenefficientfoundation} have achieved significant progress in the field of natural language processing. They perform well in a wide range of downstream text-based tasks, including language understanding \citep{minaee2024largelanguagemodelssurvey, zhu-etal-2024-large} and reasoning \citep{plaat2024reasoninglargelanguagemodels}. Naturally, language models (LMs) are increasingly being used to process and understand tabular data \citep{fang2024largelanguagemodelsllmstabular, zhang2024surveytablereasoninglarge}, enabling reasoning for downstream tasks such as table-based question answering \citep{pasupat-liang-2015-compositional} and table-based fact verification \citep{chen2020tabfactlargescaledatasettablebased}.

However, the data structure of tables inherently possess a unique two-dimensional structure that contrasts with the linear text, which dominates the content in language model pretraining corpora. Most advanced LMs are not specifically optimized for processing tabular data. While techniques such as chain-of-thought prompting \citep{wei2023chainofthoughtpromptingelicitsreasoning} and other reasoning-enhanced methods \citep{yao2023treethoughtsdeliberateproblem} have enabled LMs to perform satisfactorily in reasoning with linear text, significant room for improvement remains in table-based reasoning \citep{chen-2023-large}. A notable gap persists in LMs’ ability to fully understand tables and effectively reason with tabular data.

\begin{figure}[htbp]
    \centering
    \includegraphics[width=\textwidth]{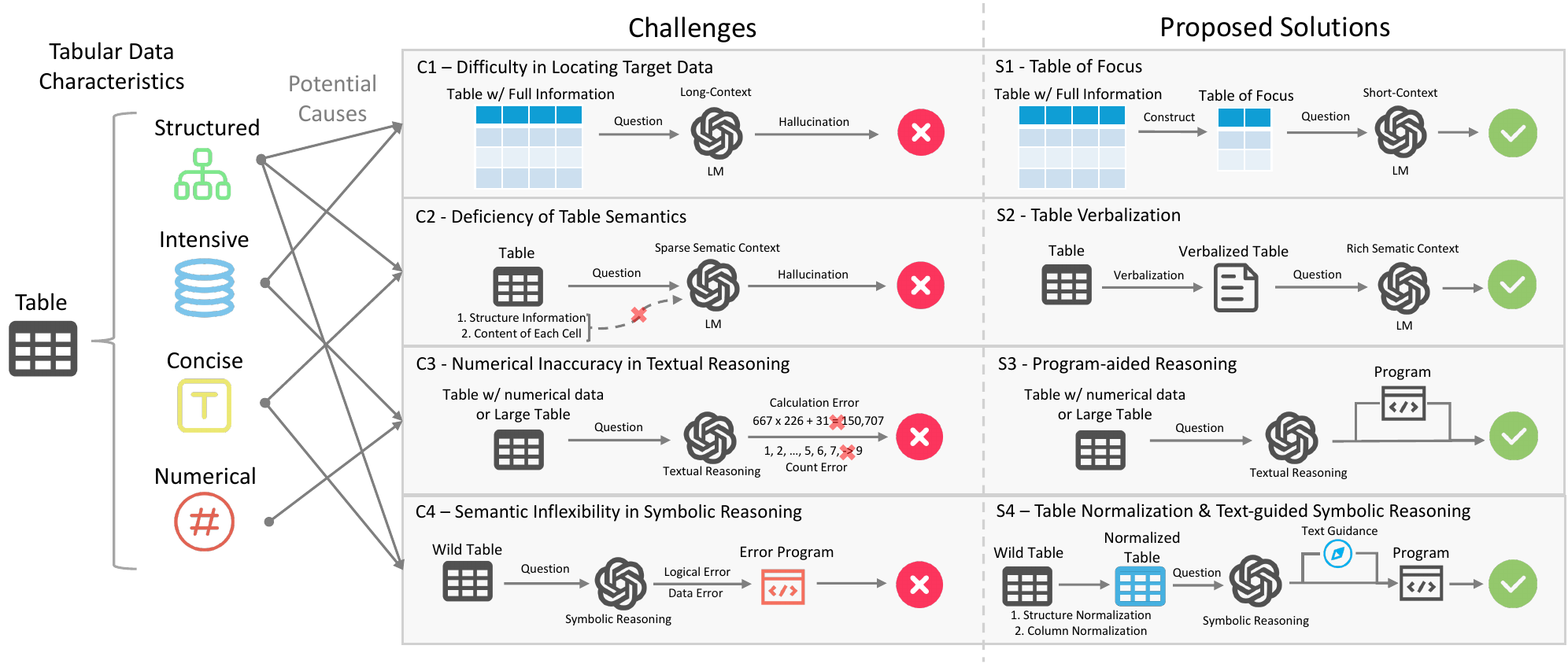}
    \caption{Overview of the challenges and proposed solutions in this work. Tabular data is inherently structured, dense, concise, and numerical. Based on these characteristics, we identify four key challenges. To address them, we propose four targeted solutions. The gray arrows between the characteristics and challenges represent the potential causes of these challenges stemming from specific characteristics. Each proposed solution corresponds to the challenge presented on the left in the same row. \method is a unified recipe developed based on these findings.}
    \label{fig:overview}
    \vspace{-10pt}
\end{figure}



Many previous studies have aimed to improve the table understanding capabilities of LMs. One efficient approach is using prompting to adapt LMs for table understanding without requiring fine-tuning, making it applicable to any advanced LM. Recent studies primarily adopt two main strategies to enhance table understanding with LMs. The first strategy involves extracting a sub-table that contains relevant content from the original table to reduce the context size, thereby making it easier for LMs to comprehend. Examples include Dater \citep{ye2023largelanguagemodelsversatile} and Chain-of-Table \citep{wang2024chainoftableevolvingtablesreasoning}, among others. The second strategy leverages SQL or Python programs to augment numerical reasoning, locate target data, and enhance table understanding of numerical information, as demonstrated by Binder \citep{cheng2023bindinglanguagemodelssymbolic} and LEVER \citep{ni2023leverlearningverifylanguagetocode}, etc. However, these studies primarily focus on a single basic aspect to enhance the performance of LMs in table understanding or design complex methods with isolated strategies. There is currently an absence of work that provides a systematic and fundamental analysis of table understanding with language models and proposes comprehensive methods for its improvement.



In this paper, we first provide extensive experiments and discussions to identify the challenges in table understanding with language models. To address these challenges, we then introduce \textbf{\method}, a recipe and comprehensive framework that integrates multiple solutions to tackle these issues effectively. In summary, this paper makes the following key contributions:
\begin{itemize}[leftmargin=*, itemsep=0pt, labelsep=5pt, topsep=0pt]
    \item \textbf{Challenges of Table Understanding.} We observe that tabular data is inherently structured, dense, concise, and numerical. Through empirical analysis, we identify four challenges associated with LMs’ table understanding: difficulty in locating target data, deficiency of table semantics, numerical inaccuracies in textual reasoning, and semantic inflexibility in symbolic reasoning. (Section~\ref{sec:challenge})
    
    \item \textbf{A Recipe for Table Understanding.} To address these challenges, we propose targeted solutions: table-of-focus, table verbalization, program-aided reasoning, table normalization, and text-guided symbolic reasoning. Building on these solutions, we introduce a framework as a unified recipe, \textbf{\method}. It also incorporates Adaptive Reasoning (AR), a flexible approach that dynamically adjusts between textual and symbolic reasoning, tailoring the reasoning process to each query. (Section~\ref{sec:recipe})
    
    \item \textbf{Extensive Experiments and Detailed Analyses.} We conduct extensive experiments and provide in-depth analyses to support our findings on table understanding with language models. Furthermore, we evaluate and demonstrate the superior performance of \method across three widely used table understanding datasets: WikiTQ, TabFact, and FetaQA. Notably, on the WikiTQ dataset, \method achieves an accuracy of 78.13\% based on GPT-4o-mini, surpassing existing baselines. (Section~\ref{sec:challenge}, Section~\ref{sec:exp}, and Appendix)
\end{itemize}

\section{Related Work}

\noindent\textbf{Reasoning with Language Models.} It has been observed that language models (LMs) can exhibit reasoning abilities when they are sufficiently large \citep{wei2022emergentabilitieslargelanguage, suzgun2022challengingbigbenchtaskschainofthought}. LMs are now widely used for various reasoning tasks, such as question answering \citep{kamalloo-etal-2023-evaluating}, decision making \citep{yang2023foundationmodelsdecisionmaking}, and mathematical reasoning \citep{ahn-etal-2024-large}. At the inference stage, techniques such as chain-of-thought prompting \citep{wei2023chainofthoughtpromptingelicitsreasoning} are used to trigger step-by-step reasoning processes and improve reasoning performance. Few-shot prompting \citep{brown2020languagemodelsfewshotlearners}, least-to-most prompting \citep{zhou2023leasttomostpromptingenablescomplex}, and program-of-thought prompting \citep{chen2023programthoughtspromptingdisentangling} have proven effective in specific scenarios. Methods like self-consistency \citep{wang2023selfconsistencyimproveschainthought} and structuring the reasoning process in forms like trees \citep{yao2023treethoughtsdeliberateproblem} or graphs \citep{Besta_2024, cao-2024-graphreason} are also useful for more complex reasoning tasks. Recently, many works have focused on using reinforcement learning \citep{lightman2023letsverifystepstep, uesato2022solvingmathwordproblems} to improve the reasoning abilities of LMs during training. Our work focuses on inference-time improvements and proposes a general framework applicable to all kinds of LMs for table understanding and reasoning.

\noindent\textbf{Fine-Tuning LMs for Table Understanding.} Several studies have focused on fine-tuning language models to enhance their understanding of tabular data. For example, based on the masked language modeling approach introduced in BERT \citep{devlin2019bertpretrainingdeepbidirectional}, models like TaPas \citep{herzig2020tapas}, Pasta \citep{gu2022pastatableoperationsawarefact}, and TUTA \citep{wang2021tuta} propose specialized pre-training methods to improve LMs’ ability to process tables. Similarly, TAPEX \citep{liu2022tapextablepretraininglearning} pre-trains an encoder-decoder model to function as a SQL executor, enabling better table comprehension. Recent advancements, such as TableLlama \citep{zhang2024tablellamaopenlargegeneralist}, TableGPT \citep{zha2023tablegptunifyingtablesnature}, and StructLLM \citep{zhuang2024structlmbuildinggeneralistmodels}, leverage open-sourced decoder-only models like Llama \citep{touvron2023llamaopenefficientfoundation} to pre-train larger models optimized for various downstream table-related tasks. Formula Tuning (Fortune) \citep{cao2025fortune} is a reinforcement learning approach that enables language models to perform symbolic table reasoning by deriving executable spreadsheet formulas.

\noindent\textbf{Adapting LMs for Table Understanding Without Fine-Tuning.} Other studies focus on adapting LMs to table-related tasks without requiring fine-tuning. For instance, Binder \citep{cheng2023bindinglanguagemodelssymbolic}, LEVER \citep{ni2023leverlearningverifylanguagetocode}, and PoTable \citep{mao2024potableprogrammingstandardlytablebased} generate SQL or Python programs, extending the capabilities of LMs to analyze tabular data. Dater \citep{ye2023largelanguagemodelsversatile}, TabSQLify \citep{nahid-rafiei-2024-tabsqlify}, ReAcTable \citep{zhang2023reactableenhancingreacttable}, TAP4LLM \citep{sui-etal-2024-tap4llm}, and Tree-of-Table \citep{ji2024treeoftableunleashingpowerllms} introduce different methods to construct sub-tables, modifying the tabular context for improved understanding. Chain-of-Table \citep{wang2024chainoftableevolvingtablesreasoning} generalizes various table operations, dynamically generating reasoning chains to create sub-tables. MIX-SC \citep{liu-etal-2024-rethinking} employs table normalization and leverages self-consistency, combining results from Python agents and textual reasoning to enhance performance. SpreadsheetEncoder \citep{dong-etal-2024-encoding} is specifically designed to interpret tabular data within spreadsheet environments. Our work also follows this direction to focus on adapting LMs without fine-tuning. We identify key challenges in table understanding and address them through our proposed method, which can be applied to any advanced LMs.


\section{Challenges in Table Understanding}
\label{sec:challenge}

As illustrated in Figure~\ref{fig:overview}, we identify and analyze the challenges in table understanding with language models (LMs) through the experiments shown in Figure~\ref{fig:challenge} and related discussions. Additionally, we propose targeted solutions to address these challenges. The detailed settings of the challenge analysis experiment are provided in Appendix~\ref{ap:analysis_settings}.

\noindent\textbf{Tabular Characteristics.} Tabular data differs from regular text, which is typically linear and sequential, due to its \textbf{structured} nature. Although tabular data can be represented as sequential text, it is fundamentally a two-dimensional array of cells. Each cell primarily contains text, but the cells are interconnected and share relationships with one another. Typically, cells within the same column represent the same feature or type, while cells in the same row correspond to a single data instance. Tables are highly efficient for data representation, often containing a large amount of information, making them inherently data-\textbf{intensive}. Moreover, the text in tables is typically \textbf{concise}, consisting of simple words and phrases rather than continuous sentences, leading to sparse semantic context. Lastly, tables frequently include substantial amounts of \textbf{numerical} data, such as dates, times, scores, and measurements, which often require specialized processing.

\subsection{Difficulty in Locating Target Data}

\begin{figure*}[t]
    \centering
    \includegraphics[width=\textwidth]{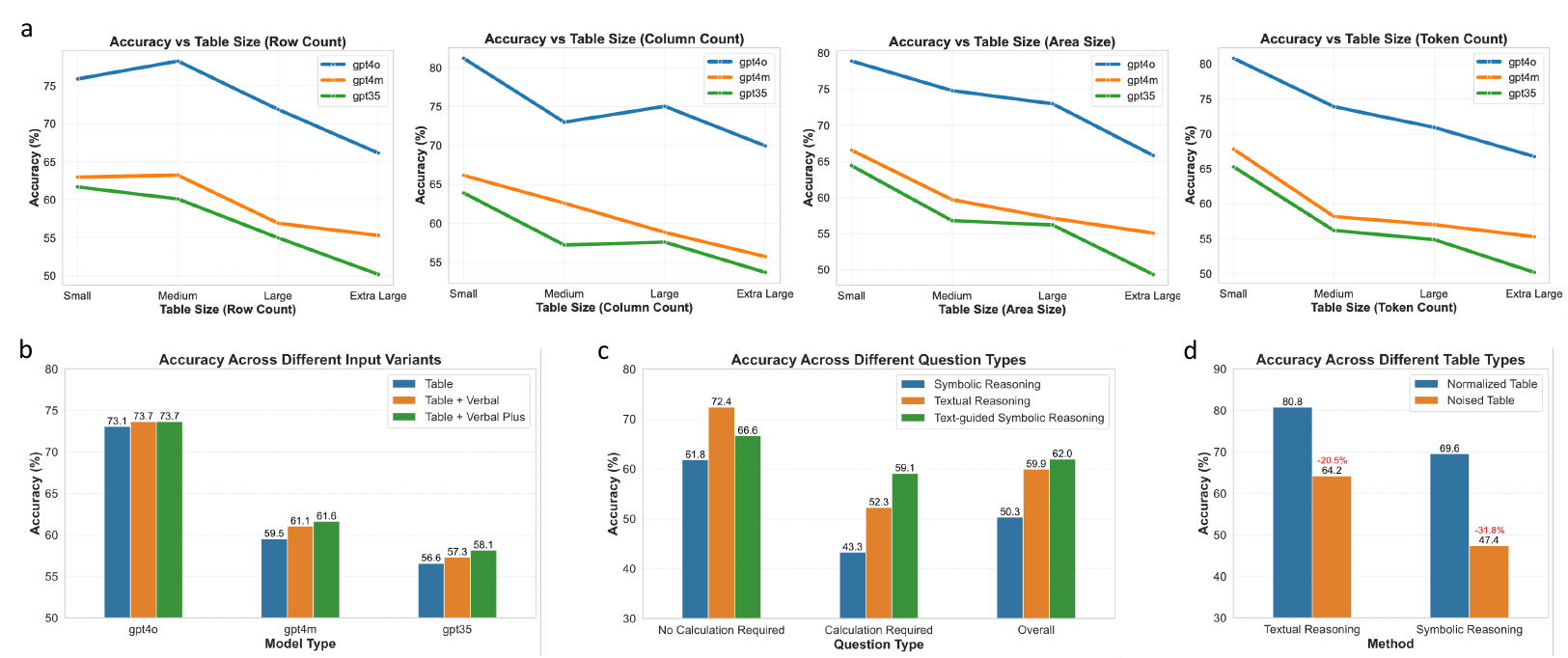}
    \caption{Experimental analysis of challenges in table understanding with language models. (a) Impact of table size on task difficulty. (b) Effect of verbalized tables with enriched semantic context. (c) Performance comparison of different reasoning methods on calculation-required versus non-calculation questions. (d) Performance differences when processing normalized versus noisy tables.}
    \label{fig:challenge}
\end{figure*}

When LMs encounter tabular data, they often struggle to locate the target data relevant to a given query, leading to misunderstandings. This challenge arises because tabular data is inherently data-intensive, typically containing large volumes of information. Additionally, the structured nature of tabular data makes it challenging for LMs to interpret individual cell contents within the broader context of headers and other structural information. This issue can lead to long-context hallucination \citep{huang2024hallucination}. Moreover, LMs are prone to neglecting information in the middle of the context \citep{liu-etal-2024-lost}, making it even harder to locate target data and further impairing their overall comprehension of the table. (Figure~\ref{fig:overview} - C1)

As shown in Figure~\ref{fig:challenge}(a), we present the changes in table understanding accuracy across four different table size metrics: row count, column count, area size, and token count, ranging from small to extra-large tables. Row count represents the number of data entries, while column count reflects the number of dimensions or attributes per entry. Area size is the product of row count and column count, and token count refers to table sizes from the perspective of LMs. All figures indicate that, regardless of the model used, overall performance tends to decline as table size increases. For weaker LMs, the performance drop is more pronounced.

To address this, we propose let LMs focusing on specific parts of the table by explicitly constructing a focused sub-table that includes only the relevant information needed for the given context. We define this as the \textbf{table-of-focus}. By narrowing the scope, table understanding becomes significantly easier, which aligns with both our previous findings and intuition. (Figure~\ref{fig:overview} - S1)

\subsection{Table Semantic Deficiency}
Tabular data is typically concise, with most cells containing simple words or phrases. Additionally, for each data entry in a row, some descriptive information may reside outside the row, such as in the top header or other structural elements. Understanding a cell in isolation is challenging and often requires a deeper comprehension of the structural relationships within the table. This leads to the problem of \textbf{sparse semantic context}, which is fundamentally different from the rich semantic context found in most data used during LMs’ pretraining \citep{dong2022tablepretrainingsurveymodel}. The semantic deficiency in tables makes it difficult for LMs to effectively understand and process tabular data. (Figure~\ref{fig:overview} - C2)

As shown in Figure~\ref{fig:challenge}(b), the \textit{Table} represents the case where the LM is provided only with the table input, while the \textit{Table+Verbal} indicates the table along with an additional description, which we refer to as a \textbf{verbalized table}. This description is generated by the LMs themselves, whereas \textit{verbal plus} refers to a description produced by more advanced LMs, which can be considered a ground-truth. We observe that verbalization helps LMs perform better on certain tables, leading to a slight overall performance improvement. This effect is more pronounced in weaker LMs, resulting in a 1.5\% increase in accuracy. Additionally, the quality of the description plays a crucial role in improvement.

To address this issue, we propose a solution where tables are first verbalized into sequential, natural text as a description and then provided to LMs alongside the original table before they directly tackle table-related tasks. It is similar to table2text \citep{parikh2020tottocontrolledtabletotextgeneration}. This transformation enriches the semantic context, making the data more aligned with the LMs’ pretraining, thereby enhancing their ability to effectively understand and process tabular data. (Figure~\ref{fig:overview} - S2)

\subsection{Numerical Inaccuracy in Textual Reasoning}
Tabular data often contains numerical values, such as dates, times, scores, and other recorded numbers, and is typically intensive. However, when LMs are used to process numerical data in textual reasoning, they often face significant limitations. LMs are prone to arithmetic calculation errors, especially when dealing with large numbers. LMs are also inefficient at handling iterative processes, particularly when the number of iteration steps is large \citep{chen2023programthoughtspromptingdisentangling}. (Figure~\ref{fig:overview} - C3)

As shown in Figure~\ref{fig:challenge}(c), questions that do not require calculations are relatively easier, allowing textual reasoning to achieve a strong performance of 72.4\%. However, when calculations are required, performance drops significantly, falling below that of the enhanced symbolic reasoning introduced later. Specifically, textual reasoning suffers a 20.1\% decline, whereas enhanced symbolic reasoning experiences a more moderate drop of 7.6\%.

Symbolic methods offer a promising solution to these challenges and have been explored extensively in prior research \citep{cheng2023bindinglanguagemodelssymbolic, ni2023leverlearningverifylanguagetocode, mao2024potableprogrammingstandardlytablebased}. Using symbolic tools, such as SQL or Python programs in combination with LMs, provides an effective approach to handling numerical data in tabular formats. (Figure~\ref{fig:overview} - S3)

\subsection{Semantic Inflexibility in Symbolic Reasoning}
Symbolic methods excel at arithmetic calculations. However, when prompting LMs to generate code for program of thought reasoning, the performance is suboptimal. Instead of truly understanding the context and generating problem-solving code, LMs often rely on memorized code from the pretraining stage \citep{yang-etal-2024-llms}. We refer to this limitation as \textbf{semantic inflexibility}. In table understanding, this challenge is exacerbated by the table’s complex structure and concise text content. In real-world scenarios, noisy tables further hinder LMs’ symbolic reasoning capabilities. Consequently, while symbolic reasoning with numerical data is highly accurate, the generated code may be incorrect due to issues in program logic or data handling, leading to errors or unintended results. (Figure~\ref{fig:overview} - C4)

As shown in Figure~\ref{fig:challenge}(c), basic symbolic reasoning performs worse overall, regardless of whether calculations are required. It indicates that basic symbolic reasoning with current LMs is ineffective. Furthermore, as illustrated in Figure~\ref{fig:challenge}(d), when processing the same content in a noisy format, symbolic reasoning suffers a larger performance drop of 31.8\%, compared to a 20.5\% decline for textual reasoning. This highlights the semantic inflexibility of symbolic reasoning when handling noisy tables.

To address this, we first normalize the table structure and content, ensuring that each column follows a consistent format. We then propose a solution where LMs first engage in textual reasoning before generating symbolic reasoning programs. This preliminary textual reasoning step serves as a guide for subsequent symbolic reasoning, improving alignment with the task context. Our approach can be seen as encouraging LMs to think more thoroughly before reasoning, aligning with techniques like plan-and-solve \citep{wang-etal-2023-plan}. By incorporating textual reasoning as a foundation, we enhance the accuracy and contextual relevance of symbolic reasoning. As demonstrated in Figure~\ref{fig:challenge}(c), this method achieves a higher accuracy of 59.1\% for calculation-required questions. (Figure~\ref{fig:overview} - S4)

\section{\method: A Recipe for Table Understanding}
\label{sec:recipe}

\begin{figure}[t]
    \centering
    \includegraphics[width=\textwidth]{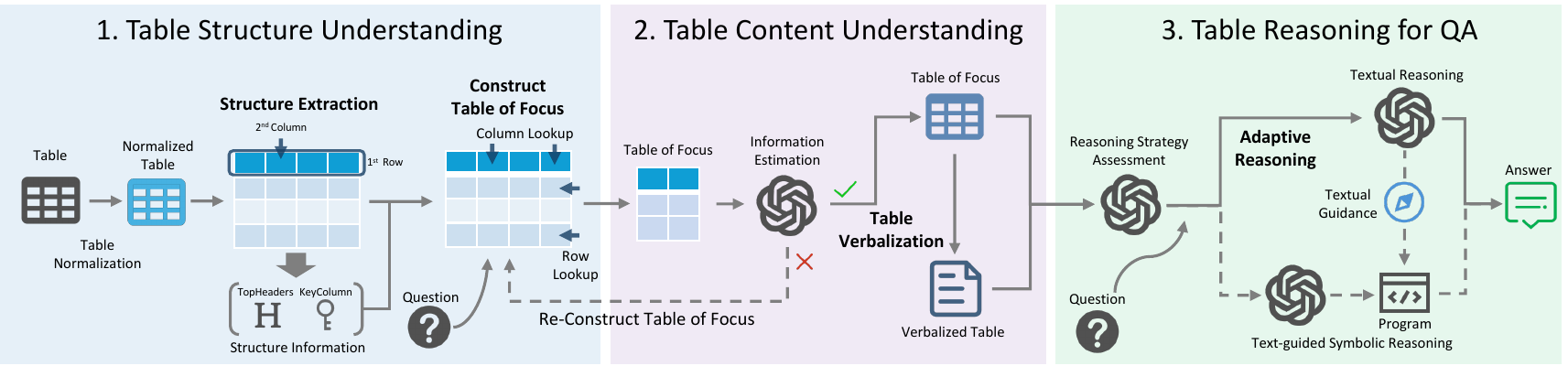}
    \caption{The framework of \method. It comprises three stages: (1) table structure understanding, where the table’s structure is analyzed, and a table-of-focus is constructed through row and column lookup; (2) table content understanding, where the table-of-focus is reconstructed based on the question, and its information is verbalized to enhance the semantic context; and (3) table reasoning for question answering, where an adaptive reasoning strategy determines whether to use textual reasoning or text-guided symbolic reasoning to derive the final answer. The dashed arrows indicate optional workflows, such as the table-of-focus re-construction and incorporating text-guided symbolic reasoning.}
    \label{fig:framework}
\end{figure}

Based on findings in Section~\ref{sec:challenge}, we introduce a recipe and comprehensive framework, \method, as shown in Figure~\ref{fig:framework}. It integrates the propose solution proposed in Section~\ref{sec:challenge} into a unified recipe for table understanding. The framework encompasses three key processes: Table Structure Understanding, Table Content Understanding, and Table Reasoning for QA. All notations are list at Appendix~\ref{ap:notation}.

\subsection{Task Formulation}
In table understanding, the objective is to determine an answer $A$ given a table $\mathbb{T}$ and a question or statement $Q$ related to it. The table $T$ is represented as a two-dimensional array of cells,
\[
\mathbb{T}_{m\times n} = \begin{bmatrix} 
C_{1,1} & C_{1,2} & \dots \\ 
C_{2,1} & C_{i,j} & \dots \\ 
\vdots & \vdots & \ddots 
\end{bmatrix}
\]
, where $C_{i,j}$ denotes the cell in the $i$-th row and $j$-th column, with the table consisting of $m$ rows and $n$ columns. In table-based question answering tasks, $Q$ represents a question, and $A$ is the expected answer in natural language. In table-based fact verification tasks, $Q$ is a statement about the table’s contents, and $A \in \{\text{True, False}\}$ is a Boolean value indicating whether the statement is correct. Therefore, the goal is to develop a system $\mathcal{F}$ that can predict the answer accurately based on the table and the given question or statement, formalized as $\mathcal{F}(\mathbb{T}, Q) = A$.

\subsection{Table Structure Understanding}
\label{sec:method_structure}

The goal of table structure understanding is to analyze the table’s structure and construct a Table-of-Focus that contains relevant content for the given question. This process reduces context length and simplifies the table as much as possible.

To enhance the efficiency of the framework, we introduce the \textbf{table peek} technique. For structure extraction and certain operations, it is often unnecessary to process the entire table; instead, inspecting only the top rows is sufficient. Given a peek size $k$, the original table $\mathbb{T}_{m\times n}$ is transformed into a peek table $\mathbb{T}_{k\times n}$, where all columns are retained, but the table is truncated to the first $k$ rows.

Given a wild table $\mathbb{T}^{W}$, we first normalize it. We begin by determining whether the table is in row-major or column-major format. If it is in column-major format, we transpose it using $\mathbb{T}=\text{Transpose}(\mathbb{T}')$. Next, we normalize and clean all columns containing numerical information, ensuring consistency in formats such as dates and numerical values, making them directly processable in bulk by a program. After this normalization process, we obtain the normalized table $\mathbb{T}^{N}$.

We begin by extracting the top headers $H$ and the key column. The top headers are used for column lookup, while the key column serves as the subject or unique identifier for each row. Next, we prompt LMs to perform column lookup and row lookup to identify the relevant rows and columns required for the task. Specifically, for column lookup, we first define the set of candidate columns as $\mathbb{C} = \text{Rank}(H)$. LMs will also rank all candidates based on their relevance to the question. We then prompt the LMs to select $b$ relevant columns based on a given question $Q$:
\[
C^0 = \text{Column Lookup}(\mathbb{T}^{N} \mid Q),
\]
where $C^0 = \{ c_i \mid c_i \in H \}$ and $|C^0| = b$. For row lookup, we instruct the LMs to generate an SQL query to efficiently filter and select $a$ relevant rows $R$:
\[
R = \text{Row Lookup}(\mathbb{T}^{N} \mid Q).
\]
Using the identified rows and columns, we construct the initial table-of-focus:
\[
\mathbb{T}^{F}_{a\times b} = \text{Table Construction}(\mathbb{T}^{N}, C^0, R),
\]
which contains only the filtered information necessary for the task.

\subsection{Table Content Understanding}
The goal of table content understanding is to enrich the semantic context of the table.

Studies have shown that LMs can assess whether sufficient information is available to answer a question \citep{cao-2024-learn, yin-etal-2023-large}. We first prompt the LMs to estimate whether the constructed Table-of-Focus  $\mathbb{T}^{F}_{a\times b}$, containing $C^0$, provides enough information to answer the given question $Q$. If not, additional column attributes from the candidate column set $\mathbf{C}$ are incrementally added from the ranked candidate headers until sufficient information is available or all relevant top headers have been utilized. Subsequently, a total of $a'$ columns from $C$ are selected for further reasoning. We use re-construction to mitigate information loss during the table-of-focus construction process. The detailed re-construction algorithm can be found in Appendix~\ref{ap:re-construction}.

Once the information sufficiency check is passed, we verbalize the table into natural language, adding descriptions to enrich the semantic context and producing a verbalized table:
\[
T^{\mathbb{T}} = \text{Verbalization}(\mathbb{T}^F_{a \times b}).
\]
This verbalized table is represented as sequential natural language text $T$ essentially rather than a structured table, preserving rich semantic context while maintaining a concise size. This transformation enhances information density, further facilitating the LMs’ reasoning for the given question.

\subsection{Table Reasoning for Question Answering}
The goal of this stage is to answer table-related questions by understanding the table precisely and calculating accurately.

We employ an \textbf{adaptive reasoning} approach. First, we prompt the LMs to determine the most appropriate reasoning strategy $S$ for the given task. In the instruction, for small tables or those without numerical data, the LMs are allowed to perform textual reasoning directly to derive the final result. For larger tables or those containing numerical data, symbolic reasoning with programmatic execution is selected.
\[
S = \text{Strategy Assessment}(\mathbb{T}^F, T^{\mathbb{T}}, Q),
\]
where $S \in \{\mathcal{T}, \mathcal{S}\}$ represents the chosen reasoning strategy, with $\mathcal{T}$  denoting textual reasoning and $\mathcal{S}$ denoting symbolic reasoning.

In symbolic reasoning, we first prompt the LMs to perform textual reasoning to generate guidance $G$ without providing the final result. This intermediate reasoning step is then used as input for symbolic reasoning, transitioning to a text-guided symbolic reasoning approach using programmatic methods. This adaptive method dynamically adjusts based on the table’s size, complexity, and the nature of the question, ensuring accurate and reliable results.
\[
A =
\begin{cases}
\text{Chain-of-Thought}(\mathbb{T}^F, T^{\mathbb{T}}, Q), & \text{if } S = \mathcal{T} \\
\mathcal{P}(\text{Program-of-Thought}(\mathbb{T}^F, T^{\mathbb{T}}, Q \mid G)), & \text{if } S = \mathcal{S}
\end{cases}
\]
where chain-of-thought and program-of-thought are two prompting techniques, $\mathcal{P}$ represents a Python or SQL program executor, $A$ is the final answer for the current table understanding task.

\section{Experiments}
\label{sec:exp}

\begin{table*}[t]
\centering
\caption{Performance comparison between \method and previous work on WikiTQ and TabFact. The values in the table represent accuracy (\%). The best result is \textbf{bold}, the second-best result is \underline{underlined}, and the improvement over the previous best result is highlighted in {\color{langgreen}green}. `-' indicates that the result values were not reported in the related papers. For all models in the table, results are obtained from a single inference run without any voting. Our method outperforms all other methods across both datasets and different language models.}
\resizebox{\textwidth}{!}{
\begin{tabular}{@{}l|ccc|ccc@{}}
\toprule[1pt]
\multirow{2}{*}{\textbf{Method}} 
& \multicolumn{3}{c|}{\textbf{WikiTQ}} 
& \multicolumn{3}{c}{\textbf{TabFact}} \\
\cmidrule(lr){2-4} \cmidrule(lr){5-7}
& \textbf{gpt-3.5-turbo\textsubscript{$\sim$175B}} & \textbf{gpt-4o-mini\textsubscript{$\sim$8B}} & \textbf{Llama-3.1\textsubscript{70B}}
& \textbf{gpt-3.5-turbo\textsubscript{$\sim$175B}} & \textbf{gpt-4o-mini\textsubscript{$\sim$8B}} & \textbf{Llama-3.1\textsubscript{70B}} \\
\midrule
Text-to-SQL \cite{rajkumar2022evaluatingtexttosqlcapabilitieslarge}     & 52.90   & -      & -      & 64.71 & -      & -      \\
End-to-End QA \cite{wang2024chainoftableevolvingtablesreasoning}   & 51.84  & -      & -      & 70.45 & -      & -      \\
Few-Shot QA \cite{wang2024chainoftableevolvingtablesreasoning}     & 52.56  & -      & -      & 71.54 & -      & -      \\
Chain-of-Thought \cite{wang2024chainoftableevolvingtablesreasoning} & 53.48  & -      & -      & 65.37 & -      & -      \\
ReAcTable \cite{zhang2023reactableenhancingreacttable}        & 52.50   & -      & -      & 74.40  & -      & -      \\
Binder \cite{cheng2023bindinglanguagemodelssymbolic}           & 56.74  & 58.86  & 50.51  & 79.17 & 84.63  & 78.16  \\
Dater \cite{ye2023largelanguagemodelsversatile}            & 52.81  & 58.33  & 43.53  & 78.01 & 80.98  & 81.57  \\
TabSQLify \cite{nahid-rafiei-2024-tabsqlify}       & \underline{64.70}  
                 & 57.02  
                 & 55.78  
                 & 79.50
                 & 78.75  
                 & 70.70   \\
Chain-of-Table \cite{wang2024chainoftableevolvingtablesreasoning}   & 59.94  & 55.60   & 62.22  & 80.20  & 84.24  & 85.62  \\
Tree-of-Table \cite{ji2024treeoftableunleashingpowerllms}    & 61.11  & -      & -      & \underline{81.92} & -      & -      \\
PoTable \cite{mao2024potableprogrammingstandardlytablebased}          & -      & \underline{64.73} & \underline{65.56} 
                 & -      & \underline{88.93} & \underline{87.06} \\
\midrule
\textbf{Ours (\method)} 
& \textbf{68.21 \textcolor{langgreen}{(+3.51)}} 
& \textbf{78.13 \textcolor{langgreen}{(+13.40)}} 
& \textbf{77.95 \textcolor{langgreen}{(+12.39)}} 
& \textbf{83.65 \textcolor{langgreen}{(+1.73)}} 
& \textbf{90.12 \textcolor{langgreen}{(+1.19)}} 
& \textbf{91.16 \textcolor{langgreen}{(+4.10)}} \\


\bottomrule[1pt]
\end{tabular}%
}
\label{tab:main_exp}
\end{table*}

\subsection{Settings}
We conduct extensive experiments to evaluate the performance of \method. Specifically, we assess its effectiveness across three different table understanding datasets: WikiTQ \citep{pasupat-liang-2015-compositional} (table-based question answering), TabFact \citep{chen2020tabfactlargescaledatasettablebased} (table-based fact verification), and FetaQA \citep{nan-etal-2022-fetaqa} (table-based free-form question answering). For WikiTQ and TabFact, following previous work \citep{wang2024chainoftableevolvingtablesreasoning, liu-etal-2024-rethinking}, we use exact match accuracy as the evaluation metric. For FetaQA, we evaluate performance using BLEU \citep{papineni-etal-2002-bleu} and ROUGE \citep{lin-2004-rouge} scores. We also conduct experiments on HiTab \citep{cheng-etal-2022-hitab} and FinQA \citep{chen2021finqa}. Tables are encoded in Markdown format before being input into language models, with or without addresses, depending on the specific case \ref{ap:case_study_tm}.

Our experiments utilize OpenAI models hosted on Microsoft Azure. Unless otherwise stated, we set the temperature to 0 to ensure stable output while keeping all other hyperparameters at their default values. The models used in our evaluation include \textit{gpt-4o} (\textit{gpt-4o-0806}), \textit{gpt-4o-mini} (\textit{gpt-4o-mini-0718}), \textit{gpt-3.5-turbo} (\textit{gpt-3.5-turbo-0125}), \textit{o1} (\textit{o1-preview-0912}), and \textit{o1-mini} (\textit{o1-mini-0912}). Additionally, we evaluate our methods on open-sourced \textit{Llama-3.1-70B} (\textit{Llama-3.1-70B-Instruct}).
For comparison, we select several strong baselines, including both classic and state-of-the-art methods such as Binder \citep{cheng2023bindinglanguagemodelssymbolic}, Dater \citep{ye2023largelanguagemodelsversatile}, and Chain-of-Table \citep{wang2024chainoftableevolvingtablesreasoning}. Performance results for other methods not in this work are cited directly from their original or related papers, with sources indicated alongside the method names in the results table.

Further analysis and additional experiments on \method can be found in the Appendix. The prompts used in \method can be found in Appendix~\ref{ap:tm_prompt}, while other prompts used in this work are provided in Appendix~\ref{ap:o_prompt}.


\subsection{Main Results}

As shown in Table~\ref{tab:main_exp}, our \method approach consistently achieves the highest performance across both WikiTQ and TabFact under different backbone models (\textit{gpt-3.5-turbo}, \textit{gpt-4o-mini}, and \textit{Llama-3.1-70B}). On WikiTQ, \method outperforms the strongest baselines by +3.51, +13.40, and +12.39 points, respectively. A similar trend is observed on TabFact, with improvements of +1.73, +1.19, and +4.10 points, demonstrating the robustness of our method across diverse large language models. Results on the FetaQA, HiTab, FinQA dataset are provided in Appendix~\ref{ap:fetaqa}, \ref{ap:hitab}, \ref{ap:finqa}. These results confirm that \method not only generalizes well across different base language models but also significantly enhances table understanding and reasoning in complex QA tasks.

Notably, methods such as Binder, Dater, TabSQLify, and Chain-of-Table exhibit subpar performance with \textit{gpt-4o-mini}, in some cases performing worse than with \textit{gpt-3.5-turbo}. Our empirical analysis suggests that these methods primarily rely on symbolic approaches to construct subtables, which often fail to leverage the strengths of chain-of-thought reasoning in textual contexts. This limitation underscores the necessity of integrating advanced textual reasoning strategies, as effectively demonstrated by our \method approach.

\subsection{Ablation Study}

To analyze the contribution of each component in \method, we conduct an ablation study on WikiTQ and TabFact. Table~\ref{tab:ablation} presents the results, and the performance drop from the full model is highlighted in red. The results demonstrate that removing any component leads to a decrease in accuracy, confirming the importance of each module in the overall framework.

\begin{wraptable}{r}{0.6\textwidth} 
\centering
\caption{Ablation results on WikiTQ and TabFact. The values in the table represent accuracy (\%), with $\triangledown$ indicating the performance drop. The {\color{langred}red} text highlights the drop magnitude. Removing any component from \method results in a decrease in performance.}
\label{tab:ablation}
\resizebox{0.6\textwidth}{!}{
\begin{tabular}{l|cc|cc}
\toprule
\textbf{Method} & \textbf{WikiTQ} & $\triangledown$ & \textbf{TabFact} & $\triangledown$ \\
\midrule
\textbf{\method  (gpt-4o-mini)} & \textbf{78.13} & -- & \textbf{90.12} & -- \\
\midrule
\multicolumn{5}{l}{\textbf{Structure}} \\
\quad w/o Structure Extraction & 74.75 & \textcolor{langred}{(-3.38)} & 88.98 & \textcolor{langred}{(-1.14)} \\
\quad w/o Column Lookup       & 77.00 & \textcolor{langred}{(-1.13)} & 90.51 & \textcolor{langred}{(-0.40)} \\
\quad w/o Row Lookup          & 76.59 & \textcolor{langred}{(-1.54)} & 89.23 & \textcolor{langred}{(-0.89)} \\
\quad w/o Table of Focus      & 76.40 & \textcolor{langred}{(-1.73)} & 89.33 & \textcolor{langred}{(-0.79)} \\
\midrule
\multicolumn{5}{l}{\textbf{Content}} \\
\quad w/o Re-Construction     & 75.55 & \textcolor{langred}{(-2.58)} & 89.72 & \textcolor{langred}{(-0.40)} \\
\quad w/o Verbalization       & 75.78 & \textcolor{langred}{(-2.35)} & 89.23 & \textcolor{langred}{(-0.89)} \\
\midrule
\multicolumn{5}{l}{\textbf{Reasoning}} \\
\quad w/o Textual Reasoning   & 73.85 & \textcolor{langred}{(-4.28)} & 88.39 & \textcolor{langred}{(-1.73)} \\
\quad w/o Symbolic Reasoning  & 76.10 & \textcolor{langred}{(-2.03)} & 89.18 & \textcolor{langred}{(-0.94)} \\
\quad w/o Textual Guidance    & 75.21 & \textcolor{langred}{(-2.92)} & 89.67 & \textcolor{langred}{(-0.44)} \\
\bottomrule
\end{tabular}
}
\end{wraptable}

\noindent{\textbf{Structure}}. The structure understanding components play a crucial role in table comprehension. Removing structure extraction results in a notable accuracy drop of 3.38\% on WikiTQ and 1.14\% on TabFact, indicating that explicitly extracting the table’s structure is essential for effective reasoning, as failing to do so can lead to errors in subsequent steps. Among lookup strategies, removing row lookup leads to a 1.54\% decrease in WikiTQ accuracy, whereas removing column lookup results in a smaller drop of 1.13\%. This suggests that row-based information retrieval is more critical than column-based lookup, as large tables typically contain a greater number of rows. Additionally, removing the table-of-focus reduces performance by 1.73\% on WikiTQ and 0.79\% on TabFact, further emphasizing its important role in structuring relevant table content to extract key information for reasoning.

\noindent{\textbf{Content}}. Table content understanding also significantly influences performance. Eliminating re-construction, which iteratively refines the Table-of-Focus based on the question, results in a 2.58\% accuracy drop on WikiTQ and 0.40\% on TabFact, highlighting the importance of this process. Similarly, removing table verbalization, which enriches the semantic context of the table by adding descriptive elements, leads to a 2.35\% decrease in WikiTQ accuracy. However, its impact on TabFact is minimal (0.23\% drop), suggesting that verbalization becomes even more beneficial for complex table understanding tasks.

\noindent{\textbf{Reasoning}}. The reasoning stage exhibits the most significant performance drop when removed. Removing textual reasoning leads to the largest accuracy decline, with a 4.28\% drop on WikiTQ and 1.73\% on TabFact, underscoring its necessity for complex reasoning tasks. Similarly, removing symbolic reasoning results in a 2.03\% and 0.79\% drop on WikiTQ and TabFact, respectively, demonstrating that symbolic reasoning enhances numerical and structured table interpretations. Finally, removing textual guidance, which improves the semantic flexibility of symbolic reasoning, reduces accuracy by 2.92\% on WikiTQ and 0.44\% on TabFact. This highlights that textual guidance is particularly beneficial and important in symbolic reasoning by ensuring alignment with the problem context. More analysis of adaptive reasoning can be found at Appendix~\ref{ap:adaptive}.


\section{Conclusion}
In this paper, we explore table understanding with language models. Given the characteristics of tabular data, we identify key challenges in table understanding. To overcome these challenges, we propose \method, a recipe and comprehensive framework that integrates multiple solutions. Extensive analyses and experiments demonstrate our findings and the effectiveness of \method. In the future, we plan to extend and refine the framework to improve its performance across diverse practical applications, where discussed in Appendix~\ref{ap:limit}.


\section*{Acknowledgments}
This work was completed during my internship in the gap year prior to starting my PhD and was motivated by core product needs within the company. I sincerely thank the organization for the resources provided, as well as all the friends and colleagues who supported me during this period.


\bibliography{iclr2026_conference}

@article{ghasemi2016process,
  title={Process mining in healthcare: a systematised literature review},
  author={Ghasemi, Mahdi and Amyot, Daniel},
  journal={International Journal of Electronic Healthcare},
  volume={9},
  number={1},
  pages={60--88},
  year={2016},
  publisher={Inderscience Publishers (IEL)}
}

@misc{li2020gftegraphbasedfinancialtable,
      title={GFTE: Graph-based Financial Table Extraction}, 
      author={Yiren Li and Zheng Huang and Junchi Yan and Yi Zhou and Fan Ye and Xianhui Liu},
      year={2020},
      eprint={2003.07560},
      archivePrefix={arXiv},
      primaryClass={cs.CV},
      url={https://arxiv.org/abs/2003.07560}, 
}

@misc{gunasekar2023textbooksneed,
      title={Textbooks Are All You Need}, 
      author={Suriya Gunasekar and Yi Zhang and Jyoti Aneja and Caio César Teodoro Mendes and Allie Del Giorno and Sivakanth Gopi and Mojan Javaheripi and Piero Kauffmann and Gustavo de Rosa and Olli Saarikivi and Adil Salim and Shital Shah and Harkirat Singh Behl and Xin Wang and Sébastien Bubeck and Ronen Eldan and Adam Tauman Kalai and Yin Tat Lee and Yuanzhi Li},
      year={2023},
      eprint={2306.11644},
      archivePrefix={arXiv},
      primaryClass={cs.CL},
      url={https://arxiv.org/abs/2306.11644}, 
}

@misc{openai2024gpt4technicalreport,
      title={GPT-4 Technical Report}, 
      author={OpenAI},
      year={2024},
      eprint={2303.08774},
      archivePrefix={arXiv},
      primaryClass={cs.CL},
      url={https://arxiv.org/abs/2303.08774}, 
}

@misc{plaat2024reasoninglargelanguagemodels,
      title={Reasoning with Large Language Models, a Survey}, 
      author={Aske Plaat and Annie Wong and Suzan Verberne and Joost Broekens and Niki van Stein and Thomas Back},
      year={2024},
      eprint={2407.11511},
      archivePrefix={arXiv},
      primaryClass={cs.AI},
      url={https://arxiv.org/abs/2407.11511}, 
}

@misc{minaee2024largelanguagemodelssurvey,
      title={Large Language Models: A Survey}, 
      author={Shervin Minaee and Tomas Mikolov and Narjes Nikzad and Meysam Chenaghlu and Richard Socher and Xavier Amatriain and Jianfeng Gao},
      year={2024},
      eprint={2402.06196},
      archivePrefix={arXiv},
      primaryClass={cs.CL},
      url={https://arxiv.org/abs/2402.06196}, 
}

@inproceedings{zhu-etal-2024-large,
    title = "Can Large Language Models Understand Context?",
    author = "Zhu, Yilun  and
      Moniz, Joel Ruben Antony  and
      Bhargava, Shruti  and
      Lu, Jiarui  and
      Piraviperumal, Dhivya  and
      Li, Site  and
      Zhang, Yuan  and
      Yu, Hong  and
      Tseng, Bo-Hsiang",
    editor = "Graham, Yvette  and
      Purver, Matthew",
    booktitle = "Findings of the Association for Computational Linguistics: EACL 2024",
    month = mar,
    year = "2024",
    address = "St. Julian{'}s, Malta",
    publisher = "Association for Computational Linguistics",
    url = "https://aclanthology.org/2024.findings-eacl.135",
    pages = "2004--2018"
}

@misc{fang2024largelanguagemodelsllmstabular,
      title={Large Language Models(LLMs) on Tabular Data: Prediction, Generation, and Understanding -- A Survey}, 
      author={Xi Fang and Weijie Xu and Fiona Anting Tan and Jiani Zhang and Ziqing Hu and Yanjun Qi and Scott Nickleach and Diego Socolinsky and Srinivasan Sengamedu and Christos Faloutsos},
      year={2024},
      eprint={2402.17944},
      archivePrefix={arXiv},
      primaryClass={cs.CL},
      url={https://arxiv.org/abs/2402.17944}, 
}

@misc{zhang2024surveytablereasoninglarge,
      title={A Survey of Table Reasoning with Large Language Models}, 
      author={Xuanliang Zhang and Dingzirui Wang and Longxu Dou and Qingfu Zhu and Wanxiang Che},
      year={2024},
      eprint={2402.08259},
      archivePrefix={arXiv},
      primaryClass={cs.CL},
      url={https://arxiv.org/abs/2402.08259}, 
}

@inproceedings{chen-2023-large,
    title = "Large Language Models are few(1)-shot Table Reasoners",
    author = "Chen, Wenhu",
    editor = "Vlachos, Andreas  and
      Augenstein, Isabelle",
    booktitle = "Findings of the Association for Computational Linguistics: EACL 2023",
    month = may,
    year = "2023",
    address = "Dubrovnik, Croatia",
    publisher = "Association for Computational Linguistics",
    url = "https://aclanthology.org/2023.findings-eacl.83",
    doi = "10.18653/v1/2023.findings-eacl.83",
    pages = "1120--1130"
}

@misc{wei2022emergentabilitieslargelanguage,
      title={Emergent Abilities of Large Language Models}, 
      author={Jason Wei and Yi Tay and Rishi Bommasani and Colin Raffel and Barret Zoph and Sebastian Borgeaud and Dani Yogatama and Maarten Bosma and Denny Zhou and Donald Metzler and Ed H. Chi and Tatsunori Hashimoto and Oriol Vinyals and Percy Liang and Jeff Dean and William Fedus},
      year={2022},
      eprint={2206.07682},
      archivePrefix={arXiv},
      primaryClass={cs.CL},
      url={https://arxiv.org/abs/2206.07682}, 
}

@misc{suzgun2022challengingbigbenchtaskschainofthought,
      title={Challenging BIG-Bench Tasks and Whether Chain-of-Thought Can Solve Them}, 
      author={Mirac Suzgun and Nathan Scales and Nathanael Schärli and Sebastian Gehrmann and Yi Tay and Hyung Won Chung and Aakanksha Chowdhery and Quoc V. Le and Ed H. Chi and Denny Zhou and Jason Wei},
      year={2022},
      eprint={2210.09261},
      archivePrefix={arXiv},
      primaryClass={cs.CL},
      url={https://arxiv.org/abs/2210.09261}, 
}

@article{huang2024hallucination,
   title={A Survey on Hallucination in Large Language Models: Principles, Taxonomy, Challenges, and Open Questions},
   ISSN={1558-2868},
   url={http://dx.doi.org/10.1145/3703155},
   DOI={10.1145/3703155},
   journal={ACM Transactions on Information Systems},
   publisher={Association for Computing Machinery (ACM)},
   author={Huang, Lei and Yu, Weijiang and Ma, Weitao and Zhong, Weihong and Feng, Zhangyin and Wang, Haotian and Chen, Qianglong and Peng, Weihua and Feng, Xiaocheng and Qin, Bing and Liu, Ting},
   year={2024},
   month={nov}
}

@inproceedings{cao-2024-learn,
    title = "Learn to Refuse: Making Large Language Models More Controllable and Reliable through Knowledge Scope Limitation and Refusal Mechanism",
    author = "Cao, Lang",
    editor = "Al-Onaizan, Yaser  and
      Bansal, Mohit  and
      Chen, Yun-Nung",
    booktitle = "Proceedings of the 2024 Conference on Empirical Methods in Natural Language Processing",
    month = nov,
    year = "2024",
    address = "Miami, Florida, USA",
    publisher = "Association for Computational Linguistics",
    url = "https://aclanthology.org/2024.emnlp-main.212",
    doi = "10.18653/v1/2024.emnlp-main.212",
    pages = "3628--3646"
}

@inproceedings{yin-etal-2023-large,
    title = "Do Large Language Models Know What They Don{'}t Know?",
    author = "Yin, Zhangyue  and
      Sun, Qiushi  and
      Guo, Qipeng  and
      Wu, Jiawen  and
      Qiu, Xipeng  and
      Huang, Xuanjing",
    editor = "Rogers, Anna  and
      Boyd-Graber, Jordan  and
      Okazaki, Naoaki",
    booktitle = "Findings of the Association for Computational Linguistics: ACL 2023",
    month = jul,
    year = "2023",
    address = "Toronto, Canada",
    publisher = "Association for Computational Linguistics",
    url = "https://aclanthology.org/2023.findings-acl.551",
    doi = "10.18653/v1/2023.findings-acl.551",
    pages = "8653--8665"
}

@article{liu-etal-2024-lost,
    title = "Lost in the Middle: How Language Models Use Long Contexts",
    author = "Liu, Nelson F.  and
      Lin, Kevin  and
      Hewitt, John  and
      Paranjape, Ashwin  and
      Bevilacqua, Michele  and
      Petroni, Fabio  and
      Liang, Percy",
    journal = "Transactions of the Association for Computational Linguistics",
    volume = "12",
    year = "2024",
    address = "Cambridge, MA",
    publisher = "MIT Press",
    url = "https://aclanthology.org/2024.tacl-1.9",
    doi = "10.1162/tacl_a_00638",
    pages = "157--173"
}

@inproceedings{wang-etal-2023-plan,
    title = "Plan-and-Solve Prompting: Improving Zero-Shot Chain-of-Thought Reasoning by Large Language Models",
    author = "Wang, Lei  and
      Xu, Wanyu  and
      Lan, Yihuai  and
      Hu, Zhiqiang  and
      Lan, Yunshi  and
      Lee, Roy Ka-Wei  and
      Lim, Ee-Peng",
    editor = "Rogers, Anna  and
      Boyd-Graber, Jordan  and
      Okazaki, Naoaki",
    booktitle = "Proceedings of the 61st Annual Meeting of the Association for Computational Linguistics (Volume 1: Long Papers)",
    month = jul,
    year = "2023",
    address = "Toronto, Canada",
    publisher = "Association for Computational Linguistics",
    url = "https://aclanthology.org/2023.acl-long.147",
    doi = "10.18653/v1/2023.acl-long.147",
    pages = "2609--2634"
}

@inproceedings{kamalloo-etal-2023-evaluating,
    title = "Evaluating Open-Domain Question Answering in the Era of Large Language Models",
    author = "Kamalloo, Ehsan  and
      Dziri, Nouha  and
      Clarke, Charles  and
      Rafiei, Davood",
    editor = "Rogers, Anna  and
      Boyd-Graber, Jordan  and
      Okazaki, Naoaki",
    booktitle = "Proceedings of the 61st Annual Meeting of the Association for Computational Linguistics (Volume 1: Long Papers)",
    month = jul,
    year = "2023",
    address = "Toronto, Canada",
    publisher = "Association for Computational Linguistics",
    url = "https://aclanthology.org/2023.acl-long.307",
    doi = "10.18653/v1/2023.acl-long.307",
    pages = "5591--5606",
}

@misc{yang2023foundationmodelsdecisionmaking,
      title={Foundation Models for Decision Making: Problems, Methods, and Opportunities}, 
      author={Sherry Yang and Ofir Nachum and Yilun Du and Jason Wei and Pieter Abbeel and Dale Schuurmans},
      year={2023},
      eprint={2303.04129},
      archivePrefix={arXiv},
      primaryClass={cs.AI},
      url={https://arxiv.org/abs/2303.04129}, 
}

@inproceedings{ahn-etal-2024-large,
    title = "Large Language Models for Mathematical Reasoning: Progresses and Challenges",
    author = "Ahn, Janice  and
      Verma, Rishu  and
      Lou, Renze  and
      Liu, Di  and
      Zhang, Rui  and
      Yin, Wenpeng",
    editor = "Falk, Neele  and
      Papi, Sara  and
      Zhang, Mike",
    booktitle = "Proceedings of the 18th Conference of the European Chapter of the Association for Computational Linguistics: Student Research Workshop",
    month = mar,
    year = "2024",
    address = "St. Julian{'}s, Malta",
    publisher = "Association for Computational Linguistics",
    url = "https://aclanthology.org/2024.eacl-srw.17",
    pages = "225--237",
}

@misc{wei2023chainofthoughtpromptingelicitsreasoning,
      title={Chain-of-Thought Prompting Elicits Reasoning in Large Language Models}, 
      author={Jason Wei and Xuezhi Wang and Dale Schuurmans and Maarten Bosma and Brian Ichter and Fei Xia and Ed Chi and Quoc Le and Denny Zhou},
      year={2023},
      eprint={2201.11903},
      archivePrefix={arXiv},
      primaryClass={cs.CL},
      url={https://arxiv.org/abs/2201.11903}, 
}

@misc{brown2020languagemodelsfewshotlearners,
      title={Language Models are Few-Shot Learners}, 
      author={Tom B. Brown and Benjamin Mann and Nick Ryder and Melanie Subbiah and Jared Kaplan and Prafulla Dhariwal and Arvind Neelakantan and Pranav Shyam and Girish Sastry and Amanda Askell and Sandhini Agarwal and Ariel Herbert-Voss and Gretchen Krueger and Tom Henighan and Rewon Child and Aditya Ramesh and Daniel M. Ziegler and Jeffrey Wu and Clemens Winter and Christopher Hesse and Mark Chen and Eric Sigler and Mateusz Litwin and Scott Gray and Benjamin Chess and Jack Clark and Christopher Berner and Sam McCandlish and Alec Radford and Ilya Sutskever and Dario Amodei},
      year={2020},
      eprint={2005.14165},
      archivePrefix={arXiv},
      primaryClass={cs.CL},
      url={https://arxiv.org/abs/2005.14165}, 
}

@misc{zhou2023leasttomostpromptingenablescomplex,
      title={Least-to-Most Prompting Enables Complex Reasoning in Large Language Models}, 
      author={Denny Zhou and Nathanael Schärli and Le Hou and Jason Wei and Nathan Scales and Xuezhi Wang and Dale Schuurmans and Claire Cui and Olivier Bousquet and Quoc Le and Ed Chi},
      year={2023},
      eprint={2205.10625},
      archivePrefix={arXiv},
      primaryClass={cs.AI},
      url={https://arxiv.org/abs/2205.10625}, 
}

@misc{chen2023programthoughtspromptingdisentangling,
      title={Program of Thoughts Prompting: Disentangling Computation from Reasoning for Numerical Reasoning Tasks}, 
      author={Wenhu Chen and Xueguang Ma and Xinyi Wang and William W. Cohen},
      year={2023},
      eprint={2211.12588},
      archivePrefix={arXiv},
      primaryClass={cs.CL},
      url={https://arxiv.org/abs/2211.12588}, 
}

@misc{wang2023selfconsistencyimproveschainthought,
      title={Self-Consistency Improves Chain of Thought Reasoning in Language Models}, 
      author={Xuezhi Wang and Jason Wei and Dale Schuurmans and Quoc Le and Ed Chi and Sharan Narang and Aakanksha Chowdhery and Denny Zhou},
      year={2023},
      eprint={2203.11171},
      archivePrefix={arXiv},
      primaryClass={cs.CL},
      url={https://arxiv.org/abs/2203.11171}, 
}

@misc{yao2023treethoughtsdeliberateproblem,
      title={Tree of Thoughts: Deliberate Problem Solving with Large Language Models}, 
      author={Shunyu Yao and Dian Yu and Jeffrey Zhao and Izhak Shafran and Thomas L. Griffiths and Yuan Cao and Karthik Narasimhan},
      year={2023},
      eprint={2305.10601},
      archivePrefix={arXiv},
      primaryClass={cs.CL},
      url={https://arxiv.org/abs/2305.10601}, 
}

@article{Besta_2024,
   title={Graph of Thoughts: Solving Elaborate Problems with Large Language Models},
   volume={38},
   ISSN={2159-5399},
   url={http://dx.doi.org/10.1609/aaai.v38i16.29720},
   DOI={10.1609/aaai.v38i16.29720},
   number={16},
   journal={Proceedings of the AAAI Conference on Artificial Intelligence},
   publisher={Association for the Advancement of Artificial Intelligence (AAAI)},
   author={Besta, Maciej and Blach, Nils and Kubicek, Ales and Gerstenberger, Robert and Podstawski, Michal and Gianinazzi, Lukas and Gajda, Joanna and Lehmann, Tomasz and Niewiadomski, Hubert and Nyczyk, Piotr and Hoefler, Torsten},
   year={2024},
   month=mar, pages={17682–17690}
}

@inproceedings{cao-2024-graphreason,
    title = "{G}raph{R}eason: Enhancing Reasoning Capabilities of Large Language Models through A Graph-Based Verification Approach",
    author = "Cao, Lang",
    editor = "Dalvi Mishra, Bhavana  and
      Durrett, Greg  and
      Jansen, Peter  and
      Lipkin, Ben  and
      Neves Ribeiro, Danilo  and
      Wong, Lionel  and
      Ye, Xi  and
      Zhao, Wenting",
    booktitle = "Proceedings of the 2nd Workshop on Natural Language Reasoning and Structured Explanations (@ACL 2024)",
    month = aug,
    year = "2024",
    address = "Bangkok, Thailand",
    publisher = "Association for Computational Linguistics",
    url = "https://aclanthology.org/2024.nlrse-1.1",
    pages = "1--12",
}

@misc{uesato2022solvingmathwordproblems,
      title={Solving math word problems with process- and outcome-based feedback}, 
      author={Jonathan Uesato and Nate Kushman and Ramana Kumar and Francis Song and Noah Siegel and Lisa Wang and Antonia Creswell and Geoffrey Irving and Irina Higgins},
      year={2022},
      eprint={2211.14275},
      archivePrefix={arXiv},
      primaryClass={cs.LG},
      url={https://arxiv.org/abs/2211.14275}, 
}

@misc{lightman2023letsverifystepstep,
      title={Let's Verify Step by Step}, 
      author={Hunter Lightman and Vineet Kosaraju and Yura Burda and Harri Edwards and Bowen Baker and Teddy Lee and Jan Leike and John Schulman and Ilya Sutskever and Karl Cobbe},
      year={2023},
      eprint={2305.20050},
      archivePrefix={arXiv},
      primaryClass={cs.LG},
      url={https://arxiv.org/abs/2305.20050}, 
}

@inproceedings{yang-etal-2024-llms,
    title = "Can {LLM}s Reason in the Wild with Programs?",
    author = "Yang, Yuan  and
      Xiong, Siheng  and
      Payani, Ali  and
      Shareghi, Ehsan  and
      Fekri, Faramarz",
    editor = "Al-Onaizan, Yaser  and
      Bansal, Mohit  and
      Chen, Yun-Nung",
    booktitle = "Findings of the Association for Computational Linguistics: EMNLP 2024",
    month = nov,
    year = "2024",
    address = "Miami, Florida, USA",
    publisher = "Association for Computational Linguistics",
    url = "https://aclanthology.org/2024.findings-emnlp.573/",
    doi = "10.18653/v1/2024.findings-emnlp.573",
    pages = "9806--9829"
}

@inproceedings{pasupat-liang-2015-compositional,
    title = "Compositional Semantic Parsing on Semi-Structured Tables",
    author = "Pasupat, Panupong  and
      Liang, Percy",
    editor = "Zong, Chengqing  and
      Strube, Michael",
    booktitle = "Proceedings of the 53rd Annual Meeting of the Association for Computational Linguistics and the 7th International Joint Conference on Natural Language Processing (Volume 1: Long Papers)",
    month = jul,
    year = "2015",
    address = "Beijing, China",
    publisher = "Association for Computational Linguistics",
    url = "https://aclanthology.org/P15-1142",
    doi = "10.3115/v1/P15-1142",
    pages = "1470--1480",
}

@misc{chen2020tabfactlargescaledatasettablebased,
      title={TabFact: A Large-scale Dataset for Table-based Fact Verification}, 
      author={Wenhu Chen and Hongmin Wang and Jianshu Chen and Yunkai Zhang and Hong Wang and Shiyang Li and Xiyou Zhou and William Yang Wang},
      year={2020},
      eprint={1909.02164},
      archivePrefix={arXiv},
      primaryClass={cs.CL},
      url={https://arxiv.org/abs/1909.02164}, 
}

@misc{wang2024chainoftableevolvingtablesreasoning,
      title={Chain-of-Table: Evolving Tables in the Reasoning Chain for Table Understanding}, 
      author={Zilong Wang and Hao Zhang and Chun-Liang Li and Julian Martin Eisenschlos and Vincent Perot and Zifeng Wang and Lesly Miculicich and Yasuhisa Fujii and Jingbo Shang and Chen-Yu Lee and Tomas Pfister},
      year={2024},
      eprint={2401.04398},
      archivePrefix={arXiv},
      primaryClass={cs.CL},
      url={https://arxiv.org/abs/2401.04398}, 
}

@misc{devlin2019bertpretrainingdeepbidirectional,
      title={BERT: Pre-training of Deep Bidirectional Transformers for Language Understanding}, 
      author={Jacob Devlin and Ming-Wei Chang and Kenton Lee and Kristina Toutanova},
      year={2019},
      eprint={1810.04805},
      archivePrefix={arXiv},
      primaryClass={cs.CL},
      url={https://arxiv.org/abs/1810.04805}, 
}

@article{herzig2020tapas,
  author={Jonathan Herzig and Pawel Krzysztof Nowak and Thomas M{\"{u}}ller and rancesco Piccinno and Julian Martin Eisenschlos},
  title={{TAPAS:} Weakly Supervised Table Parsing via Pre-training},
  journal= {CoRR},
  volume={abs/2004.02349},
  year={2020},
  url={https://arxiv.org/abs/2004.02349},
  eprinttype={arXiv},
  eprint={2004.02349},
  bibsource={dblp computer science bibliography, https://dblp.org}
}

@misc{gu2022pastatableoperationsawarefact,
      title={PASTA: Table-Operations Aware Fact Verification via Sentence-Table Cloze Pre-training}, 
      author={Zihui Gu and Ju Fan and Nan Tang and Preslav Nakov and Xiaoman Zhao and Xiaoyong Du},
      year={2022},
      eprint={2211.02816},
      archivePrefix={arXiv},
      primaryClass={cs.CL},
      url={https://arxiv.org/abs/2211.02816}, 
}

@inproceedings{wang2021tuta,
  title={Tuta: Tree-based transformers for generally structured table pre-training},
  author={Wang, Zhiruo and Dong, Haoyu and Jia, Ran and Li, Jia and Fu, Zhiyi and Han, Shi and Zhang, Dongmei},
  booktitle={Proceedings of the 27th ACM SIGKDD Conference on Knowledge Discovery \& Data Mining},
  pages={1780--1790},
  year={2021}
}

@misc{liu2022tapextablepretraininglearning,
      title={TAPEX: Table Pre-training via Learning a Neural SQL Executor}, 
      author={Qian Liu and Bei Chen and Jiaqi Guo and Morteza Ziyadi and Zeqi Lin and Weizhu Chen and Jian-Guang Lou},
      year={2022},
      eprint={2107.07653},
      archivePrefix={arXiv},
      primaryClass={cs.CL},
      url={https://arxiv.org/abs/2107.07653}, 
}

@misc{zhang2024tablellamaopenlargegeneralist,
      title={TableLlama: Towards Open Large Generalist Models for Tables}, 
      author={Tianshu Zhang and Xiang Yue and Yifei Li and Huan Sun},
      year={2024},
      eprint={2311.09206},
      archivePrefix={arXiv},
      primaryClass={cs.CL},
      url={https://arxiv.org/abs/2311.09206}, 
}

@misc{zha2023tablegptunifyingtablesnature,
      title={TableGPT: Towards Unifying Tables, Nature Language and Commands into One GPT}, 
      author={Liangyu Zha and Junlin Zhou and Liyao Li and Rui Wang and Qingyi Huang and Saisai Yang and Jing Yuan and Changbao Su and Xiang Li and Aofeng Su and Tao Zhang and Chen Zhou and Kaizhe Shou and Miao Wang and Wufang Zhu and Guoshan Lu and Chao Ye and Yali Ye and Wentao Ye and Yiming Zhang and Xinglong Deng and Jie Xu and Haobo Wang and Gang Chen and Junbo Zhao},
      year={2023},
      eprint={2307.08674},
      archivePrefix={arXiv},
      primaryClass={cs.AI},
      url={https://arxiv.org/abs/2307.08674}, 
}

@misc{zhuang2024structlmbuildinggeneralistmodels,
      title={StructLM: Towards Building Generalist Models for Structured Knowledge Grounding}, 
      author={Alex Zhuang and Ge Zhang and Tianyu Zheng and Xinrun Du and Junjie Wang and Weiming Ren and Stephen W. Huang and Jie Fu and Xiang Yue and Wenhu Chen},
      year={2024},
      eprint={2402.16671},
      archivePrefix={arXiv},
      primaryClass={cs.CL},
      url={https://arxiv.org/abs/2402.16671}, 
}

@misc{touvron2023llamaopenefficientfoundation,
      title={LLaMA: Open and Efficient Foundation Language Models}, 
      author={Hugo Touvron and Thibaut Lavril and Gautier Izacard and Xavier Martinet and Marie-Anne Lachaux and Timothée Lacroix and Baptiste Rozière and Naman Goyal and Eric Hambro and Faisal Azhar and Aurelien Rodriguez and Armand Joulin and Edouard Grave and Guillaume Lample},
      year={2023},
      eprint={2302.13971},
      archivePrefix={arXiv},
      primaryClass={cs.CL},
      url={https://arxiv.org/abs/2302.13971}, 
}

@misc{rajkumar2022evaluatingtexttosqlcapabilitieslarge,
      title={Evaluating the Text-to-SQL Capabilities of Large Language Models}, 
      author={Nitarshan Rajkumar and Raymond Li and Dzmitry Bahdanau},
      year={2022},
      eprint={2204.00498},
      archivePrefix={arXiv},
      primaryClass={cs.CL},
      url={https://arxiv.org/abs/2204.00498}, 
}

@misc{cheng2023bindinglanguagemodelssymbolic,
      title={Binding Language Models in Symbolic Languages}, 
      author={Zhoujun Cheng and Tianbao Xie and Peng Shi and Chengzu Li and Rahul Nadkarni and Yushi Hu and Caiming Xiong and Dragomir Radev and Mari Ostendorf and Luke Zettlemoyer and Noah A. Smith and Tao Yu},
      year={2023},
      eprint={2210.02875},
      archivePrefix={arXiv},
      primaryClass={cs.CL},
      url={https://arxiv.org/abs/2210.02875}, 
}

@misc{zhang2023reactableenhancingreacttable,
      title={ReAcTable: Enhancing ReAct for Table Question Answering}, 
      author={Yunjia Zhang and Jordan Henkel and Avrilia Floratou and Joyce Cahoon and Shaleen Deep and Jignesh M. Patel},
      year={2023},
      eprint={2310.00815},
      archivePrefix={arXiv},
      primaryClass={cs.DB},
      url={https://arxiv.org/abs/2310.00815}, 
}

@misc{ni2023leverlearningverifylanguagetocode,
      title={LEVER: Learning to Verify Language-to-Code Generation with Execution}, 
      author={Ansong Ni and Srini Iyer and Dragomir Radev and Ves Stoyanov and Wen-tau Yih and Sida I. Wang and Xi Victoria Lin},
      year={2023},
      eprint={2302.08468},
      archivePrefix={arXiv},
      primaryClass={cs.LG},
      url={https://arxiv.org/abs/2302.08468}, 
}

@misc{mao2024potableprogrammingstandardlytablebased,
      title={PoTable: Programming Standardly on Table-based Reasoning Like a Human Analyst}, 
      author={Qingyang Mao and Qi Liu and Zhi Li and Mingyue Cheng and Zheng Zhang and Rui Li},
      year={2024},
      eprint={2412.04272},
      archivePrefix={arXiv},
      primaryClass={cs.IR},
      url={https://arxiv.org/abs/2412.04272}, 
}

@misc{ye2023largelanguagemodelsversatile,
      title={Large Language Models are Versatile Decomposers: Decompose Evidence and Questions for Table-based Reasoning}, 
      author={Yunhu Ye and Binyuan Hui and Min Yang and Binhua Li and Fei Huang and Yongbin Li},
      year={2023},
      eprint={2301.13808},
      archivePrefix={arXiv},
      primaryClass={cs.CL},
      url={https://arxiv.org/abs/2301.13808}, 
}

@inproceedings{liu-etal-2024-rethinking,
    title = "Rethinking Tabular Data Understanding with Large Language Models",
    author = "Liu, Tianyang  and
      Wang, Fei  and
      Chen, Muhao",
    editor = "Duh, Kevin  and
      Gomez, Helena  and
      Bethard, Steven",
    booktitle = "Proceedings of the 2024 Conference of the North American Chapter of the Association for Computational Linguistics: Human Language Technologies (Volume 1: Long Papers)",
    month = jun,
    year = "2024",
    address = "Mexico City, Mexico",
    publisher = "Association for Computational Linguistics",
    url = "https://aclanthology.org/2024.naacl-long.26",
    doi = "10.18653/v1/2024.naacl-long.26",
    pages = "450--482",
}

@inproceedings{nahid-rafiei-2024-tabsqlify,
    title = "{T}ab{SQL}ify: Enhancing Reasoning Capabilities of {LLM}s Through Table Decomposition",
    author = "Nahid, Md  and
      Rafiei, Davood",
    editor = "Duh, Kevin  and
      Gomez, Helena  and
      Bethard, Steven",
    booktitle = "Proceedings of the 2024 Conference of the North American Chapter of the Association for Computational Linguistics: Human Language Technologies (Volume 1: Long Papers)",
    month = jun,
    year = "2024",
    address = "Mexico City, Mexico",
    publisher = "Association for Computational Linguistics",
    url = "https://aclanthology.org/2024.naacl-long.320",
    doi = "10.18653/v1/2024.naacl-long.320",
    pages = "5725--5737",
}

@inproceedings{sui-etal-2024-tap4llm,
    title = "{TAP}4{LLM}: Table Provider on Sampling, Augmenting, and Packing Semi-structured Data for Large Language Model Reasoning",
    author = "Sui, Yuan  and
      Zou, Jiaru  and
      Zhou, Mengyu  and
      He, Xinyi  and
      Du, Lun  and
      Han, Shi  and
      Zhang, Dongmei",
    editor = "Al-Onaizan, Yaser  and
      Bansal, Mohit  and
      Chen, Yun-Nung",
    booktitle = "Findings of the Association for Computational Linguistics: EMNLP 2024",
    month = nov,
    year = "2024",
    address = "Miami, Florida, USA",
    publisher = "Association for Computational Linguistics",
    url = "https://aclanthology.org/2024.findings-emnlp.603",
    doi = "10.18653/v1/2024.findings-emnlp.603",
    pages = "10306--10323"
}

@misc{ji2024treeoftableunleashingpowerllms,
      title={Tree-of-Table: Unleashing the Power of LLMs for Enhanced Large-Scale Table Understanding}, 
      author={Deyi Ji and Lanyun Zhu and Siqi Gao and Peng Xu and Hongtao Lu and Jieping Ye and Feng Zhao},
      year={2024},
      eprint={2411.08516},
      archivePrefix={arXiv},
      primaryClass={cs.CL},
      url={https://arxiv.org/abs/2411.08516}, 
}

@inproceedings{dong-etal-2024-encoding,
    title = "Encoding Spreadsheets for Large Language Models",
    author = "Dong, Haoyu  and
      Zhao, Jianbo  and
      Tian, Yuzhang  and
      Xiong, Junyu  and
      Zhou, Mengyu  and
      Lin, Yun  and
      Cambronero, Jos{\'e}  and
      He, Yeye  and
      Han, Shi  and
      Zhang, Dongmei",
    editor = "Al-Onaizan, Yaser  and
      Bansal, Mohit  and
      Chen, Yun-Nung",
    booktitle = "Proceedings of the 2024 Conference on Empirical Methods in Natural Language Processing",
    month = nov,
    year = "2024",
    address = "Miami, Florida, USA",
    publisher = "Association for Computational Linguistics",
    url = "https://aclanthology.org/2024.emnlp-main.1154",
    doi = "10.18653/v1/2024.emnlp-main.1154",
    pages = "20728--20748"
}

@misc{abacha2025medecbenchmarkmedicalerror,
      title={MEDEC: A Benchmark for Medical Error Detection and Correction in Clinical Notes}, 
      author={Asma Ben Abacha and Wen-wai Yim and Yujuan Fu and Zhaoyi Sun and Meliha Yetisgen and Fei Xia and Thomas Lin},
      year={2025},
      eprint={2412.19260},
      archivePrefix={arXiv},
      primaryClass={cs.CL},
      url={https://arxiv.org/abs/2412.19260}, 
}

@misc{chen2021evaluatinglargelanguagemodels,
      title={Evaluating Large Language Models Trained on Code}, 
      author={Mark Chen and Jerry Tworek and Heewoo Jun and Qiming Yuan and Henrique Ponde de Oliveira Pinto and Jared Kaplan and Harri Edwards and Yuri Burda and Nicholas Joseph and Greg Brockman and Alex Ray and Raul Puri and Gretchen Krueger and Michael Petrov and Heidy Khlaaf and Girish Sastry and Pamela Mishkin and Brooke Chan and Scott Gray and Nick Ryder and Mikhail Pavlov and Alethea Power and Lukasz Kaiser and Mohammad Bavarian and Clemens Winter and Philippe Tillet and Felipe Petroski Such and Dave Cummings and Matthias Plappert and Fotios Chantzis and Elizabeth Barnes and Ariel Herbert-Voss and William Hebgen Guss and Alex Nichol and Alex Paino and Nikolas Tezak and Jie Tang and Igor Babuschkin and Suchir Balaji and Shantanu Jain and William Saunders and Christopher Hesse and Andrew N. Carr and Jan Leike and Josh Achiam and Vedant Misra and Evan Morikawa and Alec Radford and Matthew Knight and Miles Brundage and Mira Murati and Katie Mayer and Peter Welinder and Bob McGrew and Dario Amodei and Sam McCandlish and Ilya Sutskever and Wojciech Zaremba},
      year={2021},
      eprint={2107.03374},
      archivePrefix={arXiv},
      primaryClass={cs.LG},
      url={https://arxiv.org/abs/2107.03374}, 
}

@misc{dong2022tablepretrainingsurveymodel,
      title={Table Pre-training: A Survey on Model Architectures, Pre-training Objectives, and Downstream Tasks}, 
      author={Haoyu Dong and Zhoujun Cheng and Xinyi He and Mengyu Zhou and Anda Zhou and Fan Zhou and Ao Liu and Shi Han and Dongmei Zhang},
      year={2022},
      eprint={2201.09745},
      archivePrefix={arXiv},
      primaryClass={cs.CL},
      url={https://arxiv.org/abs/2201.09745}, 
}

@article{nan-etal-2022-fetaqa,
    title = "{F}e{T}a{QA}: Free-form Table Question Answering",
    author = "Nan, Linyong  and
      Hsieh, Chiachun  and
      Mao, Ziming  and
      Lin, Xi Victoria  and
      Verma, Neha  and
      Zhang, Rui  and
      Kry{\'s}ci{\'n}ski, Wojciech  and
      Schoelkopf, Hailey  and
      Kong, Riley  and
      Tang, Xiangru  and
      Mutuma, Mutethia  and
      Rosand, Ben  and
      Trindade, Isabel  and
      Bandaru, Renusree  and
      Cunningham, Jacob  and
      Xiong, Caiming  and
      Radev, Dragomir  and
      Radev, Dragomir",
    editor = "Roark, Brian  and
      Nenkova, Ani",
    journal = "Transactions of the Association for Computational Linguistics",
    volume = "10",
    year = "2022",
    address = "Cambridge, MA",
    publisher = "MIT Press",
    url = "https://aclanthology.org/2022.tacl-1.3/",
    doi = "10.1162/tacl_a_00446",
    pages = "35--49",
}

@inproceedings{papineni-etal-2002-bleu,
    title = "{B}leu: a Method for Automatic Evaluation of Machine Translation",
    author = "Papineni, Kishore  and
      Roukos, Salim  and
      Ward, Todd  and
      Zhu, Wei-Jing",
    editor = "Isabelle, Pierre  and
      Charniak, Eugene  and
      Lin, Dekang",
    booktitle = "Proceedings of the 40th Annual Meeting of the Association for Computational Linguistics",
    month = jul,
    year = "2002",
    address = "Philadelphia, Pennsylvania, USA",
    publisher = "Association for Computational Linguistics",
    url = "https://aclanthology.org/P02-1040/",
    doi = "10.3115/1073083.1073135",
    pages = "311--318"
}

@inproceedings{lin-2004-rouge,
    title = "{ROUGE}: A Package for Automatic Evaluation of Summaries",
    author = "Lin, Chin-Yew",
    booktitle = "Text Summarization Branches Out",
    month = jul,
    year = "2004",
    address = "Barcelona, Spain",
    publisher = "Association for Computational Linguistics",
    url = "https://aclanthology.org/W04-1013/",
    pages = "74--81"
}

@inproceedings{nahid-rafiei-2024-normtab,
    title = "{N}orm{T}ab: Improving Symbolic Reasoning in {LLM}s Through Tabular Data Normalization",
    author = "Nahid, Md Mahadi Hasan  and
      Rafiei, Davood",
    editor = "Al-Onaizan, Yaser  and
      Bansal, Mohit  and
      Chen, Yun-Nung",
    booktitle = "Findings of the Association for Computational Linguistics: EMNLP 2024",
    month = nov,
    year = "2024",
    address = "Miami, Florida, USA",
    publisher = "Association for Computational Linguistics",
    url = "https://aclanthology.org/2024.findings-emnlp.203/",
    doi = "10.18653/v1/2024.findings-emnlp.203",
    pages = "3569--3585",
}

@misc{parikh2020tottocontrolledtabletotextgeneration,
      title={ToTTo: A Controlled Table-To-Text Generation Dataset}, 
      author={Ankur P. Parikh and Xuezhi Wang and Sebastian Gehrmann and Manaal Faruqui and Bhuwan Dhingra and Diyi Yang and Dipanjan Das},
      year={2020},
      eprint={2004.14373},
      archivePrefix={arXiv},
      primaryClass={cs.CL},
      url={https://arxiv.org/abs/2004.14373}, 
}

@misc{anil2023palm2technicalreport,
      title={PaLM 2 Technical Report}, 
      author={Rohan Anil and Andrew M. Dai and Orhan Firat and Melvin Johnson and Dmitry Lepikhin and Alexandre Passos and Siamak Shakeri and Emanuel Taropa and Paige Bailey and Zhifeng Chen and Eric Chu and Jonathan H. Clark and Laurent El Shafey and Yanping Huang and Kathy Meier-Hellstern and Gaurav Mishra and Erica Moreira and Mark Omernick and Kevin Robinson and Sebastian Ruder and Yi Tay and Kefan Xiao and Yuanzhong Xu and Yujing Zhang and Gustavo Hernandez Abrego and Junwhan Ahn and Jacob Austin and Paul Barham and Jan Botha and James Bradbury and Siddhartha Brahma and Kevin Brooks and Michele Catasta and Yong Cheng and Colin Cherry and Christopher A. Choquette-Choo and Aakanksha Chowdhery and Clément Crepy and Shachi Dave and Mostafa Dehghani and Sunipa Dev and Jacob Devlin and Mark Díaz and Nan Du and Ethan Dyer and Vlad Feinberg and Fangxiaoyu Feng and Vlad Fienber and Markus Freitag and Xavier Garcia and Sebastian Gehrmann and Lucas Gonzalez and Guy Gur-Ari and Steven Hand and Hadi Hashemi and Le Hou and Joshua Howland and Andrea Hu and Jeffrey Hui and Jeremy Hurwitz and Michael Isard and Abe Ittycheriah and Matthew Jagielski and Wenhao Jia and Kathleen Kenealy and Maxim Krikun and Sneha Kudugunta and Chang Lan and Katherine Lee and Benjamin Lee and Eric Li and Music Li and Wei Li and YaGuang Li and Jian Li and Hyeontaek Lim and Hanzhao Lin and Zhongtao Liu and Frederick Liu and Marcello Maggioni and Aroma Mahendru and Joshua Maynez and Vedant Misra and Maysam Moussalem and Zachary Nado and John Nham and Eric Ni and Andrew Nystrom and Alicia Parrish and Marie Pellat and Martin Polacek and Alex Polozov and Reiner Pope and Siyuan Qiao and Emily Reif and Bryan Richter and Parker Riley and Alex Castro Ros and Aurko Roy and Brennan Saeta and Rajkumar Samuel and Renee Shelby and Ambrose Slone and Daniel Smilkov and David R. So and Daniel Sohn and Simon Tokumine and Dasha Valter and Vijay Vasudevan and Kiran Vodrahalli and Xuezhi Wang and Pidong Wang and Zirui Wang and Tao Wang and John Wieting and Yuhuai Wu and Kelvin Xu and Yunhan Xu and Linting Xue and Pengcheng Yin and Jiahui Yu and Qiao Zhang and Steven Zheng and Ce Zheng and Weikang Zhou and Denny Zhou and Slav Petrov and Yonghui Wu},
      year={2023},
      eprint={2305.10403},
      archivePrefix={arXiv},
      primaryClass={cs.CL},
      url={https://arxiv.org/abs/2305.10403}, 
}

@inproceedings{maynez-etal-2023-benchmarking,
    title = "Benchmarking Large Language Model Capabilities for Conditional Generation",
    author = "Maynez, Joshua  and
      Agrawal, Priyanka  and
      Gehrmann, Sebastian",
    editor = "Rogers, Anna  and
      Boyd-Graber, Jordan  and
      Okazaki, Naoaki",
    booktitle = "Proceedings of the 61st Annual Meeting of the Association for Computational Linguistics (Volume 1: Long Papers)",
    month = jul,
    year = "2023",
    address = "Toronto, Canada",
    publisher = "Association for Computational Linguistics",
    url = "https://aclanthology.org/2023.acl-long.511/",
    doi = "10.18653/v1/2023.acl-long.511",
    pages = "9194--9213"
}

@inproceedings{cheng-etal-2022-hitab,
    title = "{H}i{T}ab: A Hierarchical Table Dataset for Question Answering and Natural Language Generation",
    author = "Cheng, Zhoujun  and
      Dong, Haoyu  and
      Wang, Zhiruo  and
      Jia, Ran  and
      Guo, Jiaqi  and
      Gao, Yan  and
      Han, Shi  and
      Lou, Jian-Guang  and
      Zhang, Dongmei",
    editor = "Muresan, Smaranda  and
      Nakov, Preslav  and
      Villavicencio, Aline",
    booktitle = "Proceedings of the 60th Annual Meeting of the Association for Computational Linguistics (Volume 1: Long Papers)",
    month = may,
    year = "2022",
    address = "Dublin, Ireland",
    publisher = "Association for Computational Linguistics",
    url = "https://aclanthology.org/2022.acl-long.78/",
    doi = "10.18653/v1/2022.acl-long.78",
    pages = "1094--1110"
}

@inproceedings{zhang-etal-2024-e5,
    title = "$E^5$: Zero-shot Hierarchical Table Analysis using Augmented {LLM}s via Explain, Extract, Execute, Exhibit and Extrapolate",
    author = "Zhang, Zhehao  and
      Gao, Yan  and
      Lou, Jian-Guang",
    editor = "Duh, Kevin  and
      Gomez, Helena  and
      Bethard, Steven",
    booktitle = "Proceedings of the 2024 Conference of the North American Chapter of the Association for Computational Linguistics: Human Language Technologies (Volume 1: Long Papers)",
    month = jun,
    year = "2024",
    address = "Mexico City, Mexico",
    publisher = "Association for Computational Linguistics",
    url = "https://aclanthology.org/2024.naacl-long.68/",
    doi = "10.18653/v1/2024.naacl-long.68",
    pages = "1244--1258"
}

@misc{MultiCoT2025,
  title        = {{MultiCoT}: Chain-of-Table Reasoning with Multiple Tables},
  author       = {{CYQIQ}},
  year         = {2025},
  howpublished = {\url{https://github.com/CYQIQ/MultiCoT}},
  note         = {GitHub repository},
  urldate      = {2025-05-02}
}

@article{chen2021finqa,
  title={FinQA: A Dataset of Numerical Reasoning over Financial Data},
  author={Chen, Zhiyu and Chen, Wenhu and Smiley, Charese and Shah, Sameena and Borova, Iana and Langdon, Dylan and Moussa, Reema and Beane, Matt and Huang, Ting-Hao and Routledge, Bryan and Wang, William Yang},
  journal={Proceedings of EMNLP 2021},
  year={2021}
}

@article{yi2025tablepilot,
    title = {TablePilot: Recommending Human-Preferred Tabular Data Analysis with Large Language Models}, 
    author = {Yi, Deyin and Liu, Yihao and Cao, Lang and Zhou, Mengyu and Dong, Haoyu and Han, Shi and Zhang, Dongmei},
    year = {2025},
    journal = {arXiv preprint arXiv: 2503.13262},
}

@misc{abhyankar2025hstarllmdrivenhybridsqltext,
      title={H-STAR: LLM-driven Hybrid SQL-Text Adaptive Reasoning on Tables}, 
      author={Nikhil Abhyankar and Vivek Gupta and Dan Roth and Chandan K. Reddy},
      year={2025},
      eprint={2407.05952},
      archivePrefix={arXiv},
      primaryClass={cs.DB},
      url={https://arxiv.org/abs/2407.05952}, 
}

@article{cao2025fortune,
    title = {Fortune: Formula-Driven Reinforcement Learning for Symbolic Table Reasoning in Language Models},
    author = {Cao, Lang and Xu, Jingxian and Liu, Hanbing and Wang, Jinyu and Zhou, Mengyu and Dong, Haoyu and Han, Shi and Zhang, Dongmei},
    journal = {arXiv preprint arXiv:2505.23667},
    year = {2025}
}
\bibliographystyle{iclr2026_conference}


\clearpage
\DoToC
\clearpage

\appendix

\section{Ethics Statement}
\label{ap:ethic}

\method introduces a general-purpose, modular framework that improves the ability of language models (LMs) to understand and reason over tabular data. Its applications span domains such as business intelligence, scientific reporting, education, and healthcare, where structured data plays a critical role. By enabling adaptive reasoning across textual and symbolic paradigms, \method improves both accuracy and transparency in question answering, verification, and analysis over tables. These advances may lead to better decision-support systems and streamlined human–AI collaboration in spreadsheet-heavy workflows.

Despite these benefits, there are potential risks. As \method automates reasoning over tabular content, it could be misused to generate misleading analyses or automate decisions without adequate human oversight. Furthermore, inaccurate reasoning, especially when symbolic operations are applied incorrectly, could result in flawed conclusions or financial misjudgments. Biases in training data might also manifest in generated answers or program logic. In addition, as the framework relies on LLMs’ capabilities, disparities across languages, domains, or spreadsheet conventions may lead to uneven performance, potentially disadvantaging users in low-resource settings.

To mitigate these concerns, we propose several safeguards. First, outputs involving symbolic reasoning should be verified via deterministic execution (e.g., code validation or unit tests) before downstream use. Second, we encourage model evaluation on a diverse range of real-world tables, including messy, hierarchical, or multilingual formats. Third, human oversight is recommended in high-stakes applications, especially when deployed in financial, legal, or healthcare settings. Fourth, interpretability tools—such as reasoning traces or program annotations—should be integrated to facilitate debugging and auditing. Finally, we advocate for transparent reporting of model limitations and publishing benchmark results across different domains and table types to promote responsible deployment.

\section{Limitations, Extendability, and Future Works}
\label{ap:limit}
Although we conduct extensive experiments and in-depth analysis, this work still has certain limitations. However, we believe that \method possesses extensibility, allowing for future refinements. These improvements may include technical refinement as well as optimizing its application in downstream applications.

\subsection{Technical Refinement}

\noindent{\textbf{Wild Table.}} In our experiments, the tables in the three datasets we use are already cleaned; therefore, we do not explicitly implement table normalization in our evaluation experiments. However, we conduct analysis experiments to highlight the importance of table normalization for handling wild tables. In practical scenarios, various tools are available for table normalization. Regular expression matching can be employed for formatting, and small language models can also be leveraged to efficiently process and normalize tables \citep{nahid-rafiei-2024-normtab}.

\noindent{\textbf{Hierarchy Table.}} In our work, we assume all tables are flat, allowing for straightforward utilization and extraction of structural information. However, many real-world tables are hierarchical, where data is organized in a tree structure, making table structure understanding more challenging. We envision two possible solutions: converting hierarchical tables into flat tables or designing a tree-based structure extraction method to effectively locate target data.

\noindent{\textbf{Table-of-Focus Construction.}} In designing the Table-of-Focus, we employ two efficient methods: LM prompting for column lookup and SQL generation for row lookup. The Table-of-Focus is then constructed based on the results of these two lookups. Many previous works \citep{ji2024treeoftableunleashingpowerllms, wang2024chainoftableevolvingtablesreasoning} have introduced complex approaches for extracting relevant sub-tables. In contrast, our method remains intentionally simple, prioritizing efficiency and adaptability. We believe that in the future, more advanced techniques may emerge to further enhance the extraction of key information.

\noindent{\textbf{Table Verbalization.}} To facilitate the implementation of \method, we utilize language models themselves to verbalize the table. However, the quality of the generated text is not optimal due to the challenges of open-ended text generation. Several existing studies, such as Table-to-Text \citep{parikh2020tottocontrolledtabletotextgeneration}, have explored this sub-task. In the future, we can enhance performance and efficiency by replacing this step with specifically trained small language models, which could further improve the semantic density of the verbalized table.

\noindent{\textbf{Adaptive Reasoning.}} Adaptive reasoning can be unstable, as language models may not always select the optimal strategy. We further explore this issue in Appendix~\ref{ap:adaptive}. In the future, training a dedicated machine learning model to guide LMs in selecting the most effective reasoning strategy could improve stability and performance.

\noindent{\textbf{Information Missing.}} The construction of the Table-of-Focus involves a trade-off between precision and recall. If recall is insufficient, essential information may be missing for final reasoning, while low precision can render the extracted content less useful. Although we use re-construction to mitigate information loss during the Table-of-Focus construction process, our analysis reveals that some information missing persist in row lookup. We further investigate this issue in Appendix~\ref{ap:missing}.

\noindent{\textbf{Efficiency.}} Efficiency is crucial in table processing and table understanding. To enhance efficiency, we incorporate the table peek technique, which reduces the context that language models need to process at certain steps. We further explore this technique in Appendix~\ref{ap:peek} and analyze the overall efficiency in Appendix~\ref{ap:efficiency}. In real-world applications, for optimal efficiency, we consider replacing certain steps with specialized small language models, balancing the trade-off between efficiency and performance .

\subsection{Downstream Applications}

\noindent{\textbf{Web Tables.}} The web contains a vast number of structured tables, including Wikipedia tables, government reports, and other online tabular data. Extracting and reasoning over these tables is crucial for applications such as fact verification, web search, and knowledge graph construction. \method enhances the ability to interpret, query, and reason over these tables, enabling more accurate and context-aware information retrieval.

\noindent{\textbf{Spreadsheets.}} Spreadsheets are widely used in business, finance, and scientific research for data management and analysis. Traditional spreadsheet tools require manual formula creation and human intervention to derive insights. In contrast, \method can automate tasks such as data summarization, trend analysis, anomaly detection, and reasoning-based computations. By integrating with tools like Microsoft Excel and Google Sheets, \method enables intelligent spreadsheet interactions, allowing users to query data using natural language and receive precise, structured responses.

\noindent{\textbf{Databases.}} Structured databases store vast amounts of relational data, typically accessed through SQL queries or predefined interfaces. However, many users lack SQL proficiency, posing barriers to efficient data retrieval. \method, with its Table-of-Focus mechanism, facilitates the quick understanding of large databases, enabling seamless querying of relational data without the need for manual SQL query writing. Additionally, it enhances database reasoning tasks, including knowledge extraction, making structured data more accessible to non-technical users.

In real-world applications, different scenarios have varying requirements, and it may not be necessary to incorporate all aspects of \method. Instead, certain components can be adapted or selectively applied based on specific needs.

Finally, as discussed above, there is still much work to be done in the future to further enhance language model-based table understanding. We hope this work serves as a recipe of comprehensive references on current state-of-the-art methods and provides guidance for future advancements in this field.

\section{Datasets Used for Evaluation}
\label{ap:datasets}

Table~\ref{tab:datasets} shows all datasets use for evluation in this study, license and source are also included.

\begin{table}[ht]
\centering
\caption{Benchmarks used for evaluation.}
\resizebox{0.75\textwidth}{!}{%
\begin{tabular}{lcccccl}
\toprule
Dataset & \# Test & Table Type & Domain & License & Source \\
\midrule
WikiTQ~\citep{pasupat-liang-2015-compositional}
    & 4{,}217 & Relational & Wikipedia & CC-BY-SA-4.0 & \href{https://github.com/ppasupat/WikiTableQuestions}{Link} \\
TabFact~\citep{chen2020tabfactlargescaledatasettablebased}
    & 2{,}024 & Relational & Wikipedia & CC-BY-4.0 & \href{https://github.com/wenhuchen/Table-Fact-Checking}{Link} \\
FetaQA~\citep{nan-etal-2022-fetaqa}
    & 1{,}165 & Relational & Wikipedia & CC-BY-SA-4.0 & \href{https://github.com/wzhouad/FetaQA}{Link} \\
FinQA~\citep{chen2021finqa}
    & 1{,}147 & Relational & Finance & MIT & \href{https://github.com/czyssrs/FinQA}{Link} \\
HiTab~\citep{cheng-etal-2022-hitab}
    & 1{,}583 & Hierarchical & Reports & C-UDA 1.0 & \href{https://github.com/microsoft/HiTab}{Link} \\
\bottomrule
\end{tabular}
}%
\label{tab:datasets}
\end{table}

\section{Detailed Settings of Challenge Analysis Experiments}
\label{ap:analysis_settings}

We conduct extensive experiments to analyze the challenges of table understanding with language models (LMs). Specifically, we perform challenge analysis experiments on the WikiTQ dataset \citep{pasupat-liang-2015-compositional}, which consists of 4,344 data instances. Following previous work \citep{wang2024chainoftableevolvingtablesreasoning, liu-etal-2024-rethinking}, we use the exact match of the final answer as the evaluation metric to measure accuracy. Our experiments utilize OpenAI models hosted on Microsoft Azure\footnote{\url{https://azure.microsoft.com/en-us/support/legal/}}. Unless otherwise stated, we set the temperature to 0 to ensure stable output while keeping all other hyperparameters at their default values. For each model, we use the following versions: \textit{gpt-4o} (\textit{gpt-4o-0806}), \textit{gpt-4o-mini} (\textit{gpt-4o-mini-0718}), \textit{gpt-3.5-turbo} (\textit{gpt-3.5-turbo-0125}), and \textit{o1} (\textit{o1-preview-0912}).

\begin{figure}[htbp]
    \centering
    \includegraphics[width=0.95\textwidth]{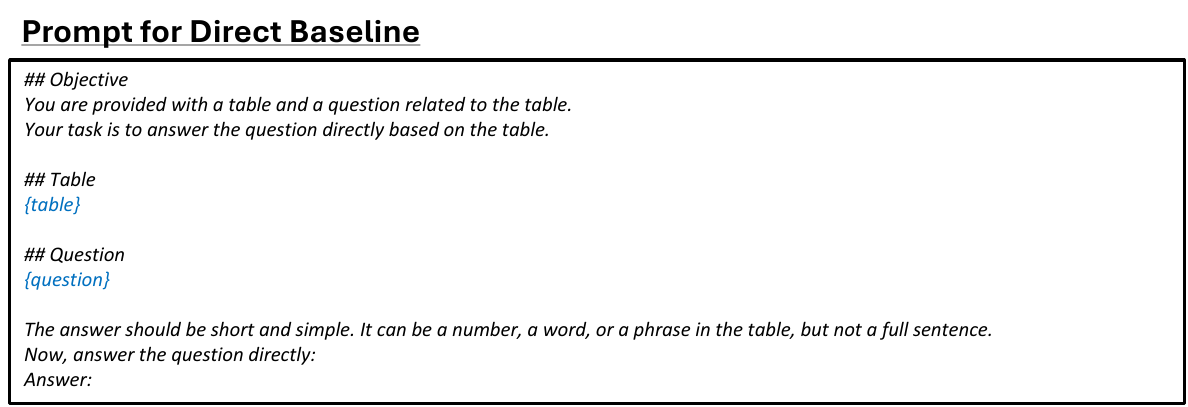}
    \centering
    \caption{Direct prompt for table understanding in analysis experiment. Blue text indicates placeholders for variables within the prompt. The prompt guides the language model to directly give the final answer based on the given table and question.}
    \label{prompt:o_direct}
\end{figure}

\noindent{\textbf{Effect of Table Size} (Figure~\ref{fig:challenge}(a)).} We evaluate how table size impacts task difficulty using a direct prompting approach (Prompt~\ref{prompt:o_direct}) with \textit{gpt-4o}, \textit{gpt-4o-mini} and \textit{gpt-3.5-turbo} to generate answers. We categorize table size based on four metrics: row count, column count, area size (computed as the product of row and column counts), and token count (measured using the \textit{cl100k\_base} encoding). The tables are divided into four size categories—small, medium, large, and extra-large—strictly partitioned into quartiles from the smallest to the largest. We then analyze results by splitting performance based on table size.

\begin{figure}[htbp]
    \centering
    \includegraphics[width=0.95\textwidth]{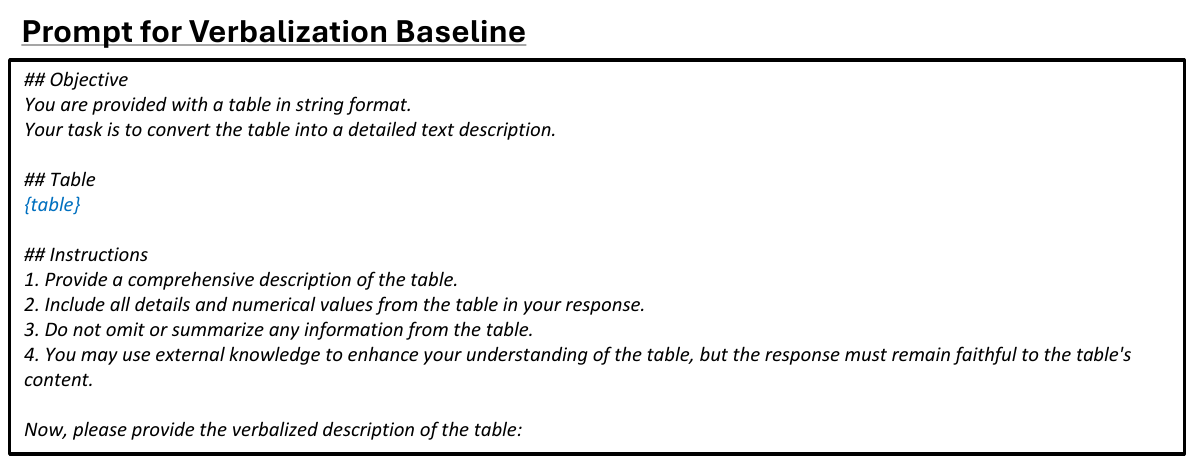}
    \centering
    \caption{Prompt for table verbalization in analysis experiment. Blue text indicates placeholders for variables within the prompt. The prompt guides the language model to verbalize a table to add detailed description.}
    \label{prompt:o_verbal}
\end{figure}

\noindent{\textbf{Effect of Verbalization} (Figure~\ref{fig:challenge}(b)).} We investigate the impact of enriching semantic context through verbalized tables by comparing three approaches. In the \textit{Table} setting, the LM processes the raw table directly using direct prompting (Prompt~\ref{prompt:o_direct}). In \textit{Table + Verbal}, the table is first verbalized using the LM itself (Prompt~\ref{prompt:o_verbal}), and both the original and verbalized tables are then provided as input. Lastly, in \textit{Table + Verbal Plus}, the verbalized table is generated using \textit{gpt-4o}, further enhancing the semantic richness of the input.

\begin{figure}[htbp]
    \centering
    \includegraphics[width=0.95\textwidth]{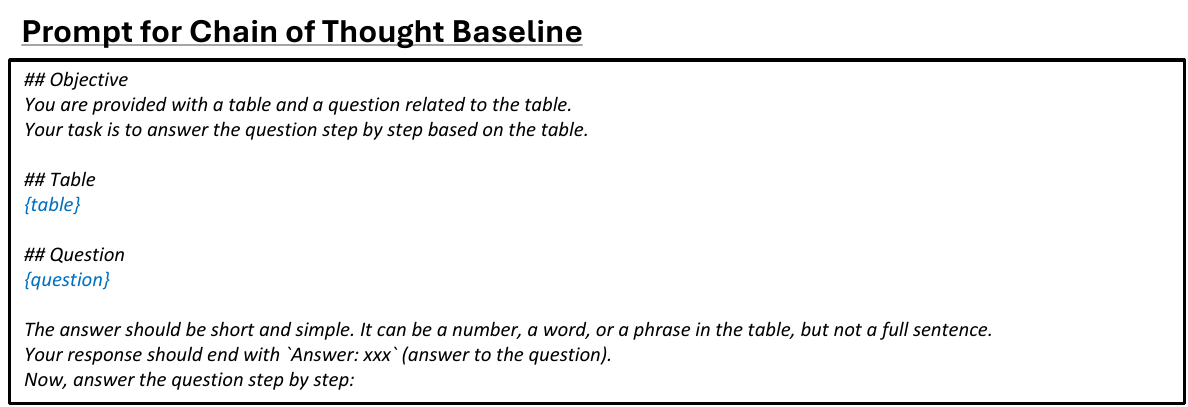}
    \centering
    \caption{Chain of thought prompt for table understanding in analysis experiment. Blue text indicates placeholders for variables within the prompt. The prompt guides the language model to give the answer step by step based on the given table and question.}
    \label{prompt:o_cot}
\end{figure}

\begin{figure}[htbp]
    \centering
    \includegraphics[width=0.95\textwidth]{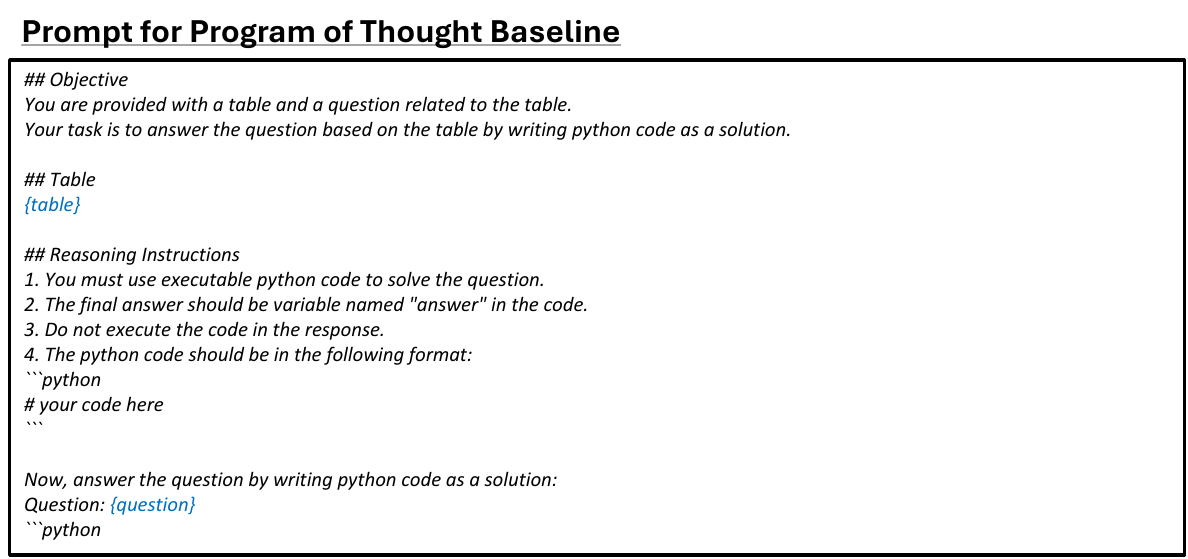}
    \centering
    \caption{Program of thought prompt for table understanding in analysis experiment. Blue text indicates placeholders for variables within the prompt. The prompt guides the language model to generate code to derive the answer based on the given table and question.}
    \label{prompt:o_pot}
\end{figure}

\begin{figure}[htbp]
    \centering
    \includegraphics[width=0.95\textwidth]{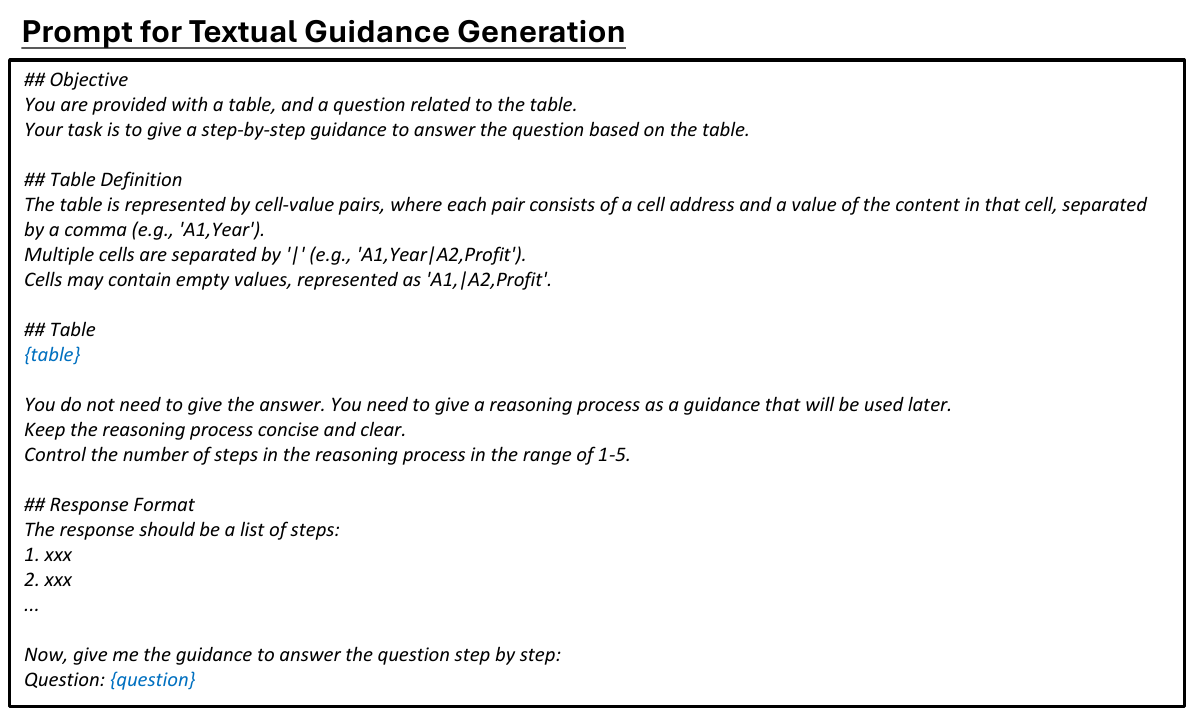}
    \centering
    \caption{Prompt for textual guidance generation in analysis experiment. Blue text indicates placeholders for variables within the prompt. The prompt guides the language model to generate textual guidance that used for symbolic reasoning.}
    \label{prompt:o_guide}
\end{figure}

\begin{figure}[htbp]
    \centering
    \includegraphics[width=0.95\textwidth]{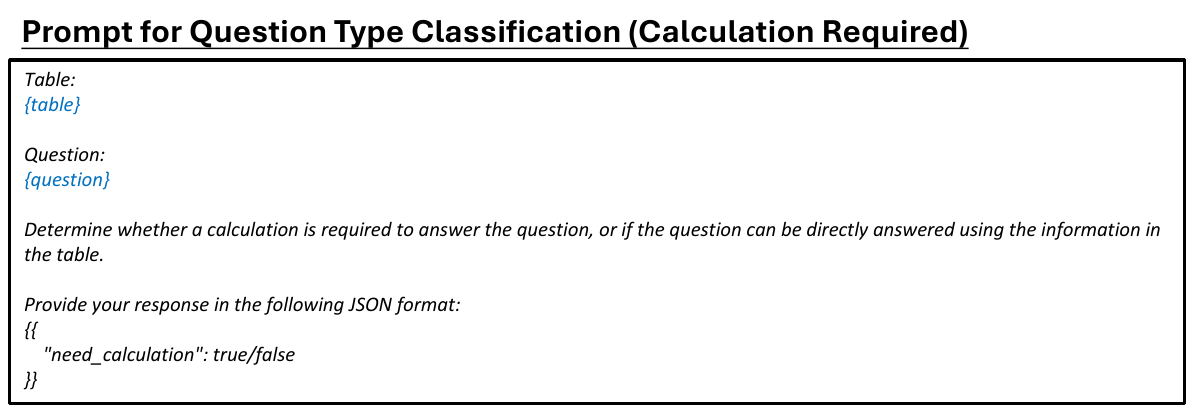}
    \centering
    \caption{Prompt for classifying a question type based on whether calculation is required in the analysis experiment. Blue text indicates placeholders for variables within the prompt.}
    \label{prompt:o_classify_qa}
\end{figure}

\noindent{\textbf{Comparison of Reasoning Methods} (Figure~\ref{fig:challenge}(c)).} We compare different reasoning approaches—textual reasoning (Prompt~\ref{prompt:o_cot}), symbolic reasoning (Prompt~\ref{prompt:o_pot}), and text-guided symbolic reasoning (Prompt~\ref{prompt:o_guide})—on calculation-required versus non-calculation questions using \textit{gpt-3.5-turbo}. To classify WikiTQ questions into calculation-required or not, we use \textit{o1} (Prompt~\ref{prompt:o_classify_qa}), identifying 2,692 calculation-required questions and 1,652 non-calculation questions. The results are then analyzed based on this classification.

\begin{figure}[htbp]
    \centering
    \includegraphics[width=0.95\textwidth]{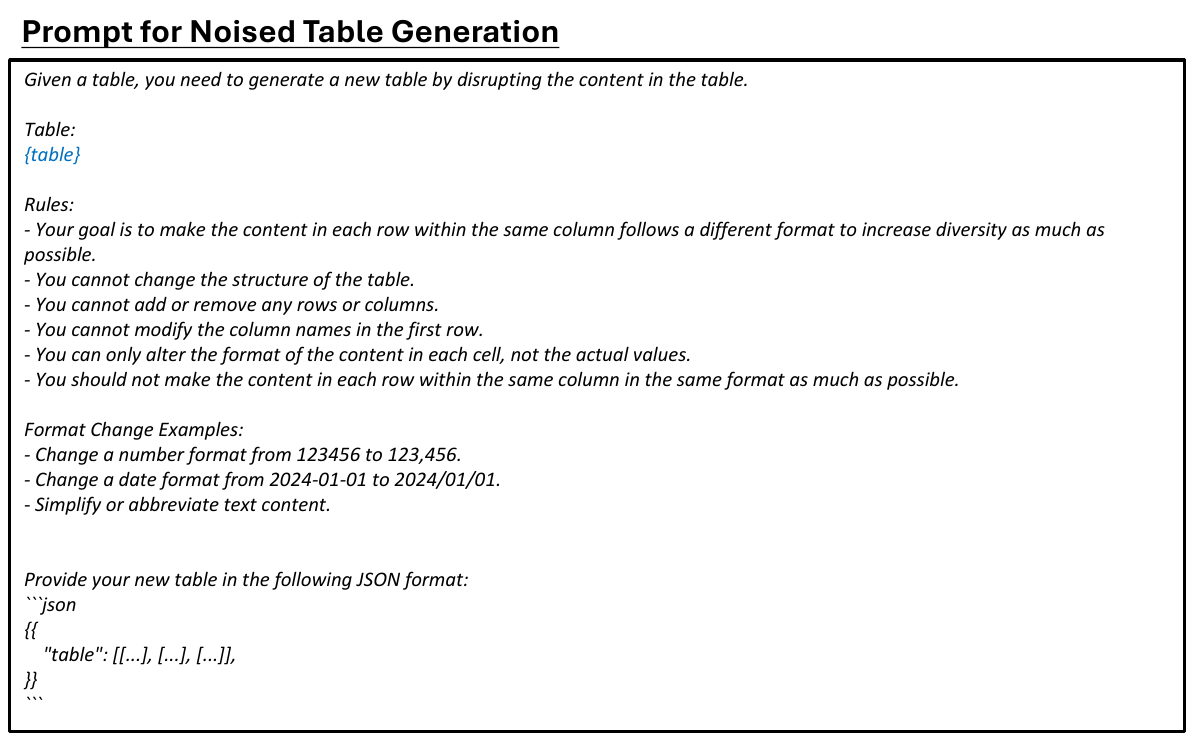}
    \centering
    \caption{Prompt for generating noised tables in the analysis experiment. Blue text represents placeholders for variables within the prompt. The prompt instructs the language model to add noise by altering the cell content format based on a given table.}
    \label{prompt:o_noise}
\end{figure}

\noindent{\textbf{Impact of Noisy Tables} (Figure~\ref{fig:challenge}(d)).} We investigate how performance varies between normalized and noisy tables. To generate noisy tables, we use \textit{o1} (Prompt~\ref{prompt:o_noise}), instructing it to introduce noise into table contents while preserving actual values and diversifying entries within columns. Additionally, each table has a 50\% chance of being randomly transposed from the default row-major format to the column-major format. We then filter the generated tables through a combination of human verification and \textit{o1} checks to ensure that answers remain derivable from the noisy tables. After filtering, 2,565 noisy tables remain. We evaluate textual reasoning (Prompt~\ref{prompt:o_cot}) and symbolic reasoning (Prompt~\ref{prompt:o_pot}) on both the noisy and original normalized tables using \textit{gpt-4o-mini}.

\section{Extended Experiments on Additional Table Understanding Benchmarks}
\label{ap:add_exp}
We perform additional experiments on diverse table‑understanding tasks to further assess the robustness of \method.

\subsection{Evaluation on Free-form QA with the FetaQA Dataset}
\label{ap:fetaqa}

\begin{table}[h]
    \centering
    \caption{Performance comparison on FetaQA. The values are multiplied by 100, and the percentage improvement represents the performance gain compared to the end-to-end QA of the base model. The results demonstrate that \method achieves strong performance in long-form question answering.}
    \resizebox{0.9\textwidth}{!}{
    \label{tab:fetaqa_results}
    \begin{tabular}{lcccc}
        \toprule
        \textbf{Methods} & \textbf{BLEU} & \textbf{ROUGE-1} & \textbf{ROUGE-2} & \textbf{ROUGE-L} \\
        \midrule
        Fine-Tuning (T5-large) \citep{ye2023largelanguagemodelsversatile} & 30.54 & 63 & 41 & 53 \\
        End-to-End QA (Codex) \citep{chen2021evaluatinglargelanguagemodels} & 27.96 & 62 & 40 & 52 \\
        \midrule
        End-to-End QA (PaLM 2) \citep{wang2024chainoftableevolvingtablesreasoning} & 28.37 & 63 & 41 & 53 \\
        Dater (PaLM 2) \citep{ye2023largelanguagemodelsversatile} & 29.47 & 63 & 41 & 53 \\
        Chain-of-Table (PaLM 2) \citep{wang2024chainoftableevolvingtablesreasoning} &
        32.61 (+14.9\%)&
        66 (+4.8\%)&
        44 (+7.3\%)&
        56 (+5.7\%)\\
        \midrule
        End-to-End QA (gpt-4o) & 24.91 & 62.05 & 41.29 & 50.36 \\
        \textbf{Ours (Tablemaster - gpt4o)} &
        \textbf{28.94 \textcolor{langgreen}{(+16.2\%)}} &
        \textbf{66.06 \textcolor{langgreen}{(+6.5\%)}} &
        \textbf{45.29 \textcolor{langgreen}{(+9.7\%)}} &
        \textbf{54.56 \textcolor{langgreen}{(+8.3\%)}} \\
        \bottomrule
    \end{tabular}
    }
\end{table}
\begin{figure*}[htbp]
    \centering
    \includegraphics[width=\textwidth]{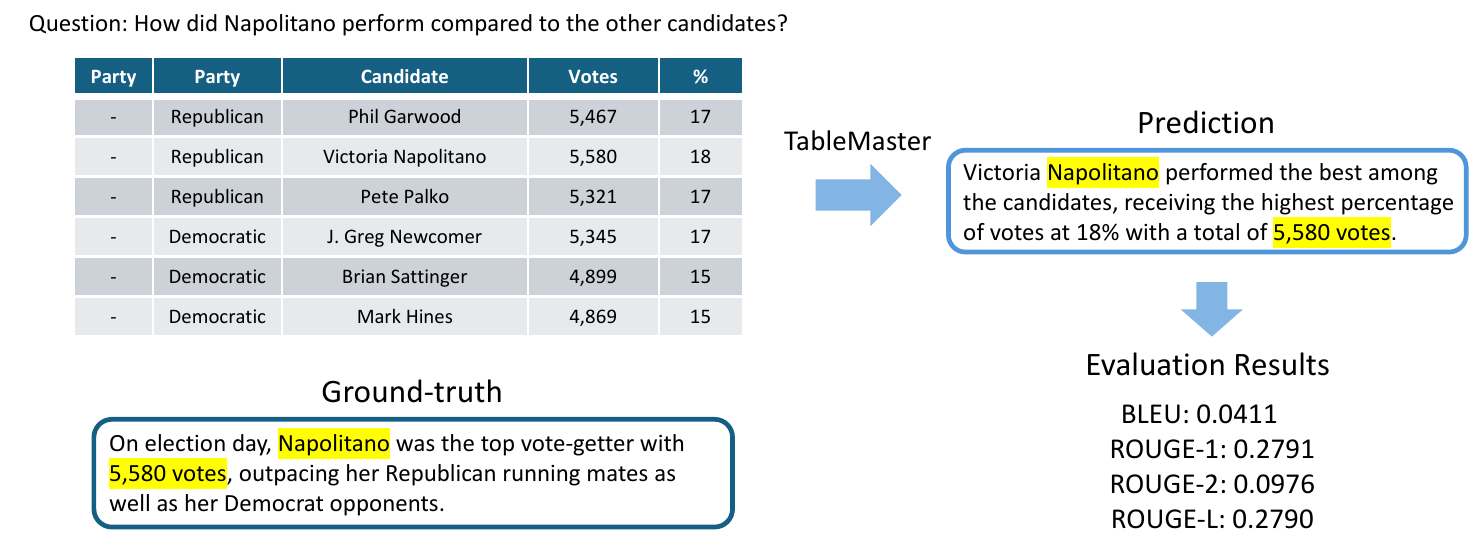}
    \caption{An example (fetaqa-164) from the FetaQA dataset where the result is accurate, but the evaluation metric assigns a low score.}
    \label{fig:fetaqa_example}
\end{figure*}

PaLM 2 has been deprecated \citep{anil2023palm2technicalreport} and is no longer accessible. Therefore, we use a comparable language model, \textit{gpt-4o}, to conduct experiments on FetaQA and compare the results with previous methods. Additionally, we use 20 exemplars for few-shot in-context learning to align with the dataset’s format.

Table~\ref{tab:fetaqa_results} shows that \method improves free-form question answering performance on FetaQA compared to the base End-to-End QA model, achieving improvements of 16.2\% in BLEU and 6.5\% in ROUGE-1. These improvements surpass those of Chain-of-Table when compared to its respective End-to-End QA baseline.

However, the improvement of \method over baseline methods remains marginal, with some values even falling below those of previous approaches in absolute terms. We believe this does not fully reflect the model’s actual performance in free-form QA. We attribute this to the n-gram text similarity metrics used in ROUGE-1/2/L \citep{lin-2004-rouge}, which are known to be insensitive to improvements gained from in-context learning \citep{maynez-etal-2023-benchmarking}. These metrics struggle to capture stylistic and structural enhancements in free-form text generation. Since models rely on instructions and a limited number of examples, they may not fully adapt to the expected output format, leading to an underestimation of performance gains.

To further investigate this, we analyze a specific case, FetaQA-164, as shown in Figure~\ref{fig:fetaqa_example}. In this instance, the BLEU and ROUGE metrics assign low scores, as only two words match in the entire sentence. However, manual review confirms that the generated answer is indeed correct—these two words are the most important, and the overall meaning of the response is both accurate and superior to the ground truth. This highlights the limitations of ROUGE in evaluating free-form QA and suggests that qualitative analysis is essential for a more comprehensive assessment of model improvements. Nonetheless, based on quantitative analysis, \method is overall effective.

\subsection{Evaluation on Hierarchical Tables with the HiTab Dataset}
\label{ap:hitab}

UUnlike WikiTQ and TabFact, the HiTab dataset~\citep{cheng-etal-2022-hitab} contains \emph{hierarchical} tables that violate the flat row–column assumption. Such structure challenges models to reason across multi‑level headers and parent–child cell relations.

We employ a \textit{GPT‑4o} backbone and follow HiTab’s official data split. MultiCoT \citep{MultiCoT2025} extends \textbf{Chain‑of‑Table} reasoning to multiple tables; both MultiCoT and \method operate on the same converted relational tables. E5 \citep{zhang-etal-2024-e5} represents the current SOTA on HiTab, being explicitly tailored for complex tables. Because our framework (\method) is designed for relational tables, we introduce a lightweight \emph{relational‑table converter} via prompting \textit{o1}: each hierarchical table is decomposed into several relational subtables, while contextual tags are propagated to preserve structural cues.

\begin{table}[ht]
\centering
\caption{Accuracy (\%) on the HiTab dataset. “After Converting” rows evaluate models on relational tables produced by our converter; “Direct” reports results on the original hierarchical tables.}
\begin{tabular}{l c}
\toprule
\textbf{Method} & \textbf{Accuracy} \\ 
\midrule
\multicolumn{2}{l}{\textit{After Converting to Relational Tables}} \\[2pt]
\quad MultiCoT (original~\citep{MultiCoT2025})                & 64.0 \\ 
\quad MultiCoT (optimized prompt)                            & 70.0 \\ 
\quad MultiCoT (optimized prompt + verbalized table)         & 73.5 \\ 
\quad \textbf{\method}                                   & \textbf{74.2} \\ 
\midrule
\multicolumn{2}{l}{\textit{Direct}} \\[2pt]
\quad E5~\citep{zhang-etal-2024-e5}                           & 77.3 \\ 
\bottomrule
\end{tabular}
\label{tab:hitab_results}
\end{table}

Table~\ref{tab:hitab_results} shows that \method outperforms the strongest chain‑of‑table baseline by +0.7 point (from 73.5 to 74.2). Residual errors are mainly due to information loss during the conversion step. Future work will integrate hierarchical relations directly into the reasoning module.

\subsection{Evaluation on Numerical Reasoning with the FinQA Dataset}
\label{ap:finqa}

FinQA \citep{chen2021finqa} requires multi‑step numerical reasoning over financial reports—e.g., computing growth rates or combining multiple cells with arithmetic operators.  Hence it evaluates whether \method can execute numerical formulas correctly, not just extract text spans. We keep the same training recipe as in. Two backbones are considered: \textit{GPT‑4o‑mini} (4m) and \textit{GPT‑4o} (4o).

\begin{table}[ht]
\centering
\caption{Accuracy on FinQA. \method consistently boosts numerical‑reasoning accuracy over both backbones.}
\begin{tabular}{lcc}
\toprule
\textbf{Method} & \textbf{Accuracy} (\%) & \textbf{$\Delta$} \\
\midrule
GPT‑4o‑mini         & 50.7 & -- \\
\textbf{\method} (4m)  & \textbf{66.4} & +15.7 \\
\cmidrule(lr){1-3}
GPT‑4o              & 63.1 & -- \\
\textbf{\method} (4o)   & \textbf{70.9} & +7.8 \\
\bottomrule
\end{tabular}
\label{tab:finqa_results}
\end{table}

Table~\ref{tab:finqa_results} shows that \method delivers impressive improvements on both backbones, demonstrating that its symbolic reasoning module reliably handles complex calculations in the financial domain.

\section{Table Understanding Baselines}
\begin{table}[ht]
\centering
\caption{Results of our reproduced baselines on WikiTQ and TabFact. The values in the table represent accuracy (\%).}
\label{tab:baselines}
\begin{tabular}{@{}l l | c c@{}} 
\toprule[0.8pt]
\textbf{Base LLM} & \textbf{Method} & \textbf{WikiTQ} & \textbf{TabFact}\\
\midrule
o1-preview\textsubscript{$\sim$300B} & Direct 
  & 84.60 & 92.05 \\
o1-mini\textsubscript{$\sim$100B} & Direct
  & 83.49 & 91.35 \\
\midrule
\multirow{4}{*}{gpt-4o\textsubscript{$\sim$200B}}
 & Direct          & 73.07 & 84.73 \\
 & Chain of Thought  & 83.98 & 91.90 \\
 & Program of Thought & 74.63 & 90.02 \\
 \cmidrule(lr){2-4}
 & \textbf{\method (gpt-4o)} & \textbf{84.55} & \textbf{94.52} \\
\midrule
\multirow{4}{*}{gpt-4o-mini\textsubscript{$\sim$8B}}
 & Direct          & 59.53 & 71.25 \\
 & Chain of Thought  & 72.97 & 87.40 \\
 & Program of Thought & 61.83 & 85.18 \\
 \cmidrule(lr){2-4}
 & \textbf{\method (gpt-4o-mini)} & \textbf{78.13} & \textbf{90.12} \\
\midrule
\multirow{4}{*}{gpt-3.5-turbo\textsubscript{$\sim$175B}}
 & Direct          & 56.58 & 70.90 \\
 & Chain of Thought  & 59.92 & 69.52 \\
 & Program of Thought & 50.32 & 68.82 \\
 \cmidrule(lr){2-4}
 & \textbf{\method (gpt-3.5-turbo)} & \textbf{68.21} & \textbf{83.65} \\
\bottomrule[0.8pt]
\end{tabular}
\end{table}

To better facilitate future research, we evaluate different reasoning methods across various base models. Table~\ref{tab:baselines} presents the accuracy results of our reproduced baselines on WikiTQ and TabFact, comparing different base LLMs and reasoning methods. The table includes evaluations on \textit{o1-preview} ($\sim$300B), \textit{o1-mini} ($\sim$100B), \textit{gpt-4o} ($\sim$200B), \textit{gpt-4o-mini} ($\sim$8B), and \textit{gpt-3.5-turbo} ($\sim$175B). The exact number of parameters for several LMs (e.g., GPT, o1) has not been publicly disclosed. Most parameter counts are estimates reported to provide context for understanding model performance. For more precise information, please refer to the original or future official documentation \citep{abacha2025medecbenchmarkmedicalerror}. Each model is tested with various reasoning strategies, including Direct, chain of thought, and Program of Thought, alongside our proposed \method.

Across all base models, TableMaster consistently achieves the highest accuracy. For gpt-4o, TableMaster reaches 84.55\% on WikiTQ and 94.52\% on TabFact, outperforming both chain of thought (83.98\%, 91.90\%) and Program of Thought (74.63\%, 90.02\%). Similarly, for gpt-4o-mini, TableMaster achieves 78.13\% on WikiTQ and 90.12\% on TabFact, significantly improving over the Direct method (59.53\%, 71.25\%) and surpassing chain of thought (72.97\%, 87.40\%).

The performance gap is even more pronounced for gpt-3.5-turbo, where TableMaster reaches 68.21\% on WikiTQ and 83.65\% on TabFact, significantly outperforming both chain of thought (59.92\%, 69.52\%) and Program of Thought (50.32\%, 68.82\%). Interestingly, we observe that while \method’s improvement is limited on gpt-4o, the weaker the base model, the greater the performance improvement. While o1-preview and o1-mini achieve high accuracy with the Direct method (84.60\%, 92.05\% for o1-preview and 83.49\%, 91.35\% for o1-mini), the results of \method on gpt-4o demonstrate that our method is capable of achieving state-of-the-art performance across different LL architectures.

Additionally, we find that chain of thought reasoning is highly effective, achieving strong accuracy across models. Even a simple chain of thought approach outperforms previous methods that rely solely on symbolic reasoning \citep{mao2024potableprogrammingstandardlytablebased}, indicating that chain of thought should be retained as a key component in the reasoning framework.

These results confirm that \method enhances table reasoning performance across various LLMs, effectively outperforming both direct prompting and traditional reasoning strategies, particularly in cases where table complexity and reasoning demands are higher.

\section{Performance Analysis Under Different Table Sizes}
\begin{table}[ht]
\centering
\caption{Performance Comparison Across Table Sizes (Token).}
\label{tab:size_analysis}
\begin{tabular}{@{}lcccc@{}}
\toprule
\multirow{2}{*}{\textbf{Method}} & \multicolumn{3}{c}{\textbf{Table Size (Token)}} \\ \cmidrule(l){2-4}
 & \textbf{Small (<2k)} & \textbf{Medium (2k $\sim$ 4k)} & \textbf{Large (>4k)} \\ \midrule
Binder \citep{cheng2023bindinglanguagemodelssymbolic} & 56.54 & 26.13 & 6.41 \\
Dater \citep{ye2023largelanguagemodelsversatile} & 62.50 & 42.34 & 34.62 \\
Chain-of-Table \citep{wang2024chainoftableevolvingtablesreasoning} & 68.13 & 52.25 & 44.87 \\
\textbf{\method (gpt-3.5-turbo)} & \textbf{69.01} & \textbf{58.00} & \textbf{56.73} \\
\midrule
\method (gpt-4o-mini) & 78.71 & 70.50 & 70.19 \\ \bottomrule
\end{tabular}
\end{table}
\begin{figure*}[htbp]
    \centering
    \includegraphics[width=0.9\textwidth]{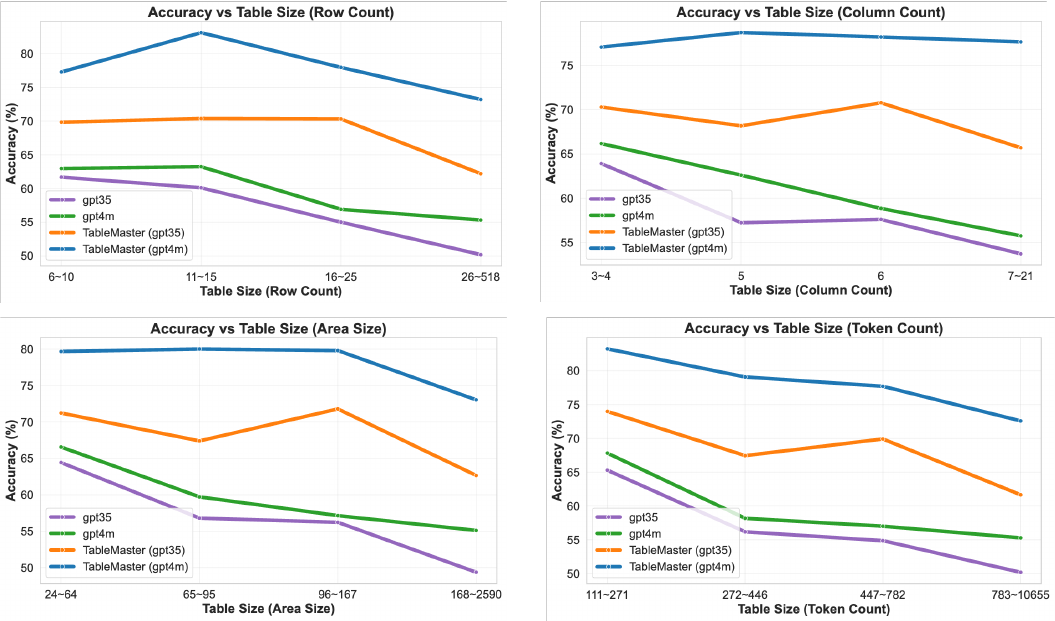}
    \caption{Performance Comparison Across Table Sizes (Row Count, Column Count, Area Size, Token Count).}
    \label{fig:table_size_analysis}
\end{figure*}

Table~\ref{tab:size_analysis} presents a performance comparison across different table sizes, categorized into small (<2k tokens), medium (2k$\sim$4k tokens), and large (>4k tokens). The results compare several methods, including Binder \citep{cheng2023bindinglanguagemodelssymbolic}, Dater \citep{ye2023largelanguagemodelsversatile}, and Chain-of-Table \citep{wang2024chainoftableevolvingtablesreasoning}, against \method. All methods are evaluated using \textit{gpt-3.5-turbo}, with additional results of \method provided for \textit{gpt-4o-mini}.

Across all table sizes, \method consistently outperforms baseline methods. Specifically, for \textit{gpt-3.5-turbo}, \method achieves the highest performance in all table size categories, scoring 69.01\% on small tables, 58.00\% on medium tables, and 56.73\% on large tables. This demonstrates its ability to maintain robust performance even as table size increases, significantly outperforming Binder, Dater, and Chain-of-Table, especially on medium and large tables, where the performance gap becomes more pronounced.

Furthermore, \method with \textit{gpt-4o-mini} achieves even stronger performance, with accuracy scores of 78.71\% (small tables), 70.50\% (medium tables), and 70.19\% (large tables). These results highlight that leveraging stronger base models further enhances \method’s effectiveness, making it particularly well-suited for large-scale table reasoning tasks. Notably, when transitioning from medium to large tables, \method (\textit{gpt-4o-mini}) experiences only a 0.31\% performance drop (from 70.50\% to 70.19\%), demonstrating its strong capability in handling increasing table complexity. This minimal decline contrasts sharply with other methods, which show significantly larger drops, further reinforcing the scalability and robustness of \method in processing large-scale tabular data.

Figure~\ref{fig:table_size_analysis} illustrates the accuracy trends of different models across various table sizes, categorized based on row count, column count, area size, and token count. The models evaluated in this study include \textit{gpt-3.5-turbo} (\textit{gpt35}), \textit{gpt-4o-mini} (\textit{gpt4m}), \method (\textit{gpt35}), and \method (\textit{gpt4m}). The results provide insights into how these models handle increasing table complexity and size, revealing the comparative strengths and limitations of each approach. The size split in this study is strictly partitioned into quartiles, ranging from the smallest to the largest tables.

\noindent{\textbf{Row Count.}} The top-left plot analyzes accuracy trends as row count increases. \method (\textit{gpt4m}) consistently outperforms other models, maintaining high accuracy levels even with an increasing number of rows. In contrast, \textit{gpt-3.5-turbo} (\textit{gpt35}) starts with the highest accuracy, peaking in the 11–15 row range before experiencing a decline as row count further increases. Smaller models such as \textit{gpt35} and \textit{gpt4m} exhibit a sharper decline, highlighting the challenge of processing larger tables with more rows.

\noindent{\textbf{Column Count.}} The top-right plot examines model performance as column count increases. \method (\textit{gpt4m}) again achieves strong performance, peaking at around five columns before showing a slight decline. This result highlights the effectiveness of \textbf{table-of-focus re-construction}, demonstrating that column re-selection can effectively adapt to scenarios with many columns. While \textit{gpt35} initially maintains the highest accuracy, other models experience a steeper drop as the number of columns increases. These trends suggest that column-heavy tables pose greater challenges for reasoning compared to row-heavy tables, likely due to the increased dimensional complexity and interdependencies between attributes.

\noindent{\textbf{Area Size.}} The bottom-left plot evaluates the relationship between accuracy and table area size, calculated as the product of row and column counts. \method (\textit{gpt4m}) reaches peak performance in the mid-range (96–167 area size) before slightly declining for larger tables. \textit{gpt35} initially performs well but deteriorates as table area size increases, while \textit{gpt4m} and \textit{gpt35} show a noticeable decline overall, reinforcing that larger tables significantly impact accuracy across models.

\noindent{\textbf{Token Count.}} The bottom-right plot assesses accuracy as a function of table token count, which reflects the amount of textual information models need to process. \method (\textit{gpt4m}) consistently achieves the highest accuracy, followed by \method (\textit{gpt35}). A general downward trend is observed across all models as token count increases, indicating that larger input lengths negatively affect performance. Notably, \textit{gpt35} experiences the sharpest drop, suggesting its lower capacity for handling long-context table data compared to \textit{gpt4m}.

Overall, these findings confirm that \method is highly scalable and generalizable across different table sizes, consistently outperforming previous methods, particularly in handling larger and more complex tables. Its robust performance and gradual decline in accuracy as table size increases make it a reliable and efficient solution for table-based reasoning tasks.

\section{Performance Analysis Under Different Table Peek Sizes}
\label{ap:peek}

\begin{figure*}[h]
  \centering
  \begin{subfigure}[h]{0.47\textwidth}
    \centering
    \includegraphics[width=\linewidth]{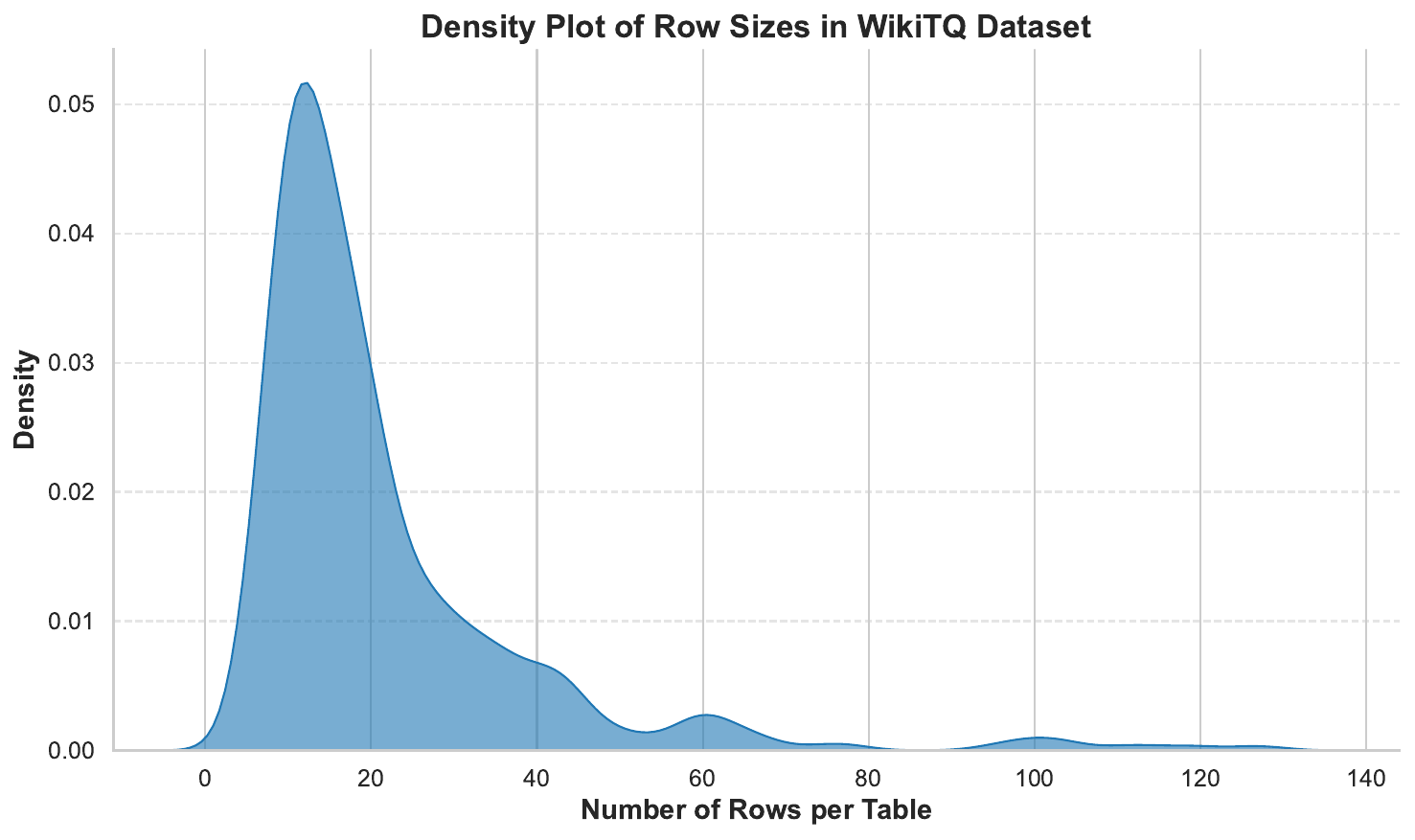}
    \caption{}
    \label{fig:wikitq_row}
  \end{subfigure}
  \hfill
  \begin{subfigure}[h]{0.47\textwidth}
    \centering
    \includegraphics[width=\linewidth]{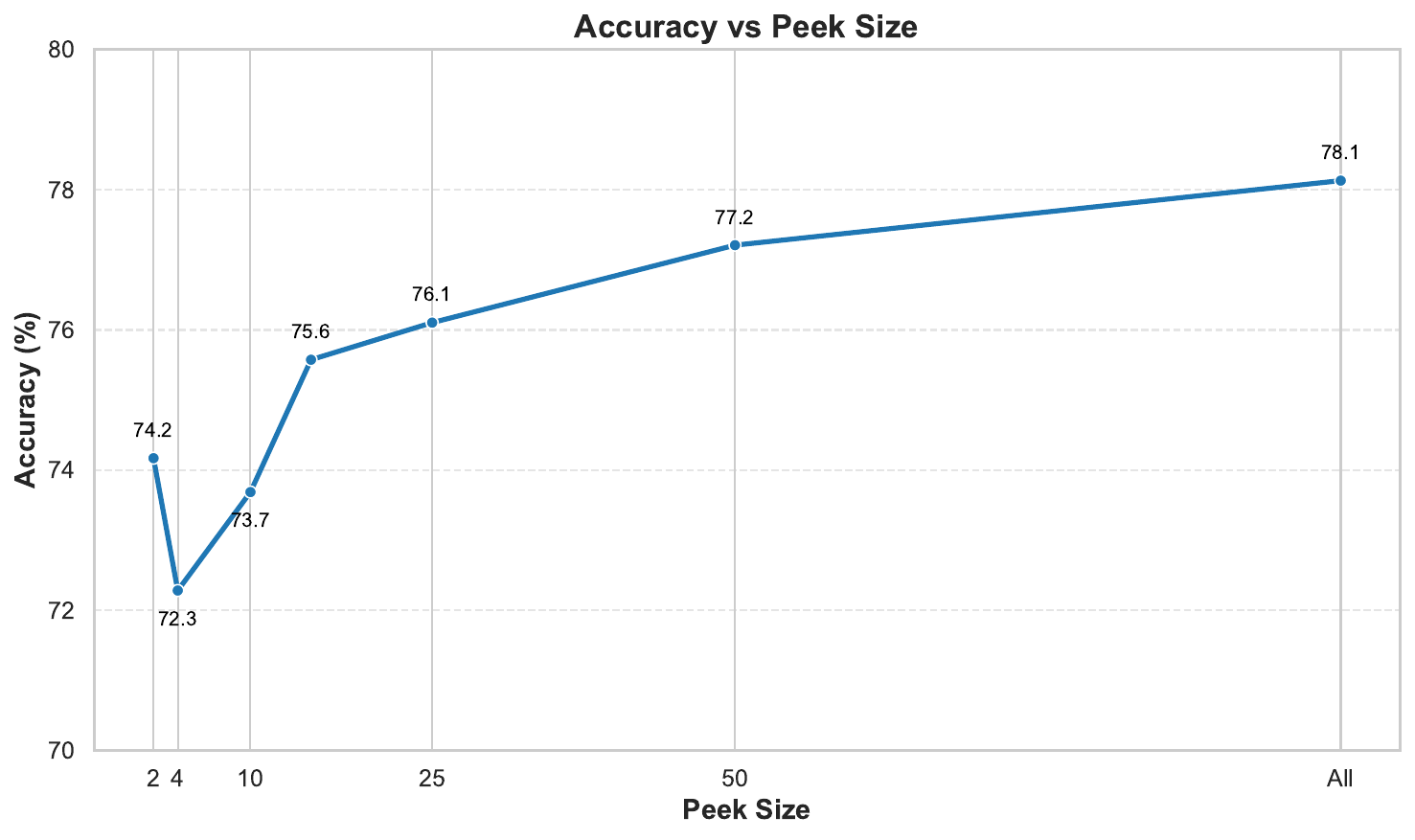}
    \caption{}
    \label{fig:peek}
  \end{subfigure}
    \caption{The row count distribution in the WikiTQ dataset and the analysis of accuracy variation with different peek sizes.}
  \label{fig:peek_combined}
\end{figure*}

We propose the concept of table peek, which enhances the efficiency of \method for table understanding tasks by reducing the context that language models need to process at certain steps.

To analyze the effectiveness of this approach, we first examine the row count distribution in the WikiTQ dataset, as shown in Figure~\ref{fig:wikitq_row}. To improve visualization, we remove 72 extreme outliers with exceptionally large row counts. The resulting density plot illustrates that the majority of tables contain fewer than 20 rows, with a pronounced peak around 10 rows. As the number of rows increases, the density gradually declines, indicating that large tables are relatively uncommon. Although a small number of tables exceed 100 rows, their frequency is minimal.

The line graph in Figure~\ref{fig:peek} illustrates the variation in accuracy with different peek sizes, where the peek size determines the number of rows considered during processing. Initially, accuracy is relatively low when only a small number of rows (e.g., 2–4) are used, reaching its minimum at a peek size of 4. We hypothesize that this occurs because, at a peek size of 2, the table includes only the top headers and a single example row, which may provide a clear structure for the language model to follow. However, at a peek size of 4, the table includes three example rows, potentially causing the language model to overfit the first few rows and misinterpret the overall table structure. This misalignment may lead to ineffective SQL generation for row lookup, resulting in a temporary drop in accuracy.

As the peek size increases, accuracy improves significantly, showing a sharp rise up to 25 rows. Beyond this point, the accuracy continues to improve but at a slower rate, eventually reaching its peak when the entire table is utilized (`All'). This trend suggests that a moderate peek size can effectively balance efficiency and accuracy, eliminating the need to process the full table while still maintaining strong performance.

\section{Efficiency Analysis of \method}
\label{ap:efficiency}

\subsection{Theoretical Analysis}

Efficiency is a critical factor in table-understanding methods. We analyze the efficiency of \method theoretically, following the notations introduced in Section~\ref{sec:recipe}. Our analysis considers the length of the table input as the primary computational cost, excluding any additional prompts or external information, and does not account for output length. This is because, in most cases, the output is relatively short compared to the large volume of data in the table. Specifically, we define the computational cost in terms of the total area size of the table that the language model processes.

Below are the main components of our efficiency analysis:
\begin{itemize}
    \item \textbf{Structure extraction:} $k \times n$
    \item \textbf{Row lookup:} $k \times n$
    \item \textbf{Column lookup:} $n$
    \item \textbf{Table-of-Focus Re-Construction} $a \times b \times e$
    \item \textbf{Table Verbalization:} $a \times b$,
    \item \textbf{Reasoning Strategy Assessment:} $a \times b$,
    \item \textbf{Reasoning:} $1.5\, a \times b$ (where the factor 1.5 accounts for textual processes weighted as 1 and symbolic processes weighted as 2)
\end{itemize}

 Here, $k$ represents the size of table peek, and $e$ represents the number of table-of-focus re-constructions after information estimation. $a$ and $b$ denote the dimensions of the table-of-focus $\mathbb{T}_{a \times b}$. Combining these components, the total computational cost is given by:

\begin{equation}
\label{eq:cost}
\text{Total Cost}
= (2\,k + 1) \times n
+ (e + 2.5) \times (a \times b).
\end{equation}

\subsection{Empirical Analysis}

\begin{figure*}[htbp]
    \centering
    \includegraphics[width=0.4\textwidth]{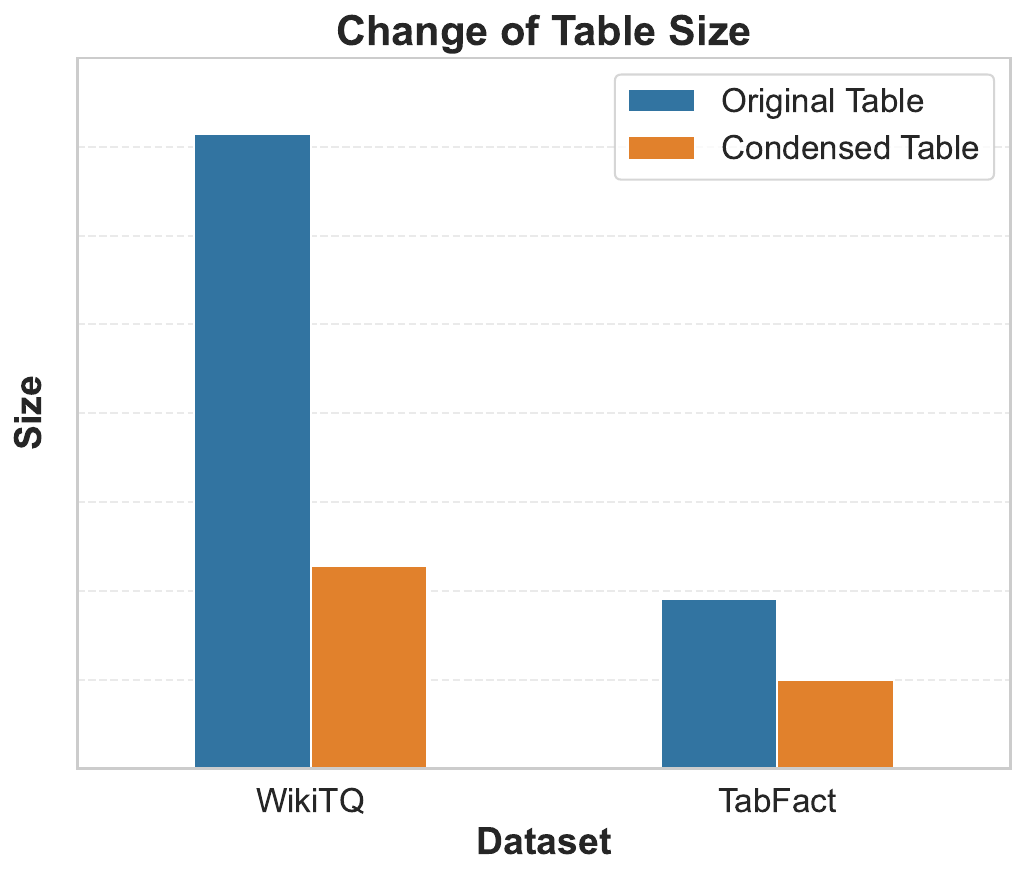}
    \caption{Changes in Table Condensation After Table-of-Focus Construction in Table Structure Understanding.}
    \label{fig:condense}
\end{figure*}

The bar chart in Figure~\ref{fig:condense} illustrates the change in table area size before and after table condensation for the WikiTQ and TabFact datasets. The y-axis represents the table size, while the x-axis categorizes the datasets. Each dataset has two bars: the blue bar represents the original table size, and the orange bar represents the condensed table size after table-of-focus construction. WikiTQ exhibits a significant reduction in table size, approximately 1:3, with the condensed table being much smaller than the original. In contrast, TabFact also undergoes condensation but to a lesser extent, around 1:2. This suggests that WikiTQ tables require more substantial structural modifications to focus on relevant content, while TabFact tables need comparatively less condensation.

As shown in Equation~\ref{eq:cost}, the theoretical cost is independent of the number of rows $m$, while $a \times b$ reflects the size of the small sub-table, which is influenced by the estimated table condensation ratio 2.5. As stated in Table~\ref{tab:reconstruct}, the reconstruction occurs 1.5 averagely, so $e$ is typically 1.5. In an ideal scenario, if the peek size is negligible, the cost is approximately $1.6 \times (m \times n)$. In the worst-case scenario, where the entire table must be examined and all content is required, the cost reaches $6 \times (m \times n)$ approximately.The estimation range for each table is 1.6 to 6 times the original table size.

Recent advancements in table understanding, such as \textsc{Chain-of-Table} \citep{wang2024chainoftableevolvingtablesreasoning} and \textsc{Tree-of-Table} \citep{ji2024treeoftableunleashingpowerllms}, involve a step-by-step evolution of tables through a long chain of transformations. In each new step, both the original table and the newly generated sub-table must be processed by language models. Additionally, their iterative process is complex, unstable, and difficult to analyze theoretically. In contrast, our approach is general and comprehensive, avoiding the trivial overhead of sub-table extraction. Instead, it focuses on holistic reasoning while maintaining ideal efficiency.

On the first 100 examples of TabFact, we evaluate \method against the representative baseline \textsc{Chain-of-Table}~\citep{wang2024chainoftableevolvingtablesreasoning} using GPT-4o, ensuring a fair comparison without self-consistency decoding.

\begin{table}[h]
\centering
\caption{Token usage comparison on the TabFact subset using GPT-4o.}
\begin{tabular}{lcc}
\toprule
\textbf{Method} & \textbf{Prompt Tokens} & \textbf{Completion Tokens} \\
\midrule
Chain-of-Table & 13,209.6 & 914.2 \\
TableMaster (Ours) & 3,393.5 & 738.6 \\
\bottomrule
\end{tabular}
\label{tab:token_usage}
\end{table}

As shown in Table~\ref{tab:token_usage}, \method consumes substantially fewer tokens while maintaining strong performance. While our design may appear intricate, it integrates practical features such as fallbacks to early-exit or full-table reasoning, thereby avoiding unnecessary table transformations. In contrast, \textsc{Chain-of-Table} continuously evolves the table without fallback safeguards, leading to much higher token usage. This limitation is not acknowledged in the original paper, whereas our framework explicitly incorporates token efficiency as a design principle.

\section{Detailed Algorithm of Table-of-Focus Re-Construction}
\label{ap:re-construction}

Here, we provide a detailed description of the Table-of-Focus Re-Construction algorithm, as shown in Algorithm~\ref{alg:tof}.

\begin{algorithm}[!]
\caption{Algorithm of Table-of-Focus Re-Construction}
\label{alg:tof}
\begin{algorithmic}[1]
\REQUIRE $\mathbb{T}$: The original table
\REQUIRE $Q$: The question
\REQUIRE $R$: Selected rows
\REQUIRE $C^0$: Initially selected columns
\REQUIRE $\mathbb{C}$: Ranked candidate column indices
\ENSURE $\mathbb{T}^{F}$: Final table-of-focus
\ENSURE $C$: Updated selected columns

\STATE Initialize $C^{candidate} \leftarrow \{ c \in \mathbb{C} \mid c \notin C^0 \}$  
\STATE Initialize $C \leftarrow \mathrm{Copy}(C^0)$  

\WHILE{\algorithmictrue}  
    \STATE $\mathbb{T}^{F} \leftarrow \text{extractTable}(\mathbb{T}, R, C)$  
    \STATE $E \leftarrow \text{estimateInformation}(\mathbb{T}^{F}, Q)$  
    \IF{$E\ \OR\ \text{len}(C^{candidate}) = \emptyset$}  
        \STATE \textbf{break}  
    \ELSE  
        \STATE $c \leftarrow \mathrm{popFront}(C^{candidate})$ \hfill \COMMENT{Select the next candidate column}  
        \STATE $C \leftarrow C \cup \{ c \}$  
    \ENDIF  
\ENDWHILE

\STATE \textbf{return} $\mathbb{T}^{F}, C$
\end{algorithmic}
\end{algorithm}

The Table-of-Focus Re-Construction Algorithm iteratively refines a table by selecting relevant columns to form the final table-of-focus $\mathbb{T}^{F}$. It starts by initializing the set of candidate columns $C^{candidate}$ that are not part of the initially selected columns $C^0$, and copies $C^0$ to initialize $C$. In each iteration, it extracts a sub-table $\mathbb{T}^{F}$ using the current selected columns and estimates whether the extracted sub-table contains sufficient information to answer the given question $Q$. If the information is sufficient $E = \text{True}$ or no more candidate columns remain, the process terminates. Otherwise, the next ranked candidate column is selected and added to $C$, repeating the process. The algorithm ultimately returns the refined table $\mathbb{T}^{F}$ and the updated set of selected columns, ensuring an efficient and structured approach to dynamically refining a table while balancing relevance and minimal table size.

\section{Analysis of Table-of-Focus Re-Construction}
\label{ap:reconstruct-analysis}

\begin{table}[ht]
\centering
\caption{Column Selection Statistics Before and After Table-of-Focus Re-Construction for TabFact and WikiTQ.}
\label{tab:reconstruct}
\resizebox{\textwidth}{!}{%
\begin{tabular}{lcccccc}  
\toprule
\textbf{Dataset} & \multicolumn{2}{c}{\textbf{Initial Columns}} & \multicolumn{2}{c}{\textbf{Final Columns}} & \multicolumn{2}{c}{\textbf{Added Columns}} \\
\cmidrule(lr){2-3} \cmidrule(lr){4-5} \cmidrule(lr){6-7}
 & \textbf{Number (\#)} & \textbf{Percentage (\%)} 
 & \textbf{Number (\#)} & \textbf{Percentage (\%)} 
 & \textbf{Number (\#)} & \textbf{Percentage (\%)} \\
\midrule
TabFact & 2.44 & 40.74 & 3.34 & 54.64 & 0.90 & 13.91 \\
WikiTQ  & 2.87 & 47.67 & 4.72 & 75.91 & 1.85 & 28.23 \\
\bottomrule
\end{tabular}
}
\end{table}

Table~\ref{tab:reconstruct} presents Column Selection Statistics before and after Table-of-Focus Re-Construction for two datasets: TabFact and WikiTQ. The table measures how many columns were initially selected, how many remained after refinement, and how many were newly added during the reconstruction process.

The table is structured into three main sections: Initial Columns, Final Columns, and Added Columns. Each section includes two metrics: the number of columns and the percentage of total columns in the dataset. The Initial Columns represent the starting number of columns before any refinement. The Final Columns show the number of columns retained after the reconstruction process. The Added Columns indicate the number of additional columns incorporated to enhance table comprehension.

For the TabFact dataset, the number of Initial Columns is 2.44, covering 40.74\% of the table’s total columns. After the reconstruction process, the Final Columns increase to 3.34, covering 54.64\%. This means that 0.90 additional columns were introduced averagely, which accounts for 13.91\% of the total columns. For the WikiTQ dataset, the pattern is similar but with higher values. The Initial Columns start at 2.87, representing 47.67\% of the total table. After reconstruction, the Final Columns expand to 4.72, covering 75.91\% of the table’s total structure. This increase results from 1.85 additional columns, which make up 28.23\% of the total columns.

Overall, this mechanism has been proven to be effective while remaining lightweight. The table demonstrates that Table-of-Focus Re-Construction slightly increases the number of selected columns, with a more pronounced effect in the WikiTQ dataset compared to TabFact. This suggests that WikiTQ tables require a greater degree of expansion to ensure adequate information coverage, whereas TabFact tables undergo a more moderate refinement process.

\section{Analysis of Adaptive Reasoning}
\label{ap:adaptive}

\begin{table}[ht]
    \centering
    \caption{Performance of Different Reasoning Methods Across Base LLMs}
    \resizebox{\textwidth}{!}{ 
    \begin{tabular}{ll|ccc}
        \toprule
        \textbf{Base LLM} & \textbf{Method} & \textbf{Calculation Required \textsubscript{\#2692}} & \textbf{No Calculation Required \textsubscript{\#1652}} & \textbf{Overall \textsubscript{\#4344}} \\
        \midrule
        \multirow{3}{*}{gpt-4o} 
        & Textual Reasoning & 81.17 & 88.56 & 83.98 \\
        & Symbolic Reasoning & 74.59 & 74.70 & 74.63 \\
        & Text-Guided Symbolic Reasoning & 76.49 & 77.36 & 76.82 \\
        \midrule
        \multirow{3}{*}{gpt-4o-mini} 
        & Textual Reasoning & 67.50 & 81.90 & 72.97 \\
        & Symbolic Reasoning & 61.55 & 62.29 & 61.83 \\
        & Text-Guided Symbolic Reasoning & 67.24 & 71.43 & 68.83 \\
        \midrule
        \multirow{3}{*}{gpt-3.5-turbo} 
        & Textual Reasoning & 52.27 & 72.40 & 59.92 \\
        & Symbolic Reasoning & 43.28 & 61.80 & 50.32 \\
        & Text-Guided Symbolic Reasoning & 59.10 & 66.65 & 61.97 \\
        \bottomrule
    \end{tabular}
    }
    \label{tab:llm_reasoning}
\end{table}

We consider adaptive reasoning a key component in table understanding. Concurrent work, such as \cite{abhyankar2025hstarllmdrivenhybridsqltext}, also explores this direction.

Table~\ref{tab:llm_reasoning} compares different reasoning methods—textual reasoning, symbolic reasoning, and text-guided symbolic reasoning—across various LLMs under calculation-required and no-calculation-required scenarios. This experiment is conducted using \textit{gpt-4o-mini} on the WikiTQ dataset.

For \textit{gpt-4o}, textual reasoning achieves the highest accuracy (83.98\% overall), excelling in both calculation-required (81.17\%) and no-calculation-required (88.56\%) cases. Symbolic reasoning performs worse (74.63\% overall), while text-guided symbolic reasoning offers slight improvements (76.82\%). For \textit{gpt-4o-mini}, a similar trend is observed, with textual reasoning maintaining the highest accuracy (72.97\% overall), followed by text-guided symbolic reasoning (68.83\%), and symbolic reasoning performing the worst (61.83\%). For \textit{gpt-3.5-turbo}, performance drops significantly, with textual reasoning at 59.92\%, symbolic reasoning struggling at 50.32\%, and text-guided symbolic reasoning achieving the best results (61.97\%), indicating that symbolic guidance benefits weaker models.

Symbolic reasoning is consistently outperformed by textual reasoning, while text-guided symbolic reasoning surpasses textual reasoning only in \textit{gpt-3.5-turbo} under calculation-required scenarios. One reason for this is that not all calculation-required questions necessarily benefit from symbolic reasoning; for simple calculations, textual reasoning is more effective. However, for complex calculation-required questions, text-guided symbolic reasoning is the preferred approach. This provides a key insight for prompt design of reasoning strategy assessment.

Overall, textual reasoning consistently outperforms symbolic reasoning across all models, while text-guided symbolic reasoning helps mitigate weaker numerical capabilities in smaller models. These results suggest that adaptive reasoning should prioritize textual approaches, incorporating symbolic methods selectively for numerical calculations in weaker models.

\begin{table}[ht]
    \centering
    \caption{Performance and Inference Times for Different Methods}
    \label{tab:adaptive_reason}
    \begin{tabular}{lcc}
        \toprule
        \textbf{Method} & \textbf{Accuracy (\%)} & \textbf{Inference Times (\#)} \\
        \midrule
        Chain of Thought & 72.97 & 1 \\
        Program of Thought & 61.83 & 1 \\
        Text-Guided Program of Thought & 68.83 & 1 \\
        \midrule
        Self-Consistency (5 CoT) & 74.98 & 3 \\
        Self-Consistency (5 PoT) & 63.97 & 3 \\
        Mix Self-Consistency (3+3) & 76.70 & 6 \\
        Mix Self-Consistency (5+5) & 77.46 & 10 \\
        Self-Eval & 70.58 & 2 \\
        \midrule
        Adaptive Reasoning (POT) & 71.18 & 1 \\
        Adaptive Reasoning (TPOT) & 74.08 & 1 \\
        Adaptive Reasoning (POT - Upper Bound) & 82.99 & 1 \\
        Adaptive Reasoning (TPOT - Upper Bound) & 85.06 & 1 \\
        \bottomrule
    \end{tabular}
\end{table}

Table~\ref{tab:adaptive_reason} compares the performance (accuracy \%) and inference times of various reasoning methods, including chain of thought (CoT), program of thought (PoT), text-guided program of thought (TPoT), self-consistency, and adaptive reasoning. This experiment is conducted using \textit{gpt-4o-mini} on the WikiTQ dataset.

Among single-pass methods (1 inference), chain of thought achieves 72.97\% accuracy, outperforming program of thought (61.83\%) and text-guided program of thought (68.83\%). This suggests that CoT is more effective than pure symbolic reasoning when only one inference is allowed.

\begin{figure}[htbp]
    \centering
    \includegraphics[width=0.95\textwidth]{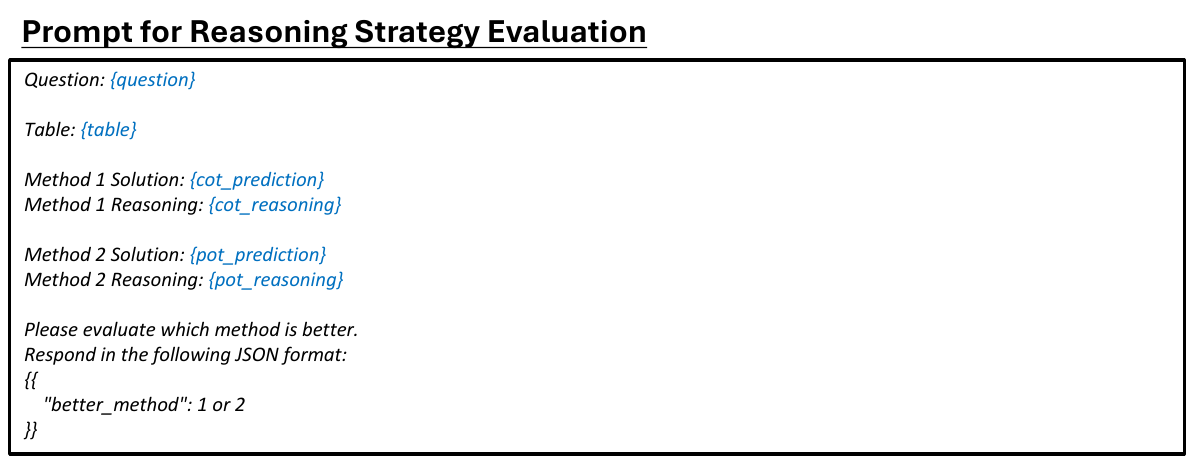}
    \centering
    \caption{Prompt for reasoning strategy evaluation in analysis experiment. Blue text indicates placeholders for variables within the prompt. The prompt guides the language model to select the better reasoning process after table reasoning.}
    \label{prompt:o_self_eval}
\end{figure}

Self-consistency methods, which perform multiple inferences to improve reliability, achieve better results. Five-shot CoT self-consistency reaches 74.98\%, while five-shot PoT self-consistency lags behind at 63.97\%. As introduced in \citep{liu-etal-2024-rethinking}, mixed self-consistency (3 CoT + 3 PoT) and (5+5) further improve accuracy to 76.70\% and 77.46\%, respectively, at the cost of increased inference time (6 and 10 passes). Self-evaluation (self-eval) first performs CoT and PoT inferences (Prompt~\ref{prompt:o_self_eval}), then selects the better result, achieving 70.58\% with 2 inferences.

Adaptive reasoning achieves competitive performance while maintaining single-pass efficiency. PoT-based adaptive reasoning reaches 71.18\%, while TPOT-based adaptive reasoning, which combines textual and text-guided symbolic methods, improves to 74.08\%. The upper-bound performance of these adaptive strategies (assuming perfect strategy selection) reaches 82.99\% (PoT) and 85.06\% (TPOT), significantly outperforming all other methods, highlighting the importance of textual guidance and strategy selection.

For the selection distribution between CoT and PoT (TPoT):  
\begin{itemize}
    \item \textbf{Self-eval:} 1,962 PoT and 2,382 CoT  
    \item \textbf{Adaptive reasoning (PoT):} 1,590 PoT and 2,754 CoT  
    \item \textbf{Adaptive reasoning (TPoT):} 1,590 PoT and 2,754 CoT  
    \item \textbf{Adaptive reasoning (PoT - upper bound):} 435 PoT and 3,909 CoT  
    \item \textbf{Adaptive reasoning (TPoT - upper bound):} 525 PoT and 3,819 CoT  
\end{itemize}  

These results suggest that language models should prioritize textual reasoning and reserve symbolic reasoning for more complex numerical calculations where it provides a clear advantage.

Overall, self-consistency enhances accuracy but requires multiple inferences, whereas adaptive reasoning effectively balances accuracy and efficiency. To further improve strategy assessment, we will explore ways to approach this upper bound in future work. This demonstrates that well-designed adaptive reasoning strategies can rival more computationally expensive self-consistency methods while maintaining efficiency.

\section{Information Missing and Table Reasoning with Full Table}
\label{ap:missing}
\begin{table}[h]
    \centering
    \caption{Performance comparison of reasoning with and without the full table on WikiTQ and TabFact.}
    \label{tab:full_table}
    \begin{tabular}{lcc}
        \toprule
        \textbf{Method} & \textbf{WikiTQ} & \textbf{TabFact} \\
        \midrule
        PoTable (Previous SOTA) \citep{mao2024potableprogrammingstandardlytablebased} & 64.73 & 88.93 \\
        TableMaster w/ Full Table in Reasoning & 78.13 & 90.12 \\
        TableMaster w/o Full Table in Reasoning & 77.23 (-0.90) & 89.58 (-0.54) \\
        \bottomrule
    \end{tabular}
\end{table}

As discussed in our limitations, the table-of-focus process may sometimes lead to the loss or omission of key relevant information. This issue is inevitable when attempting to locate specific data. If no relevant data exists within the selected portion, the reasoning result will naturally be incorrect.

In our experiments, we found that when using the table-of-focus and its verbalized representation for reasoning, 265 out of 4,344 questions in the WikiTQ dataset had no available answers. This led to a performance drop, as the language model responded with an inability to provide an answer. To address this, we replaced the table-of-focus with the original full table, combined with verbalized table-of-focus as input in those questions. The performance under this adjustment is shown in Table~\ref{tab:full_table}, reaching 77.23\% in WikiTQ. When we directly replaced the table-of-focus with the full table for all questions in WikiTQ, the performance increased to 78.13\%, resulting in a slight improvement of 0.9\%. Two results are similar.

We believe this approach does not contradict previous steps such as structure extraction and table-of-focus selection. These steps remain valuable, as the extracted target data is retained in the verbalized table, where the information density is higher and semantic context is richer. The language model prioritizes this high-density information, and if it is insufficient, it can then reference the global information from the full table. This demonstrates the complementary nature of the full table and the verbalized table-of-focus. From an efficiency perspective, it is preferable to use the sub-table for reasoning initially and only switch to the full table when necessary.

To highlight the performance of \method, we report the best scores of 78.13 and 90.12 in the main results table. Regardless, our method consistently outperforms the previous state-of-the-art, PoTable \citep{mao2024potableprogrammingstandardlytablebased}, on both WikiTQ and TabFact.

\section{Case Study}
\subsection{Case Study of Table Verbalization}
\label{ap:case_study_verbal}  
\begin{figure*}[htbp]
    \centering
    \includegraphics[width=\textwidth]{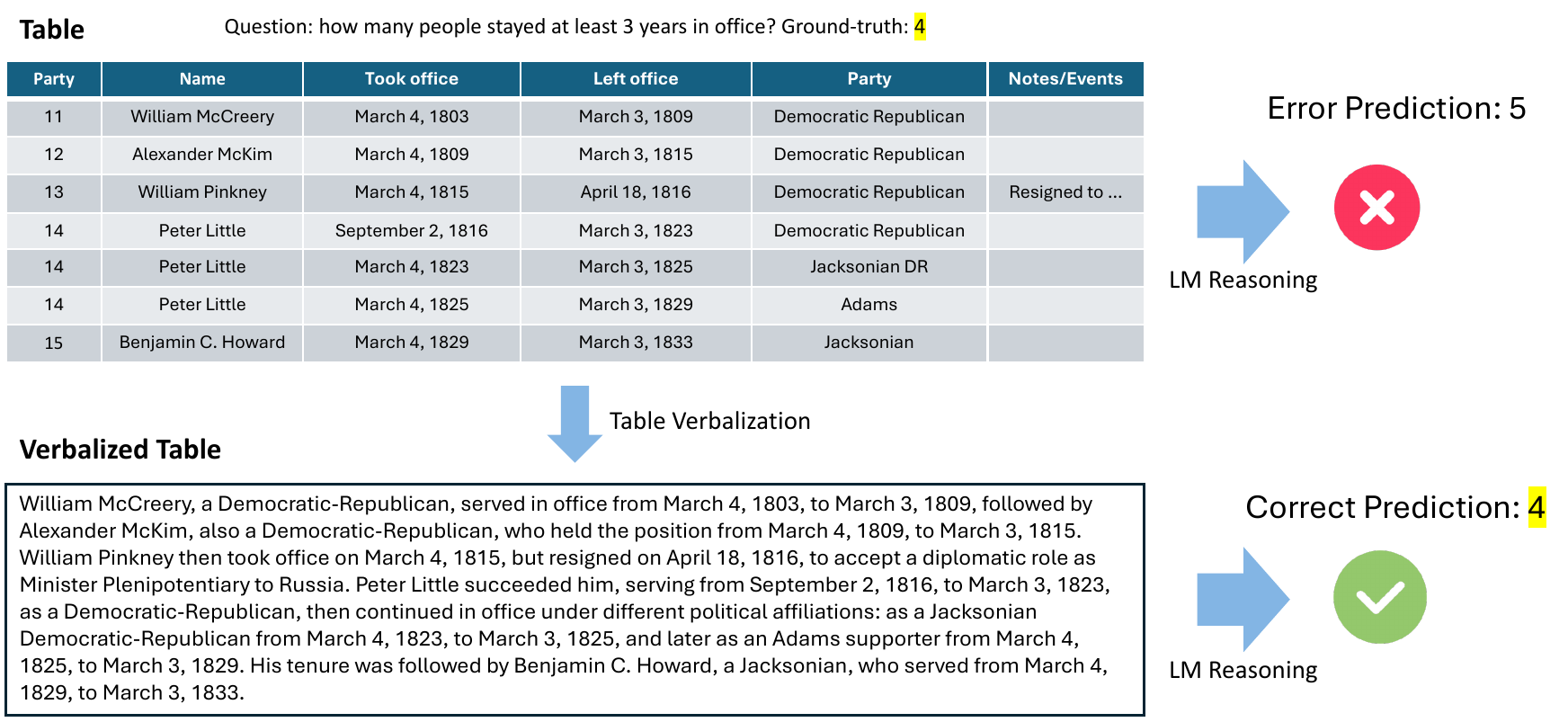}
    \caption{Case study on the impact of table verbalization. The data is from the WikiTQ dataset.}
    \label{fig:case_study_verbal}
\end{figure*}

Table verbalization brings a slight overall improvement in table understanding and is particularly effective in cases where deeper comprehension of the table’s context is required to answer questions accurately.

Figure~\ref{fig:case_study_verbal} presents a case study on the impact of table verbalization in helping language models reason about structured data. The setup includes a table listing U.S. congressional representatives, their terms in office, political affiliations, and notable events. The question posed is: \textit{How many people stayed at least 3 years in office?}, with a ground-truth answer of 4.

When the table is input directly, the model incorrectly predicts 5, as it mistakenly counts rows rather than identifying unique individuals. This suggests that the model relies on simple row counting instead of truly understanding the data. However, with the verbalized table, the model accurately interprets the descriptions, grasps the actual meaning, and correctly answers with 4.

\subsection{Case Study of \method}  
\label{ap:case_study_tm}  

\begin{figure*}[htbp]
    \centering
    \includegraphics[width=\textwidth]{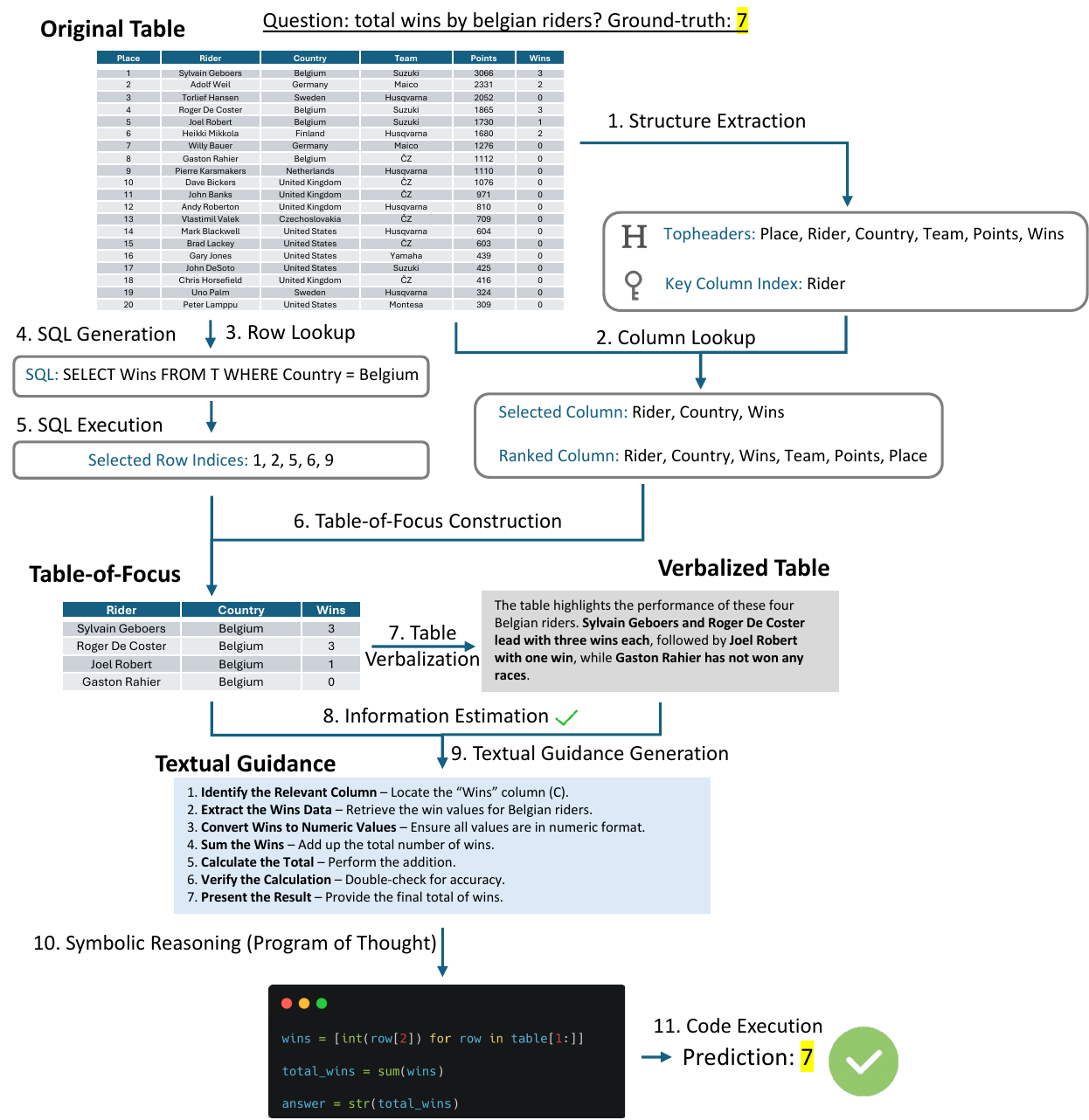}
    \caption{Case study of \method. The data is from the WikiTQ dataset.}
    \label{fig:case_study_tm}
\end{figure*}

As shown in Figure~\ref{fig:case_study_tm}, we present a case study of \method to illustrate its detailed workflow in answering the question: “Total wins by Belgian riders?” with a ground-truth answer of 7. The process begins with structure extraction, identifying key columns such as Rider, Country, and Wins. Next, column lookup selects relevant data, and row lookup filters the rows containing Belgian riders. SQL generation and execution retrieve only the relevant records where Country = Belgium.

The refined table is then constructed into a table-of-focus, keeping only the necessary columns. Table verbalization converts structured data into a description to enrich semantic context, providing insights into the number of wins for each Belgian rider. A textual guidance module generates a structured step-by-step explanation of the counting process, ensuring clarity in symbolic reasoning. The reasoning and execution phase involves symbolic reasoning (Program of Thought), where a Python snippet correctly extracts and sums the wins, leading to the correct prediction of 7.

This case study highlights \method’s ability to accurately extract, process, and reason over structured data, demonstrating its effectiveness in table-based question answering with a combination of structured queries, reasoning steps, and code execution.

\clearpage
\section{Prompt Design in \method}
\label{ap:tm_prompt}

\begin{figure}[htbp]
    \centering
    \includegraphics[width=0.95\textwidth]{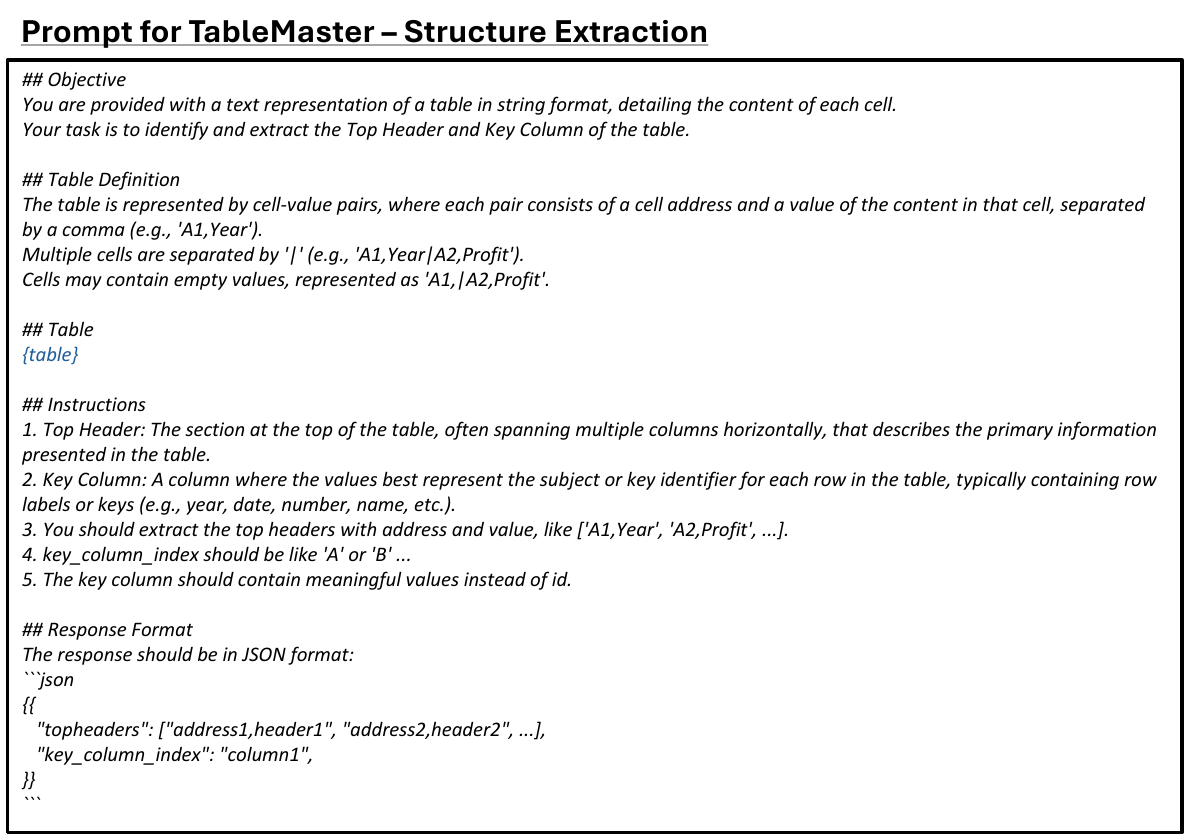}
    \centering
    \caption{Prompt for structure extraction in \method. Blue text indicates placeholders for variables within the prompt. The prompt guides the language model in extracting the table's structure.}
    \label{prompt:tm_structure_extraction}
\end{figure}

\begin{figure}[htbp]
    \centering
    \includegraphics[width=0.95\textwidth]{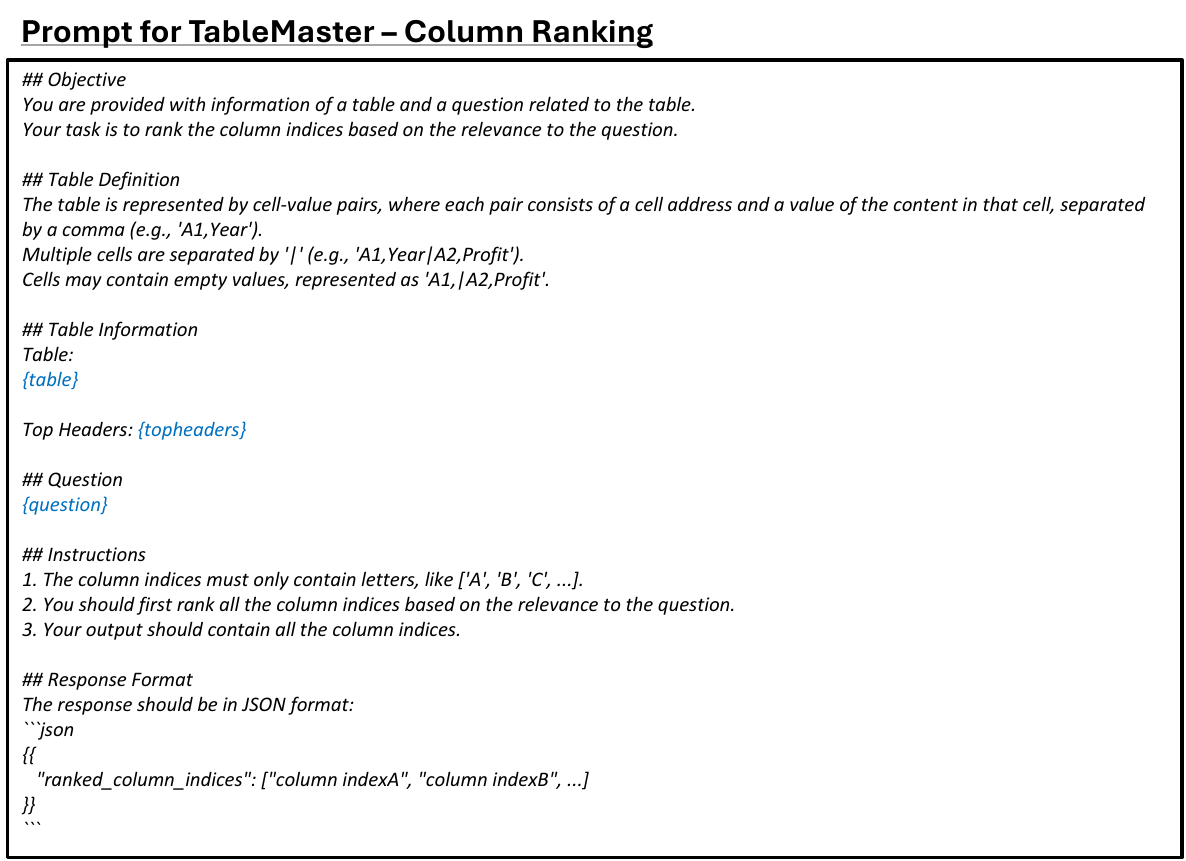}
    \centering
    \caption{Prompt for column ranking in \method. Blue text indicates placeholders for variables within the prompt. The prompt guides the language model to rank the priority of all columns based on the given table, top headers, and related question.}
    \label{prompt:tm_column_ranking}
\end{figure}

\begin{figure}[htbp]
    \centering
    \includegraphics[width=0.95\textwidth]{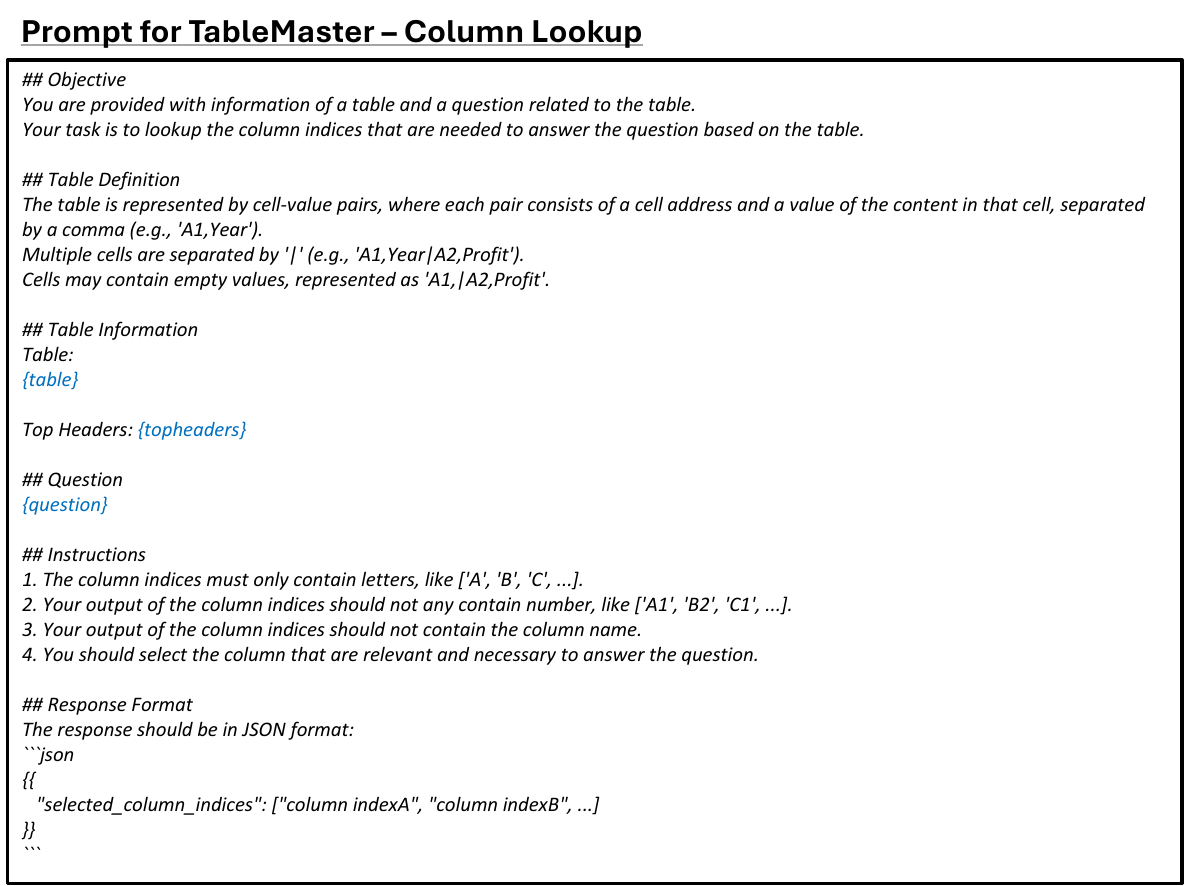}
    \centering
    \caption{Prompt for column lookup in \method. Blue text indicates placeholders for variables within the prompt. The prompt guides the language model to select relevant columns based on the given table, top headers, and related question.}
    \label{prompt:tm_column_lookup}
\end{figure}

\begin{figure}[htbp]
    \centering
    \includegraphics[width=0.95\textwidth]{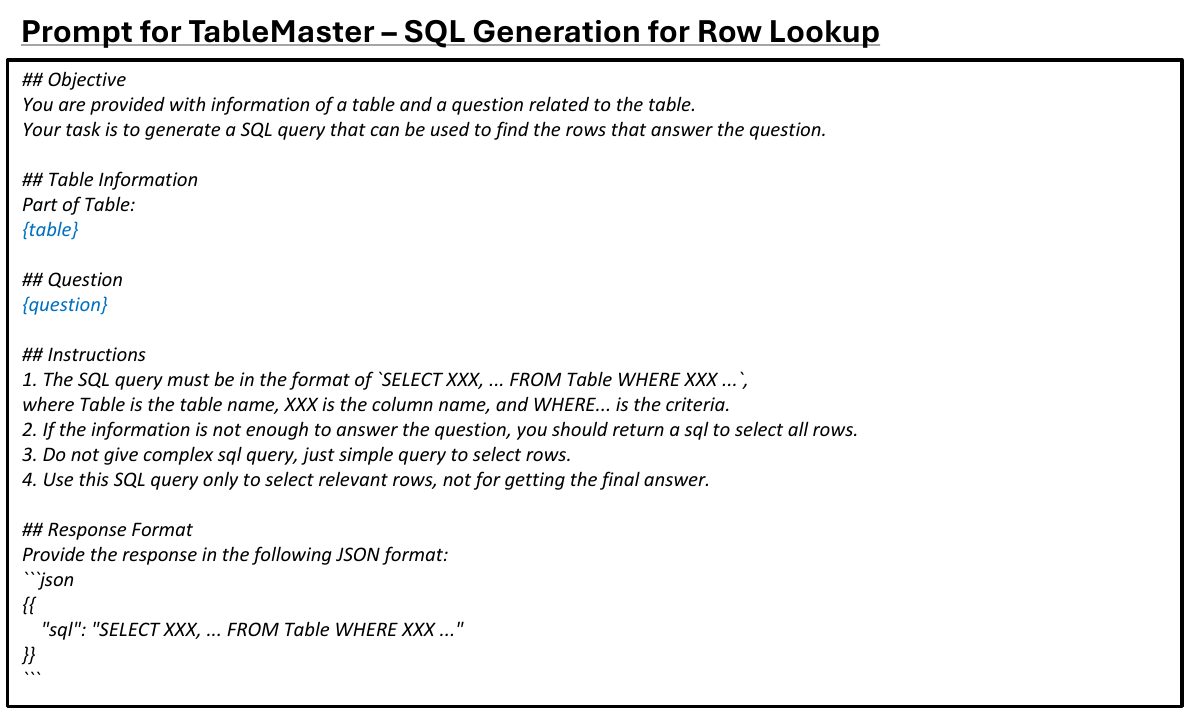}
    \centering
    \caption{Prompt for SQL generation for row lookup in \method. Blue text indicates placeholders for variables within the prompt. The prompt guides the language model to generate SQL for selecting relevant rows based on the given table and related question.}
    \label{prompt:tm_row_lookup}
\end{figure}

\begin{figure}[htbp]
    \centering
    \includegraphics[width=0.95\textwidth]{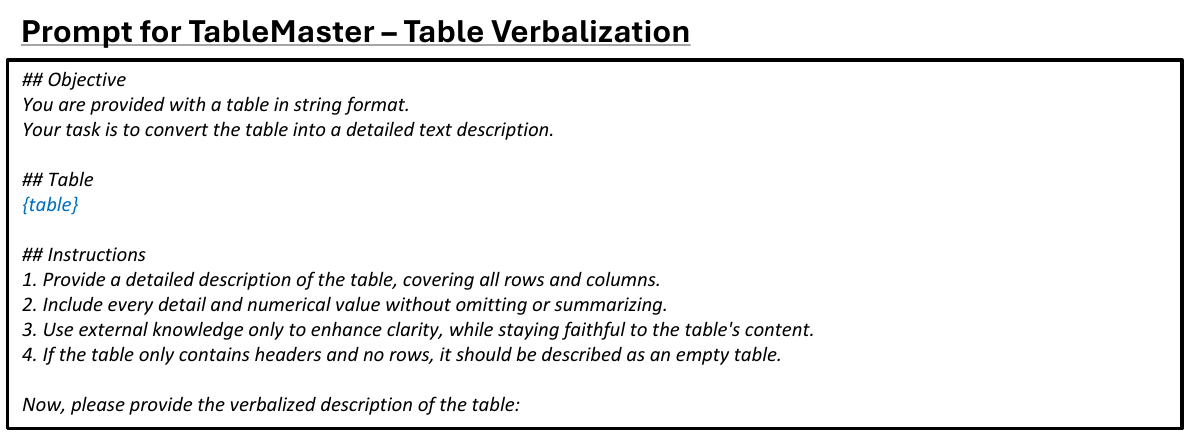}
    \centering
    \caption{Prompt for table verbalization in \method. Blue text indicates placeholders for variables within the prompt. The prompt guides the language model to verbalize the given table by adding detailed descriptions and additional knowledge about the table.}
    \label{prompt:tm_table_verbalization}
\end{figure}

\begin{figure}[htbp]
    \centering
    \includegraphics[width=0.95\textwidth]{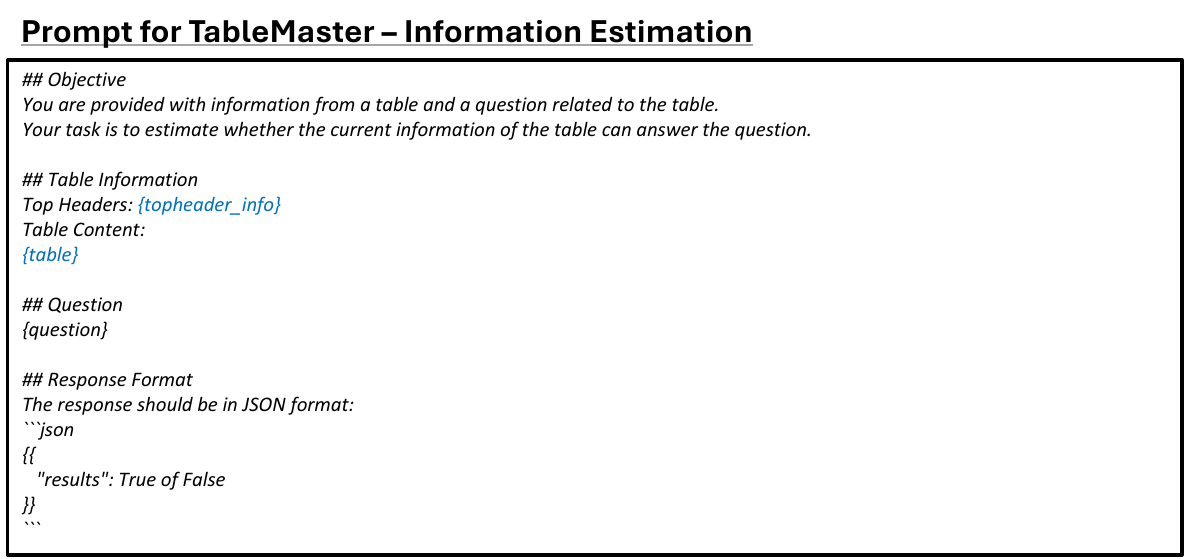}
    \centering
    \caption{Prompt for information estimation in \method. Blue text indicates placeholders for variables within the prompt. The prompt guides the language model to evaluate the given table’s content and determine whether it contains sufficient information to answer the provided question}
    \label{prompt:tm_information_estimation}
\end{figure}

\begin{figure}[htbp]
    \centering
    \includegraphics[width=0.95\textwidth]{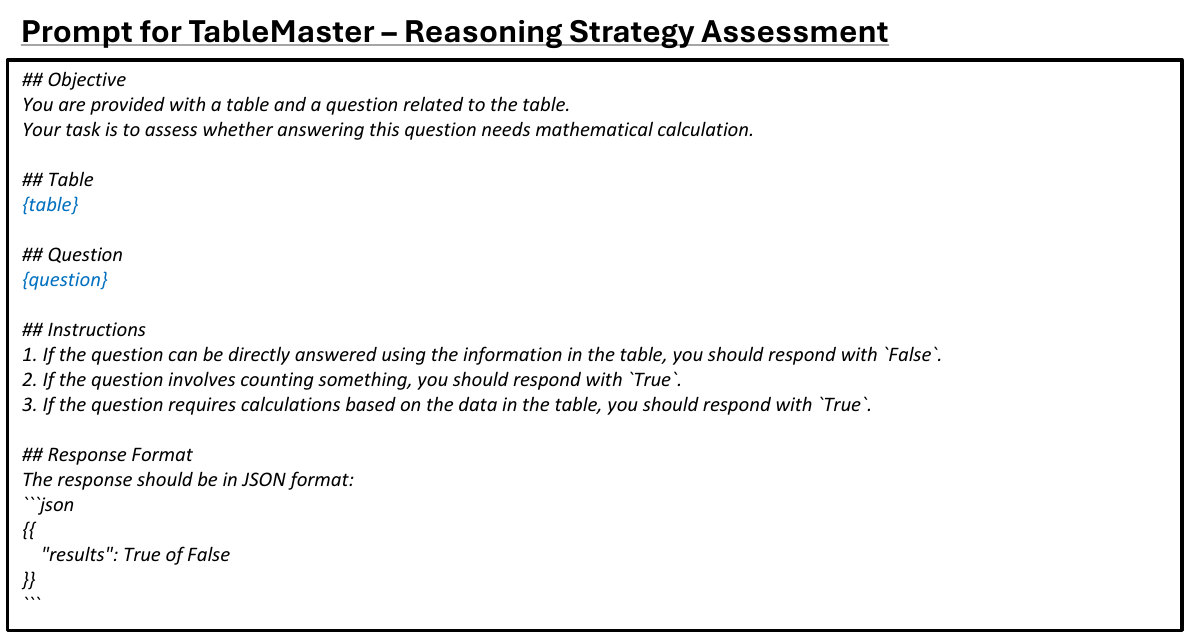}
    \centering
    \caption{Prompt for reasoning strategy assessment in \method. Blue text indicates placeholders for variables within the prompt. The prompt guides the language model to evaluate whether answering the given question requires direct information retrieval, counting, or mathematical calculations based on the table’s content. The response determines the subsequent reasoning strategy.}
    \label{prompt:tm_strategy_assessment}
\end{figure}

\begin{figure}[htbp]
    \centering
    \includegraphics[width=0.95\textwidth]{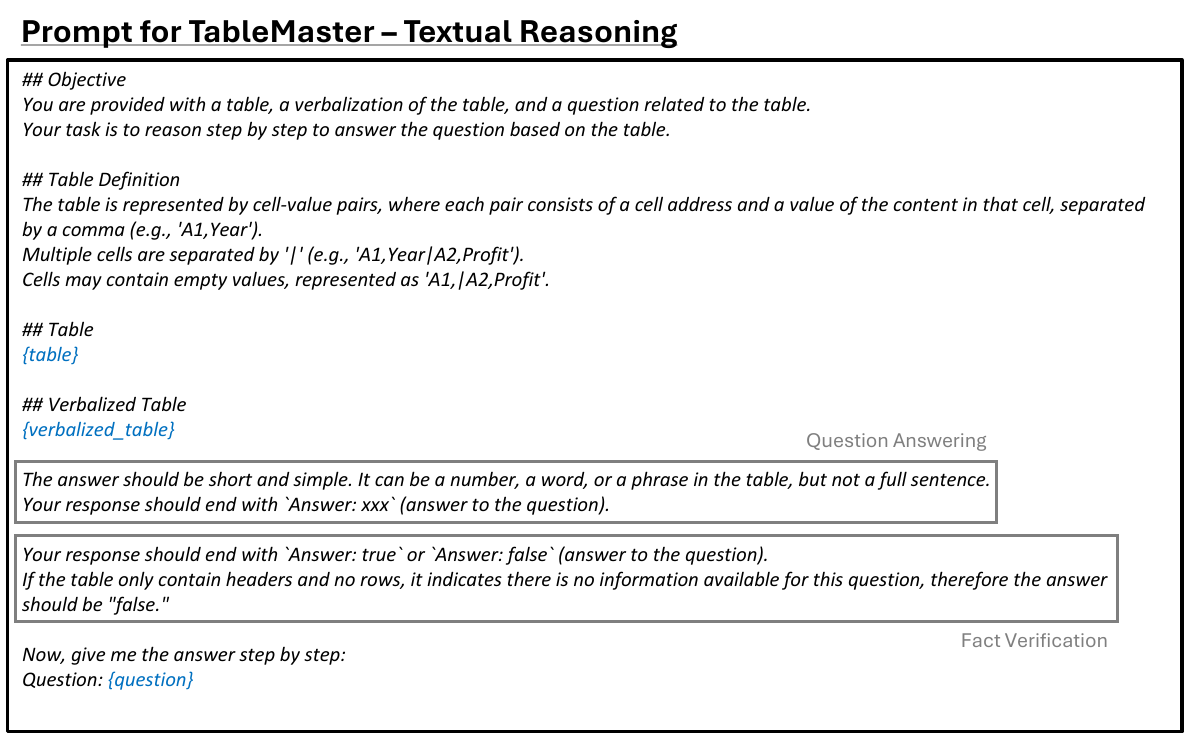}
    \centering
    \caption{Prompt for textual reasoning in \method. Blue text represents placeholders for variables within the prompt, while the grey region indicates optional sections to adapt the prompt for question-answering or fact-verification tasks. The prompt guides the language model to answer the question step by step.}
    \label{prompt:tm_textual_reasoning}
\end{figure}

\begin{figure}[htbp]
    \centering
    \includegraphics[width=0.95\textwidth]{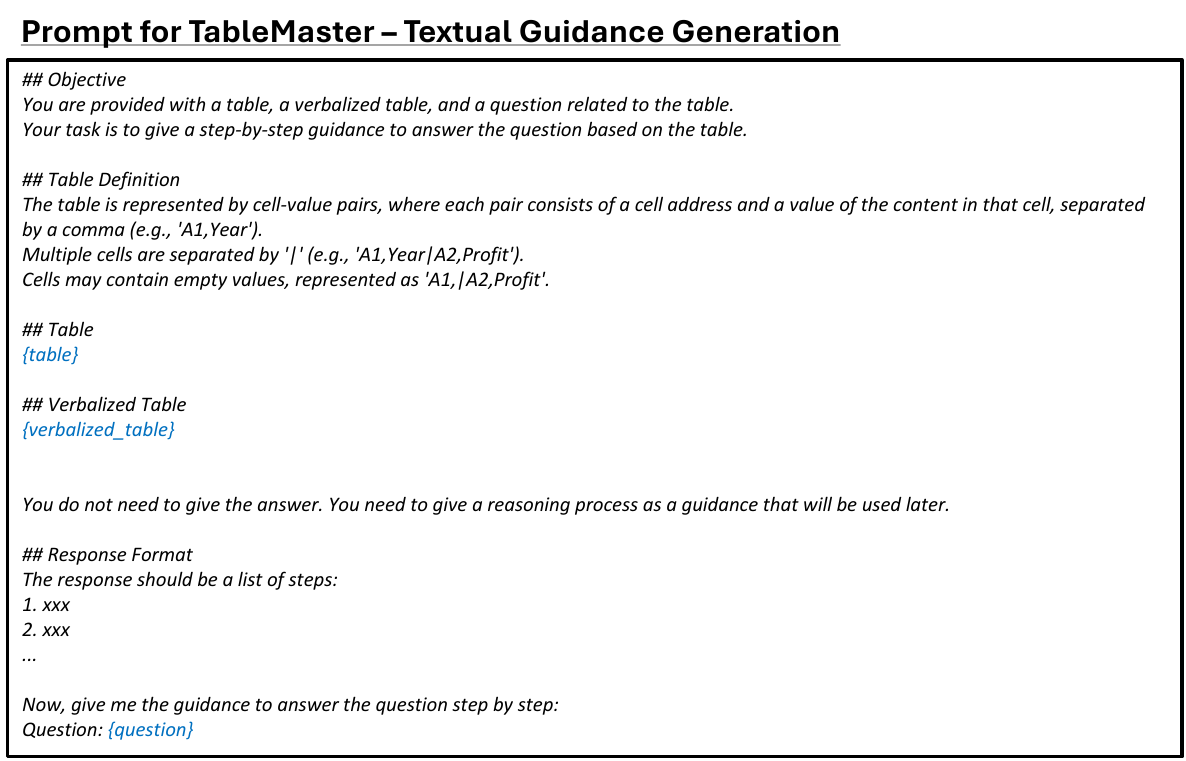}
    \centering
    \caption{Prompt for textual guidance generation in \method. Blue text indicates placeholders for variables within the prompt. The prompt guides the language model to generate textual guidance that can be utilized for subsequent symbolic reasoning.}
    \label{prompt:tm_textual_guidance}
\end{figure}

\begin{figure}[htbp]
    \centering
    \includegraphics[width=0.95\textwidth]{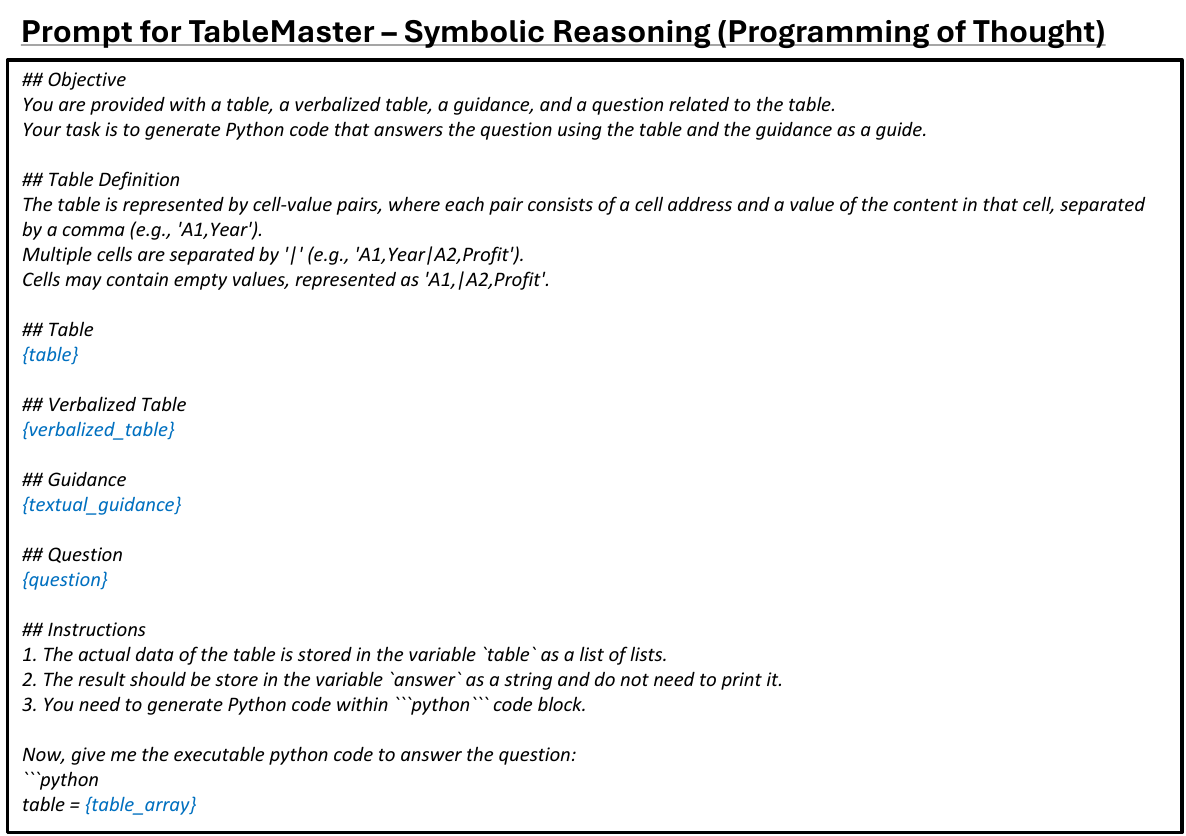}
    \centering
    \caption{Prompt for symbolic reasoning in \method. Blue text indicates placeholders for variables within the prompt. The prompt guides the language model to generate Python code to answer the question.}
    \label{prompt:tm_symbolic_reasoning}
\end{figure}

\begin{figure}[htbp]
    \centering
    \includegraphics[width=0.95\textwidth]{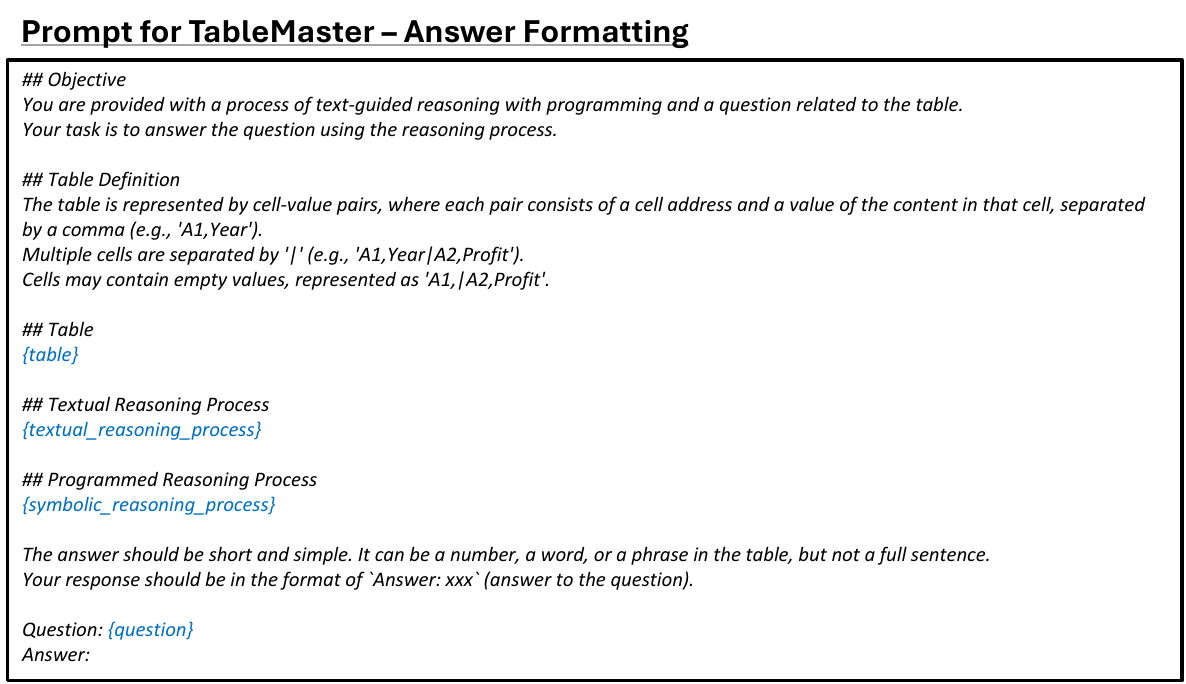}
    \centering
    \caption{Prompt for answer formatting in \method. Blue text indicates placeholders for variables within the prompt. The prompt guides the language model to format the final answer based on the given table, question, and reasoning process.}
    \label{prompt:tm_answer_formatting}
\end{figure}

\clearpage
\section{Prompts Used in Analysis Experiments}
\label{ap:o_prompt}

\begin{figure}[htbp]
    \centering
    \includegraphics[width=0.95\textwidth]{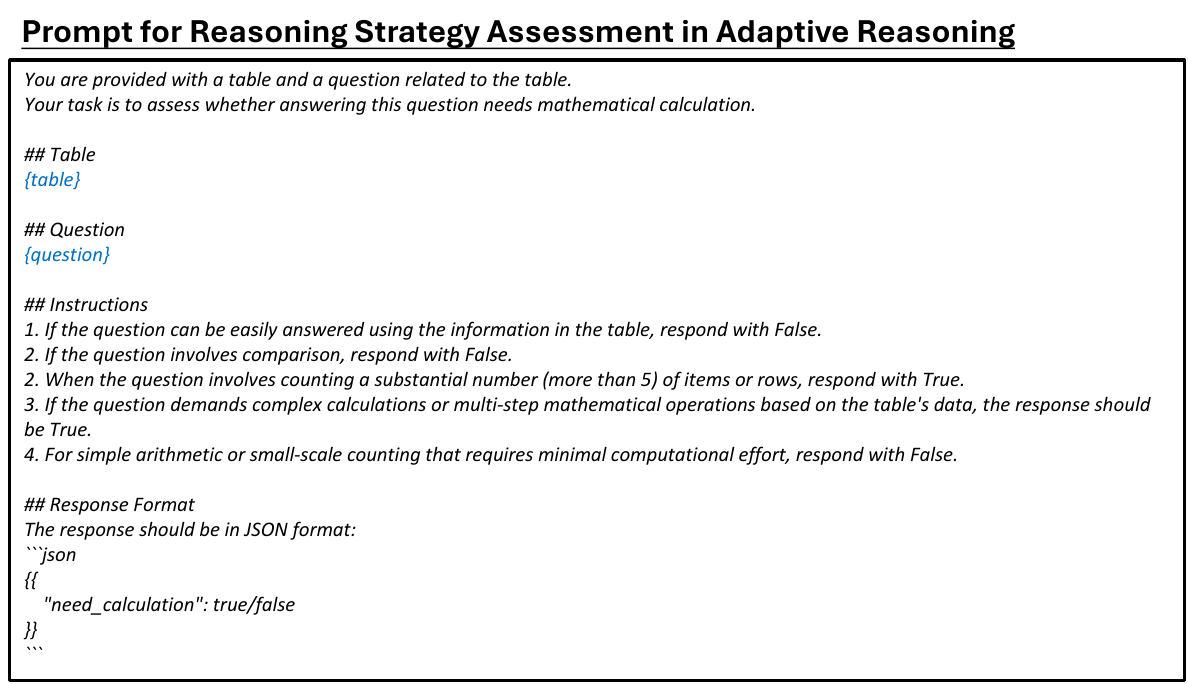}
    \centering
    \caption{Prompt for reasoning strategy evaluation in analysis experiment. Blue text indicates placeholders for variables within the prompt. The prompt guides the language model to select the better reasoning strategy before table reasoning.}
    \label{prompt:o_adaptive}
\end{figure}

\clearpage
\section{Notation Table}
\label{ap:notation}

Table~\ref{tab:notation} provides a comprehensive list of the notations used throughout this paper, along with their corresponding descriptions. This table serves as a quick reference to help readers better understand
the concepts presented in our work.

\begin{table*}[h]
\caption{Notation used throughout the paper}
\label{tab:notation}
\begin{center}
\begin{tabular}{cl}
\toprule
\textbf{Notation} & \textbf{Description} \\
\midrule
\multicolumn{2}{l}{\textit{General}} \\
$Q$ & Given question or query \\
$A$ & Generated answer \\
$\mathbb{T}$ & Input table \\
$\mathbb{T}^{W}$ & Wild table before normalization \\
$\mathbb{T}^{N}$ & Normalized table \\
$\mathbb{T}^{F}$ & Table-of-Focus \\
$C_{i,j}$ & Cell in the $i$-th row and $j$-th column \\
$m, n$ & Number of rows and columns in the table \\
\midrule
\multicolumn{2}{l}{\textit{Table Structure Understanding}} \\
$H$ & Set of top headers \\
$K$ & Key column serving as row identifier \\
$\mathbb{C}$ & Candidate column set \\
$C^0$ & Selected relevant columns \\
$R$ & Selected relevant rows \\
$k$ & Peek size for table processing \\
\midrule
\multicolumn{2}{l}{\textit{Table Content Understanding}} \\
$T^{\mathbb{T}}$ & Verbalized table (natural language text) \\
$a', b'$ & Number of refined columns and rows after reconstruction \\
\midrule
\multicolumn{2}{l}{\textit{Table Reasoning}} \\
$S$ & Selected reasoning strategy \\
$\mathcal{T}$ & Textual reasoning strategy \\
$\mathcal{S}$ & Symbolic reasoning strategy \\
$G$ & Textual reasoning guidance \\
$\mathcal{P}$ & Program executor (Python/SQL) \\
\bottomrule
\end{tabular}
\end{center}
\end{table*}

\clearpage
\section{The Use of Large Language Models (LLMs)}
\label{ap:iclr_llm_state}
In this work, large language models (LLMs) were used \emph{only to aid with writing and polishing the manuscript}. Specifically, LLMs were employed for grammar correction, phrasing suggestions, and improving readability. All research ideas, methodological contributions, theoretical analyses, and experiments were entirely conceived, designed, and executed by the authors without the involvement of LLMs. The authors take full responsibility for the scientific content of the paper.


\end{document}